\documentclass[journal]{IEEEtran}
\usepackage{amssymb}

\usepackage[cmex10]{amsmath}

\usepackage{pifont,calc}
\usepackage{cite}

\usepackage{graphicx}
\usepackage{array}
\usepackage[caption=false]{subfig}
\usepackage{caption,setspace}

\usepackage{cases} 
\usepackage{bm} 

\usepackage{mathrsfs}
\usepackage{url}
\usepackage{ulem}
\usepackage{algorithm}
\usepackage{algpseudocode}

\usepackage{arydshln}
\usepackage{multirow}
\usepackage{multicol}

\makeatletter

\newcommand{\Rmnum}[1]{\expandafter\@slowromancap\romannumeral #1@}
\makeatother

\ifCLASSINFOpdf
\else
\fi

\begin{document}
%
\title{Two Ridge Solutions for the Incremental Broad Learning
System on Added Nodes} 
%
%
%

\author{Hufei~Zhu
\thanks{H. Zhu is with the Faculty of Intelligent Manufacturing, Wuyi University, Jiangmen 529020, Guangdong, China (e-mail:
hufeizhu93@hotmail.com).}
}

%
%

\markboth{Journal of \LaTeX\ Class Files,~Vol.~14, No.~8, August~2015}%
{Shell \MakeLowercase{\textit{et al.}}: Bare Demo of IEEEtran.cls for IEEE Journals}
%

%
%
%
%
%



\maketitle

\begin{abstract}

The original Broad Learning System (BLS) on new added nodes  and its existing efficient implementation both assume the ridge parameter $\lambda \to 0$
in the ridge inverse to approximate
 the generalized
inverse, and compute    the generalized
inverse solution for the output weights.
In this paper,
   we propose two ridge solutions
   for the output weights in the BLS on added nodes,
        where $\lambda \to 0$ is no longer assumed, and  $\lambda$ can be  any positive real number.
 One of  the proposed ridge solutions
computes the
     output weights from the inverse Cholesky factor,
     which is updated efficiently by
     extending the existing inverse Cholesky factorization.
     The other proposed ridge solution
     computes
     the
     output weights from the ridge inverse, and updates the ridge inverse
     by extending the Greville's method
     that is a classical tool to compute the generalized inverse of partitioned matrices.
     For the proposed efficient ridge solution based on the inverse Cholesky factor,
     we also develop another implementation that is numerically more stable when the ridge parameter $\lambda$ is very small.


   The proposed ridge solution based on the ridge inverse
   and the numerically more stable implementation of the proposed efficient ridge solution
   require the same complexity as the original BLS and the existing efficient
    BLS, respectively.
   Moreover, the speedups
  of the proposed efficient ridge solution
  to the original BLS  and the existing efficient BLS
  are
  $300\%$
   and more than
   $167\%$
    respectively, when the computational complexities for each update are compared,
  and the speedups are
  $199\% \sim 252\%$
    and $131\% \sim 158\%$,
      respectively, when the total training time is compared by  numerical experiments.
  On the other hand, our numerical experiments
    show that   both the proposed ridge solutions
   for  BLS achieve better testing accuracies than the original BLS and the existing efficient
   BLS.
\end{abstract}


%
\IEEEpeerreviewmaketitle

%
%

%
%

%
%
%
%

\section{Introduction}
%
%
%
%
%
%
%
Deep neural networks (DNNs), including the deep belief networks
(DBN) \cite{DBNhinton},\cite{Hinton504}, the deep Boltzmann machines (DBM) \cite{salakhutdinov2009deep}, and
the convolutional neural networks (CNN) \cite{726791, DBLP:journals/corr/SimonyanZ14a},  have been successfully adopted in many applications \cite{7120131,6872541},
 particularly in  image and speech recognition~\cite{ DeepLearningImage, DeepLearningSpeech}.
However, most DNNs suffer from the time-consuming training process, since
  usually they possess complicated structures and need to tune a huge number of hyperparameters by back propagation of error~\cite{BL_Ref_19,BL_Ref_22}.
  Furthermore, DNNs are  treated as black boxes in most cases, which makes theoretic analysis of DNNs very difficult.

Single layer feedforward neural networks (SLFN)
have been
widely applied in classification and
regression because of their universal approximation capability~\cite{BL_Ref_18,BL_Ref_19,BL_Ref_20}.
 The traditional
Gradient-descent-based learning algorithms~\cite{BL_Ref_22,BL_Ref_23}
can be utilized to adjust
the input and output weights of SLFN.
However, the Gradient-descent-based  algorithms
 are very slow in the training process,  and may easily halt at local optima.
 Furthermore, their generalization performance is
 much
  affected by
parameter settings
 (e.g., the setting of learning rate).  Then
 the random vector functional-link neural network (RVFLNN)
has been proposed~\cite{BL_Ref_19} as
 a different train method,
 which generates the
 input weights and the biases randomly, and learns only the   output weights.
  RVFLNN
eliminates the disadvantage of slow training process,
 and
also offers the generalization capability in function approximation~\cite{BL_Ref_20}.


For a new added node or input in the RVFLNN,
the dynamic step-wise updating algorithm   proposed in \cite{27_ref_BL_trans_paper}
  can update the output weights easily by only computing the generalized inverse of that added node or input,  which is suitable for  time-variety data with moderate size.
Then  to process time-variety big data with
high dimension,
the scheme proposed in \cite{27_ref_BL_trans_paper}
 was improved in \cite{BL_trans_paper}
 to propose the Broad Learning
System (BLS). For BLS,  several BLS variants have
 been proposed,
 which include cascade, recurrent, and broad-deep combination structures,
and the
 proof  of the universal approximation capability  has been given~\cite{BL_trans_paperApproximate}.

In the original BLS algorithm~\cite{BL_trans_paper},  the output weights  are  the generalized
inverse solution~\cite{best_ridge_inv_paper213}
computed from
 the generalized
inverse, and the generalized
inverse is
updated efficiently by the Greville's method~\cite{cite_general_inv_book}. Recently in \cite{BLSpaper2021zhf1}, we proposed an efficient implementation of the original BLS algorithm on added nodes,
which computes the output weights  from the inverse Cholesky factor of the Hermitian matrix
in
 the generalized inverse,  and updates  the inverse Cholesky factor efficiently.
With respect to the original BLS,
the efficient BLS in \cite{BLSpaper2021zhf1}
 requires less than  $\frac{2}{3}$ of
  complexity, and
achieves the same
testing accuracy
when the tiny differences caused
by the numerical errors are neglected.
Both the BLS algorithms in \cite{BL_trans_paper,BLSpaper2021zhf1}
 are based on the generalized inverse with the ridge regression approximation, where
  the ridge parameter $\lambda \to 0$ is assumed in the ridge inverse~\cite{best_ridge_inv_paper213}
  to
  approximate
   the generalized inverse.

   In this paper,
   we will
   propose two ridge solutions~\cite{best_ridge_inv_paper213}
   for the BLS
   with new
    added nodes,
        where it is no longer required to assume the ridge parameter $\lambda \to 0$ (i.e.,  $\lambda$ can be  any positive real
number).        
In one of  the proposed ridge solutions,
the output weights
are computed
   from
 the inverse Cholesky factor of the
   Hermitian matrix in
    the ridge inverse,
  and   the inverse Cholesky factorization~\cite{my_inv_chol_paper} is extended to update the inverse Cholesky factor efficiently.
  In the other proposed ridge solution,
  the output weights are computed from the ridge inverse, and
  the Greville's method~\cite{cite_general_inv_book} for the generalized inverse
  is extended to update  the ridge inverse.
    For the proposed efficient ridge solution based on the inverse Cholesky factor,
     we also develop another implementation that is numerically more stable when the ridge parameter $\lambda$ is very small.
      In simulations,
      both the proposed  ridge solutions
    usually achieve better testing accuracy than the existing generalized inverse solutions~\cite{BL_trans_paper,BLSpaper2021zhf1} for BLS.
    Moreover,
  the  proposed efficient ridge solution
 requires less complexity than  the efficient generalized inverse solution in \cite{BLSpaper2021zhf1}.




 Since the training processing of big data with high dimension may exceed the capacity of a single computational node, there is a need to distribute computing tasks across multiple computational nodes, which can also be called as workers~\cite{SurveyOnDistributedML}.
Specifically, a distributed implementation is usually necessary if data is inherently distributed or too big to store on a single worker.
In distributed systems, algorithms have to be chosen and implemented to enable parallel computation.
In the distributed machine learning systems,   two fundamental parallelization approaches
are data-parallelism and model-parallelism~\cite{SurveyOnDistributedML}, which parallelize the data and the model, respectively,
and can be applied simultaneously.  Data-parallelism partitions training samples  into multiple workers
and then all workers apply nearly the same algorithms to different groups of training samples.
On the other hand,  model-parallelism partitions the model into multiple workers, of which each
processes the exact copy of all training samples and operates on different parts of the model.
In this paper, we will introduce the memory-saving parallel implementation of the proposed ridge solution based on the inverse
Cholesky factor, and try to avoid the square-root and division operations in that implementation.

  This paper is organized as follows. Section \Rmnum{2} introduces the existing
  generalized inverse solutions for the
   BLS with added nodes,
  and summarizes the construction model and learning procedure of BLS.
  Then for BLS,
  we propose the  ridge solution by inverse Cholesky factorization in Section \Rmnum{3},
 and propose the  ridge solution
 based on the ridge inverse in Section \Rmnum{4}.
Section \Rmnum{5}  compares the expected computational complexities of the presented BLS algorithms,
evaluates these BLS algorithms by numerical experiments, and introduces several implementation aspects of
  the inverse
Cholesky factorization, which include parallelization, memory saving, square-root free and division free.
Finally, we make conclusion in Section \Rmnum{6}.

\section{Existing Generalized Inverse Solutions for the BLS with Added Nodes}

In RVFLNN,
the expanded input matrix ${{\mathbf{A}}}=\left[ {{\mathbf{X}}}| \xi ({{\mathbf{X}}}{{\mathbf{W}}_{{{h}}}}+{{\mathbf{\beta }}_{{{h}}}}) \right]$,
where 
${{\mathbf{W}}_{{{h}}}}$ and
        ${{\mathbf{\beta }}_{{{h}}}}$  are random,
${{\mathbf{X}}}$ is  the input data,
        and $\xi$  is
the activation function.
The corresponding output
\begin{equation}\label{Y2AW65897}
{\mathbf{\hat{Y}}}={{\mathbf{A}}}{{\mathbf{W}}},
 \end{equation}
 where ${{\mathbf{W}}}$ is the output weight matrix.
For (\ref{Y2AW65897}),
the generalized inverse solution~\cite{best_ridge_inv_paper213}
  is
\begin{equation}\label{W2AinvY989565}
\mathbf{\bar W}=\mathbf{A}_{{}}^{+ }\mathbf{Y},
\end{equation}
where ${{\mathbf{Y}}}$ denotes the  labels and the generalized inverse
 \begin{equation}\label{Ainv2AtAinvAt9096}
\mathbf{A}_{{}}^{+}={{(\mathbf{A}_{{}}^{T}\mathbf{A})}^{-1}}\mathbf{A}_{{}}^{T}.
 \end{equation}

 \subsection{Ridge Regression Approximation of the Generalized Inverse}

In a flatted neural network, the generalized inverse solution (\ref{W2AinvY989565})
 is equivalent to the least square solution
\begin{equation}\label{W2WAWYls54579}
\mathop {argmin:}\limits_{\bf{\bar W}} {\left\| {{\bf{A}}{\bf{\bar W}}  - {\bf{Y}}} \right\|_2}
 \end{equation}
 of the linear equation
 (\ref{Y2AW65897}),
 which
 is
a very convenient approach to obtain the output weights~\cite{27_ref_BL_trans_paper, BL_trans_paper}.
Although the least square solution (\ref{W2WAWYls54579})
is aimed to minimize training errors,  usually it can not achieve the minimum generalization
errors, especially for ill-conditioned problems.
To
improve
the generalization performance, instead of the least square solution (\ref{W2WAWYls54579}),
we can use
the regularized
least-square solution
\begin{equation}\label{MinWAWYlamdaW95894}
\mathop {argmin:}\limits_{\bf{ W}} \left\| {{\bf{A  W}} - {\bf{Y}}} \right\|_2^2 + \lambda \left\| {\bf{ W}} \right\|_2^2,
 \end{equation}
where $\left\| \bullet \right\|_2^2$ denotes the $l_2$ norm, and
$\lambda>0$
 is the constraint on the sum of the
squared weights ${\bf{W}}$. The solution (\ref{MinWAWYlamdaW95894}) is equivalent to
the ridge solution~\cite{best_ridge_inv_paper213}
 \begin{equation}\label{W2AY9845729}
\mathbf{W} = {{{{\mathbf{A}}}}^{\dagger}}{{\mathbf{Y}}},
\end{equation}
where the ridge inverse
\begin{equation}\label{Generelized_inv_def1} 
\mathbf{A}_{{}}^{\dagger }={{(\mathbf{A}_{{}}^{T}\mathbf{A}+\lambda \mathbf{I})}^{-1}}\mathbf{A}_{{}}^{T}.
\end{equation}

In (\ref{Generelized_inv_def1}), the ridge parameter $\lambda$ can be any positive real number.
When  $\lambda \to 0$,
the
ridge solution
 (\ref{W2AY9845729}) degenerates into the generalized
inverse solution
(\ref{W2AinvY989565}),
 and the ridge inverse  degenerates into the
 generalized inverse~\cite[equation (3)]{BL_trans_paper},  i.e.,
 \begin{equation}\label{AinvLimNumda0AAiA1221}
\underset{\lambda \to 0}{\mathop{\lim }}\,\mathbf{A}^{\dagger}=\underset{\lambda \to 0}{\mathop{\lim }}\,{{(\mathbf{A}_{{}}^{T}\mathbf{A}+\lambda \mathbf{I})}^{-1}}\mathbf{A}_{{}}^{T}
=\mathbf{A}_{{}}^{+ }.
\end{equation}
In \cite{BL_trans_paper},
the ridge regression approximation of the generalized inverse, i.e.,
 (\ref{AinvLimNumda0AAiA1221}),
 is applied to compute the generalized inverse.


 \subsection{Broad Learning Model on Added Nodes}

In the BLS,
 the input data ${{\mathbf{X}}}$  is mapped to form the feature nodes,
 and then the feature nodes are  enhanced as the enhancement
nodes.  At last,  the connections of all the feature and enhancement nodes
 are fed into the output.

%
%

 The BLS  projects the input data $\mathbf{X}$
 to obtain the $i$-th group of mapped features ${{\mathbf{\ddot Z}}_{i}}$
 by
\begin{equation}\label{Z2PhyXWb985498}
{{\mathbf{\ddot Z}}_{i}}=\phi (\mathbf{X}{{\mathbf{W}}_{{{e}_{i}}}}+{{\mathbf{\beta }}_{{{e}_{i}}}}),
\end{equation}
where the  weights ${{\mathbf{W}}_{{{e}_{i}}}}$ and the biases ${{\mathbf{\beta }}_{{{e}_{i}}}}$ are
   generated randomly and then fine-tuned slightly
   with the linear inverse problem~\cite{BL_trans_paper}.
   All the
    $n$ groups of  features nodes can be concatenated into
\begin{equation}\label{Zi2z1zi988689}{{\mathbf{\ddot Z}}^{n}}\equiv \left[ \begin{matrix}
   {{\mathbf{\ddot Z}}_{1}} & \cdots  & {{\mathbf{\ddot Z}}_n}  \\
\end{matrix} \right],
\end{equation}
 which are enhanced
 to obtain  the
$j$-th group of enhancement nodes ${{\mathbf{\ddot H}}_{j}}$
  by
\begin{equation}\label{HjipsenZjWbelta09885}{{\mathbf{\ddot H}}_{j}}=\xi ({{\mathbf{\ddot Z}}^{n}}{{\mathbf{W}}_{{{h}_{j}}}}+{{\mathbf{\beta }}_{{{h}_{j}}}}),
\end{equation}
 where
    ${{\mathbf{W}}_{{{h}_{j}}}}$ and
    ${{\mathbf{\beta }}_{{{h}_{j}}}}$  are random.
 Then
 all the $m$ groups of enhancement nodes can also be concatenated into
\begin{equation}\label{Hj2H1Hj9859348}{{\mathbf{\ddot H}}^{m}}\equiv \left[ {{\mathbf{\ddot H}}_{1}},\cdots ,{{\mathbf{\ddot H}}_{m}} \right].
\end{equation}


All the feature and enhancement nodes
 can be denoted
  as the expanded input matrix
${{\mathbf{\ddot A}}_n^{m}}=\left[ {{\mathbf{\ddot Z}}^{n}}|{{\mathbf{\ddot H}}^{m}} \right]$,
where  the subscript $n$  and the superscript $m$ of ${{\mathbf{\ddot A}}_n^{m}}$ represent
$n$ groups of feature nodes and
$m$ groups of enhancement nodes,
  respectively.
Finally,
the connections of all the feature and enhancement nodes
 are fed into the output by
\begin{equation}\label{Y2ZiHjWj948934}
\mathbf{\hat{Y}}=\left[ {{\mathbf{\ddot Z}}^{n}}|{{\mathbf{\ddot H}}^{m}} \right]{{\mathbf{\ddot W}}_n^{m}}={{\mathbf{\ddot A}}_n^{m}}{{\mathbf{\ddot W}}_n^{m}},
\end{equation}
where
 the desired connection weights ${{\mathbf{\ddot W}}_n^{m}}$
are computed by (\ref{W2AinvY989565})
 from $({{\mathbf{\ddot A}}_n^{m}})^{+}$,
the pseudoinverse
with the ridge regression.


It can be seen that
all the $n$ groups of feature nodes are enhanced by (\ref{HjipsenZjWbelta09885}) synchronously, to obtain the $j$-th group of enhancement nodes ${{\mathbf{\ddot H}}_{j}}$.
In \cite{BL_trans_paper}, a different construction is also proposed, which connects each group of feature nodes to a group of enhancement nodes. That construction with $n$ groups of feature and enhancement nodes
 can be denoted as
\begin{equation*}\label{}\mathbf{\hat{Y}}=\left[ {{\mathbf{\ddot Z}}_{1}},\xi ({{\mathbf{\ddot Z}}_{1}}{{\mathbf{W}}_{{{h}_{1}}}}+{{\mathbf{\beta }}_{{{h}_{1}}}})|\cdots {{\mathbf{\ddot Z}}_n},\xi ({{\mathbf{\ddot Z}}_n}{{\mathbf{W}}_{{{h}_k}}}+{{\mathbf{\beta }}_{{{h}_k}}}) \right]{{\mathbf{\ddot W}}_n^{n}}.
\end{equation*}

\begin{algorithm}
\caption{:~\bf The Broad Learning Algorithm:  Computation of Output Weights and Increment of  Feature Nodes and Enhancement Nodes}
\begin{algorithmic}[1]
\Require Training samples $\mathbf{X}$ and labels ${{\mathbf{Y}}}$
\Ensure Output weights $\mathbf{W}$
\For{$i=1:n$}
\State Fine-tune random ${{\mathbf{W}}_{{{e}_{i}}}}$ and  ${{\bm{\beta }}_{{{e}_{i}}}}$;
\State Compute ${{\mathbf{\ddot Z}}_{i}}=\phi (\mathbf{X}{{\mathbf{W}}_{{{e}_{i}}}}+{{\bm{\beta }}_{{{e}_{i}}}})$;
\EndFor
\State Concatenate the feature nodes into ${{\mathbf{\ddot Z}}^{n}}\equiv \left[ \begin{matrix}
   {{\mathbf{\ddot Z}}_{1}} & \cdots  & {{\mathbf{\ddot Z}}_n}  \\
\end{matrix} \right]$;
\For{$j=1:m$}
\State Random   ${{\mathbf{W}}_{{{h}_{j}}}}$ and
    ${{\bm{\beta }}_{{{h}_{j}}}}$;
\State Compute ${{\mathbf{\ddot H}}_{j}}=\xi ({{\mathbf{\ddot Z}}^{n}}{{\mathbf{W}}_{{{h}_{j}}}}+{{\bm{\beta }}_{{{h}_{j}}}})$;
\EndFor
\State   Set the enhancement nodes group 
${{\mathbf{\ddot H}}^{m}}\equiv \left[ {{\mathbf{\ddot H}}_{1}},\cdots,{{\mathbf{\ddot H}}_{m}} \right]$;
\State Set ${{\mathbf{\ddot A}}_n^{m}}=\left[ {{\mathbf{\ddot Z}}^{n}}|{{\mathbf{\ddot H}}^{m}} \right]$;
\State Compute the intermediate result and  $\mathbf{\ddot W}_n^{m}$;  
\While{\emph{The target training error  is not reached}}
\If{\emph{only enhancement nodes are added}}
\State Random   ${{\mathbf{W}}_{{{h}_{m+1}}}}$ and
    ${{\bm{\beta }}_{{{h}_{m+1}}}}$;
\State Compute ${{\mathbf{\ddot H}}_{m+1}}=\xi ({{\mathbf{\ddot Z}}^{n}}{{\mathbf{W}}_{{{h}_{m+1}}}}+{{\bm{\beta }}_{{{h}_{m+1}}}})$;
\State Set ${\mathbf{\ddot A}}_n^{m+1}=\left[ {{\mathbf{\ddot A}}_n^{m}}|{{\mathbf{\ddot H}}_{m+1}}\right]$;
\State  Update the intermediate result, and update $\mathbf{\ddot W}_n^{m}$
\Statex  \quad \qquad  into $\mathbf{\ddot W}_n^{m+1}$;
\State $m=m+1$;
\Else[\emph{feature nodes are added}]
\State Fine-tune random ${{\mathbf{W}}_{{{e}_{n+1}}}}$ and  ${{\bm{\beta }}_{{{e}_{n+1}}}}$;
\State Compute ${{\mathbf{\ddot Z}}_{n+1}}=\phi (\mathbf{X}{{\mathbf{W}}_{{{e}_{n+1}}}}+{{\bm{\beta }}_{{{e}_{n+1}}}})$;
\State Set ${{\mathbf{\ddot Z}}^{n+1}}=\left[ {{\mathbf{\ddot Z}}^{n}}| {{\mathbf{\ddot Z}}_{n+1}} \right]$;
\For{$j=1:m$}
\State Random   ${{\mathbf{W}}_{e{{x}_{j}}}}$ and ${{\bm{\beta }}_{e{{x}_{j}}}}$;
\State Compute ${{\mathbf{\ddot H}}_{j,ex}}= \xi ({{\mathbf{\ddot Z}}_{n+1}}{{\mathbf{W}}_{e{{x}_{j}}}}+{{\bm{\beta }}_{e{{x}_{j}}}})$;
\EndFor
\State Set ${{\mathbf{\ddot H}}_{e{{x}_{m}}}}= \left[{{\mathbf{\ddot H}}_{1,ex}},\cdots ,{{\mathbf{\ddot H}}_{m,ex}}\right]$;
\State Set  $\mathbf{\ddot A}_{n+1}^{m}=\left[ \mathbf{\ddot A}_{n}^{m}| {{\mathbf{\ddot Z}}_{n+1}}|{{\mathbf{\ddot H}}_{e{{x}_{m}}}} \right]$;
\State  Update the intermediate result, and update $\mathbf{\ddot W}_n^{m}$
\Statex  \quad \qquad  into $\mathbf{\ddot W}_{n+1}^{m}$;
\State   $n=n+1$;
\EndIf
\EndWhile
\State $\mathbf{W}=\mathbf{\ddot W}_{n}^{m}$;
\end{algorithmic}
\end{algorithm}

\subsection{Incremental Learning for the Original BLS (i.e., Orig.)}

In this subsection, we introduce the incremental learning algorithm for
the original BLS~\cite{BL_trans_paper},
which is abbreviated as  \textbf{Orig.} in this paper.
When the
 model cannot reach the desired
accuracy,
 additional nodes can be inserted
 to achieve a better performance, and
 the incremental learning  algorithm for the original BLS can remodel
 the system  in an incremental way
 without retraining
  the whole network.
 In BLS, the inserted nodes can be some
 enhancement nodes, or some feature nodes with the corresponding
 enhancement nodes.




 When some enhancement nodes are inserted,
 the inserted nodes can be
represented
   as the
$(m+1)$-th group of enhancement nodes, i.e., ${{\mathbf{\ddot H}}_{m+1}}$ defined by (\ref{HjipsenZjWbelta09885}).
Accordingly,
    the expanded input matrix  ${{\mathbf{\ddot A}}_n^{m}}$
is updated into
\begin{equation}\label{Abig_original}
{\mathbf{\ddot A}}_n^{m+1}=\left[ {{\mathbf{\ddot A}}_n^{m}}|\mathbf{\ddot H}_{m+1}^{{}} \right].
\end{equation}
Moreover, when some feature nodes and the corresponding enhancement nodes are inserted,
the added feature nodes can be denoted as
  the $(n+1)$-th group of  feature nodes, i.e.,
     ${{\mathbf{\ddot Z}}_{n+1}}$
defined by (\ref{Z2PhyXWb985498}), and then
 ${{\mathbf{\ddot Z}}_{n+1}}$
is
  substituted
  into  (\ref{HjipsenZjWbelta09885}) to get
 the added  enhancement nodes  in  the $j$-th group (corresponding to  ${{\mathbf{\ddot Z}}_{n+1}}$),
 i.e.,
${{\mathbf{\ddot H}}_{j,ex}}=\xi \left({{\mathbf{\ddot Z}}_{n+1}}{{\mathbf{W}}_{e{{x}_{j}}}}+{{\bm{\beta }}_{e{{x}_{j}}}}\right)$.
The  $m$ groups of added enhancement nodes
is concatenated into
${{\mathbf{\ddot H}}_{e{{x}_{m}}}}=
\left[{{\mathbf{\ddot H}}_{1,ex}}, {{\mathbf{\ddot H}}_{2,ex}}, \cdots , {{\mathbf{\ddot H}}_{m,ex}} \right]$
  by  (\ref{Hj2H1Hj9859348}).
Then   ${{\mathbf{\ddot Z}}_{n+1}}$
and  ${{\mathbf{\ddot H}}_{e{{x}_{m}}}}$  are
applied to update
the
expanded
input
matrix  ${{\mathbf{\ddot A}}_n^{m}}$
 into~\cite{BL_trans_paper}
\begin{equation}\label{Anm2AnmZH120sd}
\mathbf{\ddot A}_{n+1}^{m}=\left[ \mathbf{\ddot A}_{n}^{m}|{{\mathbf{\ddot Z}}_{n+1}}|{{\mathbf{\ddot H}}_{e{{x}_{m}}}} \right].
\end{equation}


Assume  $l$ training samples  and $k$ nodes in
the expanded input
matrix
${{\mathbf{\ddot A}}_n^{m}}$.
Then
    ${{\mathbf{\ddot A}}_n^{m}} \in {\Re ^{l \times k}} $
 can be written
  as 
\begin{equation}\label{}
{{\mathbf{ A}}_k} = {{\mathbf{\ddot A}}_n^{m}}=\left[ {{\mathbf{\ddot Z}}^{n}}|{{\mathbf{\ddot H}}^{m}} \right],
 \end{equation}
  where the subscript $k$  in ${{\mathbf{ A}}_k}$ denotes the number of columns.
  We can also write both ${\mathbf{\ddot A}}_n^{m+1}$ and $\mathbf{\ddot A}_{n+1}^{m}$
 as $\mathbf{A}_{k+q}^{{}}$ with $k+q$ columns,
 to   unify
   (\ref{Abig_original}) and (\ref{Anm2AnmZH120sd})
 into
 \begin{equation}\label{Anp2AnH954734}
\mathbf{ A}_{k+q}^{{}}=\left[ \mathbf{ A}_{k}^{{}}|\mathbf{ H} \right],
 \end{equation}
 where  $\mathbf{ H}$  with  $q$ columns is defined by
 \begin{equation}\label{}
\mathbf{ H}=\mathbf{\ddot H}_{m+1}
 \end{equation}
 for  (\ref{Abig_original}),
or by
 \begin{equation}\label{}
\mathbf{ H}=\left[{{\mathbf{\ddot Z}}_{n+1}}|{{\mathbf{\ddot H}}_{e{{x}_{m}}}} \right]
 \end{equation}
 for (\ref{Anm2AnmZH120sd}).



The incremental learning algorithm  for BLS in \cite{BL_trans_paper} applies
the Greville's method~\cite{cite_general_inv_book}
  to
compute
the generalized inverse of the column-partitioned matrix $\mathbf{ A}_{k+q} =\left[ {\mathbf{ A}_k}|{{\bf{ H}}} \right]$,
 by
 \begin{equation}\label{A_1col_inv_book1matrixOrig}
 {\mathbf{ A}_{k+q}^+}={{\left[ {\mathbf{ A}_k}|{{\bf{ H}}} \right]}^{+}}=\left[ \begin{matrix}
   {{\bf{ A}}_k^+}-\mathbf{\bar D}{{\mathbf{\bar B}}^{T}}  \\
   {{\mathbf{\bar B}}^{T}}  \\
\end{matrix} \right]
\end{equation}
where the newly added sub-matrix~\footnote{In \cite{BL_trans_paper}, the generalized inverse (\ref{AinvLimNumda0AAiA1221}) is utilized to compute  ${{\bf{\bar C}}^ + }$ in (\ref{B_Matrix_def1aOrig}), where
$\underset{\lambda \to 0}{\mathop{\lim }}\,{{(\mathbf{C}_{{}}^{T}\mathbf{C}+\lambda \mathbf{I})}^{-1}}\mathbf{C}_{{}}^{T}$ is written as ${{{{\left({{\mathbf{\bar C}_{{}}^{T}\mathbf{\bar C}}} \right)}^{-1}}{{\mathbf{\bar C}}^{T}}}}$ for simplicity.}
\begin{subequations}{\label{B_Matrix_def1abOrig}}
 \begin{numcases}
{  {{\bf{\bar B}}^T} = }
{ {{\bf{\bar C}}^ + } ={{{{\left({{\mathbf{\bar C}_{{}}^{T}\mathbf{\bar C}}} \right)}^{-1}}{{\mathbf{\bar C}}^{T}}}}  \; \; \; \quad \quad  if \ {\bf{\bar C}} \ne \mathbf{0} }   &  \label{B_Matrix_def1aOrig} \\
{{{({\bf{I}} + {{\bf{\bar D}}^T}{\bf{\bar D}})}^{ - 1}}{{\bf{\bar D}}^T} {{\bf{ A}}_k^+} \quad \quad \ if \ {\bf{\bar C}} = \mathbf{0} },  &  \label{B_Matrix_def1bOrig}
\end{numcases}
\end{subequations}
\begin{equation}\label{C_matrix_def_111Orig}
\mathbf{\bar C}={{\bf{H}}}-{\mathbf{ A}_k}\mathbf{\bar D},
\end{equation}
and
\begin{equation}\label{D_matrix_def_111Orig}
\mathbf{\bar D} = {{\bf{ A}}_k^+}{{\bf{H}}}\text{ }.
\end{equation}
   After  the generalized inverse  ${{\bf{ A}}_k^+}$ is  updated  into ${\mathbf{ A}_{k+q}^+}$  by (\ref{A_1col_inv_book1matrixOrig}),
 the sub-matrix ${{\mathbf{\bar B}}^{T}} $ in ${\mathbf{ A}_{k+q}^+}$
 is utilized to update
 the generalized inverse solution ${{\bf{\bar W}}_k}$ into ${{\bf{\bar W}}_{k+q}}$
by
\begin{equation}\label{W2WDBB94398Orig}
{{\bf{\bar W}}_{k+q}}{{ = }}\left[ {\begin{array}{*{20}{c}}
{{{\bf{\bar W}}_k} - {\bf{\bar D}}{{\bf{\bar B}}^T}{\bf{Y}}}\\
{{{\bf{\bar B}}^T}{\bf{Y}}}
\end{array}} \right],
 \end{equation}
 which forms the output weights.

\begin{algorithm}[htb]
\caption{:~\bf The Existing Generalized Inverse Solution for
 BLS (i.e., \textbf{Orig.}):  Initialization and Update of Generalized Inverse and Output Weights}
\begin{algorithmic}
\State \quad \quad  $\cdots \cdots$ 
\State {\footnotesize 12:}  Compute   $({{\mathbf{\ddot A}}_n^{m}})^{+}= \underset{\lambda \to 0}{\mathop{\lim }}\,{{\left( ({{\mathbf{\ddot A}}_n^{m}})^{T}{{\mathbf{\ddot A}}_n^{m}} +\lambda \mathbf{I} \right)}^{-1}} ({{\mathbf{\ddot A}}_n^{m}})^{T}$
\State \quad \;  and  $\mathbf{\ddot W}_n^{m} =({{\mathbf{\ddot A}}_n^{m}})^{+}\mathbf{Y}$;
\State \quad \quad  $\cdots \cdots$
\State {\footnotesize 18:}  Compute $({\mathbf{\ddot A}}_n^{m+1})^+$ and $\mathbf{\ddot W}_n^{m+1}$  by
\State \quad \;  ${\psi _1}\left({({{\mathbf{\ddot A}}_n^{m}})^+}, {{\mathbf{\ddot A}}_n^{m}}, \mathbf{\ddot H}_{m+1},  \mathbf{\ddot W}_n^{m}, {\mathbf{Y}}\right)$;
\State \quad \quad  $\cdots \cdots$
\State {\footnotesize 30:}  Compute $(\mathbf{\ddot A}_{n+1}^{m})^+$ and $\mathbf{\ddot W}_{n+1}^{m}$  by
\State \quad \;  ${\psi _1}\left({({{\mathbf{\ddot A}}_n^{m}})^+}, {{\mathbf{\ddot A}}_n^{m}}, \left[{{\mathbf{\ddot Z}}_{n+1}}|{{\mathbf{\ddot H}}_{e{{x}_{m}}}} \right],  \mathbf{\ddot W}_n^{m}, {\mathbf{Y}}\right)$;
\State \quad \quad  $\cdots \cdots$
\end{algorithmic}
\end{algorithm}

\begin{algorithm}
\caption{The Algorithm to Update ${{\mathbf{A}}_{k}^+}$ and ${{\mathbf{\bar W}}_{k}}$ for \textbf{Orig.}}\label{euclid}
\begin{algorithmic}[0]
\Function{${\psi _1}$}{${{\mathbf{A}}_k^+}$, ${{\mathbf{A}}_k}$, ${{\mathbf{H}}}$,  ${{\mathbf{\bar W}}_k}$, ${\mathbf{Y}}$}
\State $\mathbf{\bar D} = {{\mathbf{A}}_k^+}{{\mathbf{H}}}$
\State $\mathbf{\bar C}={{\mathbf{H}}}-{{\mathbf{A}}_k}\mathbf{\bar D}$
\State ${{\mathbf{\bar B}}^T} =\underset{\lambda \to 0}{\mathop{\lim }}\,{(\mathbf{\bar C}_{{}}^{T}\mathbf{\bar C}+\lambda \mathbf{I})^{-1}}\mathbf{\bar C}_{{}}^{T}$
\State $ {{\mathbf{A}}_{k + q}^+}=\left[ \begin{matrix}
   {{\mathbf{A}}_k^+}-\mathbf{\bar D}{{\mathbf{\bar B}}^{T}}  \\
   {{\mathbf{\bar B}}^{T}}  \\
\end{matrix} \right]$
\State ${{\mathbf{\bar W}}_{k+q}}{{ = }}\left[ {\begin{array}{*{20}{c}}
{{{\mathbf{\bar W}}_k} - {\mathbf{\bar D}}{{\mathbf{\bar B}}^T}{\mathbf{Y}}}\\
{{{\mathbf{\bar B}}^T}{\mathbf{Y}}}
\end{array}} \right]$
\State \textbf{return} ${{\mathbf{A}}_{k + q}^+}$, ${{\mathbf{\bar W}}_{k+q}}$
\EndFunction
\end{algorithmic}
\end{algorithm}

\subsection{The Existing Efficient Generalized Inverse Solution for BLS by Inverse Cholesky Factorization (i.e., Exg-E)}

The existing efficient BLS proposed in \cite{BLSpaper2021zhf1}
accelerates the original BLS~\cite{BL_trans_paper} for added nodes,
which is abbreviated as \textbf{Exg-E} in this paper.
 It computes the output weights ${{\bf{\bar W}}}$
   from the inverse Cholesky
factor of the Hermitian matrix
\begin{equation}\label{R_define12321numda}
 {{\mathbf{\bar R}}_k}=\underset{\lambda \to 0}{\mathop{\lim }}\,{({{\mathbf{ A}}_k^T}{{\mathbf{ A}}_k}+ \lambda {\mathbf{I}})}
  \end{equation}
 in
the pseudoinverse (\ref{AinvLimNumda0AAiA1221}), and updates the inverse Cholesky
factor efficiently.
Write
   the inverse Cholesky factor of ${{\mathbf{\bar R}}_k}$ as
 \begin{equation}\label{L_m_def12431before}
{{\mathbf{\bar F}}_k}{{\mathbf{\bar F}}_k^{T}}={{\mathbf{\bar R}}_k^{-1}}=
\underset{\lambda \to 0}{\mathop{\lim }}\,{({{\mathbf{ A}}_k^T}{{\mathbf{ A}}_k}+ \lambda {\mathbf{I}})^{-1}}.
\end{equation}
Then we can substitute (\ref{L_m_def12431before}) into
  (\ref{AinvLimNumda0AAiA1221})
  to
denote ${{\mathbf{ A}}_k^{+}}$  as
 \begin{equation}\label{A_gen_inv_def_by_L_112pseudoInv}
{{\mathbf{ A}}_k^{+}}={{\mathbf{\bar F}}_k}{{\mathbf{\bar F}}_k^{T}}{{\mathbf{ A}}_k^T},
\end{equation}
which is substituted
  into (\ref{W2AinvY989565}) to compute
 $\mathbf{\bar W}_k$
 by
\begin{equation}\label{W2AinvY989565forwk322Jun17fdef}
\mathbf{\bar W}_k= {{\mathbf{\bar F}}_k}{{\mathbf{\bar F}}_k^{T}}{{\mathbf{A}}_k^T}\mathbf{Y}.
\end{equation}


 In \cite{BLSpaper2021zhf1},
  $\mathbf{\bar C}$  is computed by
      \begin{equation}\label{C2HAFFAH03290kdf43fd}
{\mathbf{\bar C}} = {\mathbf{ H}}-{{\mathbf{ A}}_k}{{\mathbf{\bar F}}_k}{{\mathbf{\bar F}}_k^{T}}{{\mathbf{ A}}_k^T}{{\mathbf{ H}}}.
 \end{equation}
 Let us substitute (\ref{A_gen_inv_def_by_L_112pseudoInv})
into (\ref{D_matrix_def_111Orig}) to obtain
 \begin{equation}\label{C_def121numda2GetD}  
{\mathbf{\bar D}} = {{\mathbf{\bar F}}_k}{{\mathbf{\bar F}}_k^{T}}{{\mathbf{ A}}_k^T}{{\mathbf{ H}}},
\end{equation}
and then we can write  (\ref{C2HAFFAH03290kdf43fd})  as  (\ref{C_matrix_def_111Orig}) with $\mathbf{\bar D}$ defined by (\ref{C_def121numda2GetD}).
Finally, $\mathbf{\bar C}$ is applied  to
 update ${{\mathbf{\bar F}}_k}$ into
${{\mathbf{\bar F}}_{k+q}}$
  by
\begin{equation}\label{L_big_BLK_def1BarBar}{{\mathbf{\bar F}}_{k+q}}=\left[ \begin{matrix}
   {{\mathbf{\bar F}}_k} & \mathbf{\bar T}  \\
   \mathbf{0} & \mathbf{\bar G}  \\
\end{matrix} \right]
\end{equation}
with~\footnote{In (\ref{ZF_def_L_2_items3a}), $\mathbf{\bar G}$ denotes the upper-triangular inverse Cholesky factor of $\underset{\lambda \to 0}{\mathop{\lim }}\,{(\mathbf{\bar C}_{{}}^{T}\mathbf{\bar C}+\lambda \mathbf{I})}$.}
\begin{subnumcases}{\label{ZF_def_L_2_items3ab}}
 \mathbf{\bar G}{{\mathbf{\bar G}}^{T}}= \underset{\lambda \to 0}{\mathop{\lim }}\,{(\mathbf{\bar C}_{{}}^{T}\mathbf{\bar C}+\lambda \mathbf{I})^{-1}} &  \label{ZF_def_L_2_items3a}\\
\mathbf{\bar T}=-{\mathbf{\bar D}} \mathbf{\bar G},  & \label{ZF_def_L_2_items3b}
\end{subnumcases}
while $\mathbf{\bar T}$ and $\mathbf{\bar G}$ in (\ref{L_big_BLK_def1BarBar})
  are applied to update ${{\mathbf{\bar W}}_k}$ into
\begin{equation}\label{W2AMHEWNH3495pseudoInv329ak}{ {\mathbf{\bar W}}_{k+q} }=\left[ \begin{matrix}
   {{\mathbf{\bar W}}_k}+\mathbf{\bar T}{{\mathbf{\bar G}}^{T}}\left( {\mathbf{ H}}^{T}\mathbf{Y}-{{\mathbf{ H}}^T}{{\mathbf{ A}}_k}{{\mathbf{\bar W}}_k} \right)  \\
   \mathbf{\bar G}{{\mathbf{\bar G}}^{T}}\left( {\mathbf{ H}}^{T}\mathbf{Y}-{{\mathbf{ H}}^T}{{\mathbf{ A}}_k}{{\mathbf{\bar W}}_k} \right)  \\
\end{matrix} \right].
\end{equation}

\begin{algorithm}[htb]
\caption{:~\bf The Existing Efficient Generalized Inverse Solution for
 BLS (i.e., \textbf{Exg-E}):  Initialization and Update of  Inverse Cholesky Factor and Output Weights}
\begin{algorithmic}
\State \quad \quad  $\cdots \cdots$ 
\State {\footnotesize 12:}  Compute the upper-triangular ${{\mathbf{\ddot F}}_n^{m}}$  satisfying
\State \quad \;  ${{\mathbf{\ddot F}}_n^{m}}({{\mathbf{\ddot F}}_n^{m}})^{T}= \underset{\lambda \to 0}{\mathop{\lim }}\,{{\left( ({{\mathbf{\ddot A}}_n^{m}})^{T}{{\mathbf{\ddot A}}_n^{m}} +\lambda \mathbf{I} \right)}^{-1}}$,
\State \quad \;  and compute $\mathbf{\ddot W}_n^{m} ={{\mathbf{\ddot F}}_n^{m}}({{\mathbf{\ddot F}}_n^{m}})^{T} ({{\mathbf{\ddot A}}_n^{m}})^{T} \mathbf{Y}$;
\State \quad \quad  $\cdots \cdots$
\State {\footnotesize 18:}  Compute ${\mathbf{\ddot F}}_n^{m+1}$ and $\mathbf{\ddot W}_n^{m+1}$  by
\State \quad \;  ${\psi _2}\left({{{\mathbf{\ddot F}}_n^{m}}}, {{\mathbf{\ddot A}}_n^{m}}, \mathbf{\ddot H}_{m+1},  \mathbf{\ddot W}_n^{m}, {\mathbf{Y}}\right)$;
\State \quad \quad  $\cdots \cdots$
\State {\footnotesize 30:}  Compute $\mathbf{\ddot F}_{n+1}^{m}$ and $\mathbf{\ddot W}_{n+1}^{m}$  by
\State \quad \;  ${\psi _2}\left({{{\mathbf{\ddot F}}_n^{m}}}, {{\mathbf{\ddot A}}_n^{m}}, \left[{{\mathbf{\ddot Z}}_{n+1}}|{{\mathbf{\ddot H}}_{e{{x}_{m}}}} \right],  \mathbf{\ddot W}_n^{m}, {\mathbf{Y}}\right)$;
\State \quad \quad  $\cdots \cdots$
\end{algorithmic}
\end{algorithm}

\begin{algorithm}
\caption{The Algorithm to Update ${{\mathbf{\bar F}}_{k}}$ and ${{\mathbf{\bar W}}_{k}}$ for \textbf{Exg-E}}\label{euclid}
\begin{algorithmic}[0]
\Function{${\psi _2}$}{${{\mathbf{\bar F}}_k}$, ${{\mathbf{A}}_k}$, ${{\mathbf{H}}}$,  ${{\mathbf{\bar W}}_k}$, ${\mathbf{Y}}$}
\State ${\mathbf{\bar D}} = {{\mathbf{\bar F}}_k}{{\mathbf{\bar F}}_k^{T}}{{\mathbf{ A}}_k^T}{{\mathbf{ H}}}$
\State $\mathbf{\bar C}={{\bf{H}}}-{\mathbf{ A}_k}\mathbf{\bar D}$
\State Compute  
 $\mathbf{\bar G}$ satisfying $\mathbf{\bar G}{{\mathbf{\bar G}}^{T}}= \underset{\lambda \to 0}{\mathop{\lim }}\,{(\mathbf{\bar C}_{{}}^{T}\mathbf{\bar C}+\lambda \mathbf{I})^{-1}}$
\State   $ \mathbf{\bar T}=-{\mathbf{\bar D}} \mathbf{\bar G}$
\State ${{\mathbf{\bar F}}_{k+q}}=\left[ \begin{matrix}
   {{\mathbf{\bar F}}_k} & \mathbf{\bar T}  \\
   \mathbf{0} & \mathbf{\bar G}  \\
\end{matrix} \right]$
\State ${ {\mathbf{\bar W}}_{k+q} }=\left[ \begin{matrix}
   {{\mathbf{\bar W}}_k}+\mathbf{\bar T}{{\mathbf{\bar G}}^{T}}\left( {\mathbf{ H}}^{T}\mathbf{Y}-{{\mathbf{ H}}^T}{{\mathbf{ A}}_k}{{\mathbf{\bar W}}_k} \right)  \\
   \mathbf{\bar G}{{\mathbf{\bar G}}^{T}}\left( {\mathbf{ H}}^{T}\mathbf{Y}-{{\mathbf{ H}}^T}{{\mathbf{ A}}_k}{{\mathbf{\bar W}}_k} \right)  \\
\end{matrix} \right]$
\State \textbf{return} ${{\mathbf{\bar F}}_{k+q}}$, ${{\mathbf{\bar W}}_{k+q}}$
\EndFunction
\end{algorithmic}
\end{algorithm}

\subsection{Construction Model and Learning Procedure of BLS}

Both the existing BLS algorithms for the increment of nodes follow the same
construction model and learning procedure, which is summarized
in \textbf{Algorithm 1}.  
Different methods
 are utilized by the existing BLS algorithms
  to implement several steps in \textbf{Algorithm 1},
i.e.,
 the step to compute the intermediate result and $\mathbf{W}_n^{m}$ in row 12,
the step to update the intermediate result and $\mathbf{W}_n^{m+1}$  in row 18, and the step to
 update the intermediate result and $\mathbf{W}_{n+1}^{m}$ in row 30.


\textbf{Algorithms 2} and \textbf{4}
describe the  implementations
of the steps in rows 12, 18 and 30 of \textbf{Algorithm 1},
for  the original BLS  in \cite{BL_trans_paper} (i.e., \textbf{Orig.}) and
the existing efficient BLS   in \cite{BLSpaper2021zhf1} (i.e., \textbf{Exg-E}),
respectively.
The functions ${\psi _1}\left(\bullet \right)$ and
${\psi _2}\left(\bullet \right)$
utilized  in  \textbf{Algorithms 2} and \textbf{4}
are defined by \textbf{Algorithms 3} and \textbf{5},
respectively.
Notice that \textbf{Algorithm 2} includes (\ref{AinvLimNumda0AAiA1221})
 and
(\ref{W2AinvY989565}), \textbf{Algorithm 3} includes (\ref{A_1col_inv_book1matrixOrig})-(\ref{W2WDBB94398Orig}),
\textbf{Algorithm 4} includes (\ref{L_m_def12431before}) and
(\ref{W2AinvY989565forwk322Jun17fdef}), and \textbf{Algorithm 5} includes (\ref{C_matrix_def_111Orig}) and (\ref{C_def121numda2GetD})-(\ref{W2AMHEWNH3495pseudoInv329ak}).

\section{Proposed  Ridge Solution for BLS by Inverse Cholesky Factorization of a Partitioned Matrix}

 The original BLS algorithm in \cite{BL_trans_paper} is based on the Greville's method~\cite{cite_general_inv_book}, which can only compute the generalized inverse of a partitioned matrix.
 As shown in (\ref{AinvLimNumda0AAiA1221}), the ridge inverse can be viewed as an approximate generalized inverse~\cite{best_ridge_inv_paper213,BL_trans_paper}.
However, the ridge inverse is not the generalized inverse~\cite{best_ridge_inv_paper213}, e.g., the ridge inverse ${{\mathbf{A}}^{\dagger }}$ does not obey $\mathbf{A}{{\mathbf{A}}^{\dagger }}\mathbf{A}=\mathbf{A}$,
while the generalized inverse ${{\mathbf{A}}^{+}}$ obeys~\cite{cite_general_inv_book} $\mathbf{A}{{\mathbf{A}}^{+}}\mathbf{A}=\mathbf{A}$.
So
the Greville's method for the generalized inverse
 \cite{Before_general_inv_book,cite_general_inv_book} is usually inapplicable to the ridge inverse, but it can be applied to
 the ridge regression (\ref{AinvLimNumda0AAiA1221})
 in the original BLS,
  since
 the ridge parameter
 $\lambda$ is set to a very small positive real number~\cite{BL_trans_paper},
e.g.,
${{10}^{-8}}$,
and then
$\lambda \to 0$ can be assumed to assure that the ridge regression  (\ref{AinvLimNumda0AAiA1221})
is equal to  the generalized inverse.

The    BLS algorithm proposed in \cite{BLSpaper2021zhf1}  is only
an efficient implementation of
 the original BLS on added nodes,
 which is based on
 the inverse Cholesky factor of the
   Hermitian matrix in the ridge regression (\ref{AinvLimNumda0AAiA1221}).
Thus  the efficient BLS in \cite{BLSpaper2021zhf1}
also assumes the ridge parameter $\lambda \to 0$,  as  the original BLS.

For  the incremental BLS on added nodes,
 we will propose an efficient  ridge solution based on
the inverse Cholesky factor~\cite{my_inv_chol_paper} and its numerically more
stable implementation,
 which are
abbreviated as   \textbf{Chol-1}  and  \textbf{Chol-2},
respectively.
 It is
no longer required to assume the ridge parameter  $\lambda \to 0$
in the proposed efficient ridge solution,
which
   computes
 the
 output weights
  from
 the inverse Cholesky factor of the
   Hermitian matrix $\mathbf{A}^{T}\mathbf{A}  +\lambda \mathbf{I}$ in the ridge inverse (\ref{Generelized_inv_def1}),
  and updates the inverse Cholesky factor  efficiently by
 extending the  inverse Cholesky factorization in \cite{my_inv_chol_paper}.



\subsection{Efficient Inverse Cholesky Factorization of a Hermitian Matrix Partitioned into $2 \times 2$ Blocks for Proposed Chol-1}


In this subsection, we extend the
efficient inverse Cholesky factorization proposed in \cite{my_inv_chol_paper},
to develop an efficient inverse Cholesky factorization for a Hermitian matrix partitioned into $2 \times 2$ blocks, which will be utilized in
what follows.

In \cite{my_inv_chol_paper},
 the inverse Cholesky factor
  of a Hermitian matrix ${{\bf{R}}_k} \in {\Re ^{k \times k}}$
is the upper-triangular ${{\bf{F}}_k}$
that
 satisfies
\begin{equation}\label{L_m_def12431RidgeInvJun30}
{{\bf{F}}_k}{{\bf{F}}_k^{T}}={{\bf{R}}_k^{-1}}= ({{\bf{A}}_k^T}{{\bf{A}}_k}+ \lambda {\bf{I}})^{-1},
\end{equation}
where ${{\bf{ R}}_k}$ is defined by~\footnote{Notice that the ridge parameter $\lambda$ can be any positive real number
 for ${{\bf{ R}}_k}$ defined in (\ref{R_define12321numdaRidge302923}),
 while  $\lambda \to 0$ needs to be assumed for ${{\bf{\bar R}}_k}$ defined in
 (\ref{R_define12321numda}).}
\begin{equation}\label{R_define12321numdaRidge302923}
 {{\bf{ R}}_k}={{\bf{A}}_k^T}{{\bf{A}}_k}+ \lambda {\bf{I}}.
 \end{equation}
From (\ref{L_m_def12431RidgeInvJun30}),  we can deduce
\begin{equation}\label{L_m_def12431}
{\bf{F}}_k^{-T}{{\bf{F}}_k^{-1}}={{{{\bf{R}}_k}}},
\end{equation}
 which shows that the lower-triangular ${\bf{F}}_k^{-T}$ is the
 conventional
 Cholesky factor~\cite[Theorem 4.2.5]{Matrix_Computations_book} of ${{{{\bf{R}}_k}}}$.


 The
efficient inverse Cholesky factorization \cite{my_inv_chol_paper}
partitions
 a Hermitian matrix
  ${{\bf{R}}_{k+1}}$   into
 \begin{equation}\label{R_def_perhaps_No_R1zhfPaper}
 {{\bf{R}}_{k+1}}=\left[ \begin{matrix}
   {{\bf{R}}_{k}} & {\mathbf{p}}  \\
   \mathbf{p}^{T} & u  \\
\end{matrix} \right],
\end{equation}
and
updates
the inverse Cholesky factor of  ${{\bf{R}}_{k}}$ into that of ${{\bf{R}}_{k+1}}$
by
\begin{equation}\label{L_big_BLK_def1zhfPaper}{{\bf{F}}_{k+1}}=\left[ \begin{matrix}
   {{\bf{F}}_{k}} & \mathbf{t}  \\
   \mathbf{0} & g  \\
\end{matrix} \right]
\end{equation}
where
\begin{subnumcases}{\label{ZF_def_L_2_items3abzhfPaper}}
g = 1/\sqrt {u - {{\bf{p}}^T}{{\bf{F}}_{k}}{\bf{F}}_{k}^T{\bf{p}}}  &  \label{ZF_def_L_2_items3azhfPaper}\\
{\bf{t}} =  - g{{\bf{F}}_{k}}{\bf{F}}_{k}^T{\bf{p}}. & \label{ZF_def_L_2_items3bzhfPaper}
\end{subnumcases}
In (\ref{R_def_perhaps_No_R1zhfPaper})-(\ref{ZF_def_L_2_items3abzhfPaper}),
$u$  and $g$ are scalars, while ${\mathbf{p}}$ and ${\mathbf{t}}$
are column vectors of length $k$.

In the more general case, we can partition
a Hermitian matrix ${{\bf{ R}}_{k + q}} =  {{\bf{A}}_{k+q}^T}{{\bf{A}}_{k+q}}+ \lambda {\bf{I}} =
{\left[ \mathbf{ A}_{k}^{{}}|\mathbf{ H} \right]^T}\left[ \mathbf{ A}_{k}^{{}}|\mathbf{ H} \right] + \lambda {\bf{I}}$
into
 a $2 \times 2$ block Hermitian matrix, i.e.,
 \begin{equation}\label{R_def_perhaps_No_R1}{{\bf{R}}_{k+q}}=\left[ \begin{matrix}
   {{\bf{R}}_k} & {\mathbf{P}}  \\
   \mathbf{P}^{T} & {\mathbf{U}}  \\
\end{matrix} \right]=\left[ \begin{matrix}
   {{\bf{R}}_k} & {{\mathbf{ A}}_k^T}{{\mathbf{ H}}}  \\
   {{\mathbf{ H}}^T}{{\mathbf{ A}}_k} & {\bf{H}}^{T}{{\bf{H}}}+\lambda \mathbf{I}  \\
\end{matrix} \right],
\end{equation}
where $ \mathbf{U} = {\bf{H}}^{T}{{\bf{H}}}+\lambda \mathbf{I}   \in {\Re ^{q \times q}}$
and  $\mathbf{P}={{\mathbf{ A}}_k^T}{{\mathbf{ H}}}\in {\Re ^{k \times q}}$.
We can  apply (\ref{ZF_def_L_2_items3abzhfPaper}) and
(\ref{L_big_BLK_def1zhfPaper})
 to update the
inverse Cholesky factor of  ${{\bf{R}}_k}$
into that of ${{\bf{R}}_{k+q}}$
through $q$ iterations, to obtain
\begin{equation}\label{L_big_BLK_def1}{{\bf{F}}_{k+q}}=\left[ \begin{matrix}
   {{\bf{F}}_k} & \mathbf{T}  \\
   \mathbf{0} & \mathbf{G}  \\
\end{matrix} \right]
\end{equation}
where
 $\mathbf{T}\in {\Re ^{k \times q}}$ and
  $\mathbf{G} \in {\Re ^{q \times q}}$.
  However, we prefer to
update ${{\bf{F}}_k}$ into
${{\bf{F}}_{k+q}}$ through only $1$ iteration,
  by computing
\begin{subnumcases}{\label{ZF_def_L_2_items3abJun2021}}
{{\mathbf{G}}}{{\mathbf{G}}^{T}}=\left({{\bf{H}}^{T}{{\bf{H}}}+\lambda \mathbf{I}}- {{\mathbf{ H}}^T}{{\mathbf{ A}}_k}{{\bf{F}}_k}{{\bf{F}}_k^{T}}{{{\mathbf{ A}}_k^T}{{\mathbf{ H}}}}\right)^{-1} &  \label{ZF_def_L_2_items3aJun2021}\\
 \mathbf{T}=-{{\bf{F}}_k}{{\bf{F}}_k^{T}}{{{\mathbf{ A}}_k^T}{{\mathbf{ H}}}}\mathbf{G}, & \label{ZF_def_L_2_items3bJun2021}
\end{subnumcases}
which can be regarded as an extension of
 (\ref{ZF_def_L_2_items3abzhfPaper}),
and will be deduced in Appendix A.
Now we can
 update the inverse Cholesky factor of  ${{\bf{ R}}_{k}}$ into that of ${{\bf{ R}}_{k + q}}$ by
 the above-described block inverse Cholesky factorization,
  i.e.,
we can
  update ${{\bf{ F}}_k}$ into ${{\bf{ F}}_{k+q}}$
 by
 (\ref{ZF_def_L_2_items3abJun2021})
  and (\ref{L_big_BLK_def1}),
 and compute the initial ${{\bf{ F}}_{k}}$
  by (\ref{L_m_def12431RidgeInvJun30}).
In
(\ref{ZF_def_L_2_items3aJun2021}),
 the upper-triangular
 inverse Cholesky factor
  $\mathbf{G}$
  can be
computed
  by the inverse Cholesky factorization~\cite{my_inv_chol_paper}
  or
 by inverting and transposing the lower-triangular Cholesky
factor~\footnote{Similarly,  ${\bf{F}}_k$  can be obtained by inverting and transposing the Cholesky factor ${\bf{F}}_k^{-T}$, as shown in (\ref{L_m_def12431}).}.
\subsection{Utilizing the Inverse Cholesky Factor to Compute the Ridge Inverse and the Ridge Solution for Proposed Chol-1}

In this subsection, we will utilize the
  inverse Cholesky factor
to compute the ridge inverse and the ridge solution.



We can
compute
the ridge inverse
  ${{\bf{ A}}_k^\dag}$ from
    ${{\bf{F}}_{k}}$ by
 \begin{equation}\label{A_gen_inv_def_by_L_112}
{{\bf{ A}}_k^\dag}={{\bf{F}}_k}{{\bf{F}}_k^{T}}{{\bf{ A}}_k^T},
\end{equation}
which  is deduced by
 substituting (\ref{L_m_def12431RidgeInvJun30})
 into
 (\ref{Generelized_inv_def1}).
Then
to update ${{\bf{ A}}_k^\dag}$ into  ${\mathbf{ A}_{k+q}^\dag}$,
 let us
 substitute (\ref{Anp2AnH954734}) and (\ref{L_big_BLK_def1}) into (\ref{A_gen_inv_def_by_L_112}) to write
\begin{gather*}
{\mathbf{ A}_{k+q}^\dag}=\left[ \begin{matrix}
   {{\bf{F}}_k} & \mathbf{T}  \\
   \mathbf{0} & \mathbf{G}  \\
\end{matrix} \right] \left[ \begin{matrix}
   {{\bf{F}}_k} & \mathbf{T}  \\
   \mathbf{0} & \mathbf{G}  \\
\end{matrix} \right]^T \left[ {\begin{array}{*{20}{c}}
{{\bf{ A}}_k^{}}&{\bf{H}}
\end{array}} \right]^T \\
=\left[ \begin{matrix}
   {{\bf{F}}_k}{{\bf{F}}_k^{T}}{{\bf{ A}}_k^T}+\mathbf{T}{{\mathbf{T}}^{T}}{{\bf{ A}}_k^T}+\mathbf{T}{{\mathbf{G}}^{T}}{\bf{H}}^{T}  \\
   \mathbf{G}{{\mathbf{T}}^{T}}{{\bf{ A}}_k^T}+\mathbf{G}{{\mathbf{G}}^{T}}{\bf{H}}^{T}  \\
\end{matrix} \right],
\end{gather*}
into which we substitute  (\ref{A_gen_inv_def_by_L_112})
 and  (\ref{ZF_def_L_2_items3b}) to obtain
\begin{gather}
	{\mathbf{ A}_{k+q}^\dag}=
\left[ \begin{matrix}
   {{\bf{ A}}_k^\dag}+\mathbf{T}{(-{{\bf{F}}_k}{{\bf{F}}_k^{T}}{{{\mathbf{ A}}_k^T}{{\mathbf{ H}}}}\mathbf{G})^T}{{\bf{ A}}_k^T}+\mathbf{T}{{\mathbf{G}}^{T}}{\bf{H}}^{T}  \\
   \mathbf{G}{(-{{\bf{F}}_k}{{\bf{F}}_k^{T}}{{{\mathbf{ A}}_k^T}{{\mathbf{ H}}}}\mathbf{G})^T}{{\bf{ A}}_k^T}+\mathbf{G}{{\mathbf{G}}^{T}}{\bf{H}}^{T}  \\
\end{matrix} \right]  \notag \\
	=
\left[ {\begin{array}{*{20}{c}}
{\begin{array}{l}
{{\bf{ A}}_k^\dag } + {\mathbf{T}} {{\bf{G}}^T}
\left( {{\bf{H}}^T - {{\mathbf{ H}}^T}{{\mathbf{ A}}_k}{{\bf{F}}_k}{{\bf{F}}_k^T}{{\bf{ A}}_k^T}} \right)
\end{array}}\\
{{\bf{G}}{{\bf{G}}^T}\left( {{\bf{H}}^T - {{\mathbf{ H}}^T}{{\mathbf{ A}}_k}{{\bf{F}}_k}{{\bf{F}}_k^T}{{\bf{ A}}_k^T}} \right)}
\end{array}} \right]  \label{AbigByALLENN121} \\
=\left[ \begin{matrix}
   {{\bf{ A}}_k^\dag}+\mathbf{T}{{\mathbf{G}}^{T}}\left( {\bf{H}}^{T}-{{\mathbf{ H}}^T}{{\mathbf{ A}}_k}{{\bf{ A}}_k^\dag} \right)  \\
   \mathbf{G}{{\mathbf{G}}^{T}}\left( {\bf{H}}^{T}-{{\mathbf{ H}}^T}{{\mathbf{ A}}_k}{{\bf{ A}}_k^\dag} \right)  \\
\end{matrix} \right].   \label{AbigByALLENNEA084945}
\end{gather}
 Notice that  the sub-matrices $\mathbf{G}$ and
 $\mathbf{T}$  in ${{\bf{F}}_{k+q}}$  are utilized to update ${{\bf{A}}_{k}^\dag}$ into ${\mathbf{ A}_{k+q}^\dag}$ by (\ref{AbigByALLENNEA084945}).
We can also use ${{\bf{F}}_{k}}$
 to
compute the ridge solution ${{\bf{ W}}_{k}}$  by
\begin{equation}\label{W2AMHEWNH3495initial}
{ {\bf{ W}}_{k} }= {{\bf{F}}_k}{{\bf{F}}_k^{T}}\mathbf{A}_{{k}}^{T} {{\mathbf{Y}}},
\end{equation}
which is deduced
by substituting
 (\ref{A_gen_inv_def_by_L_112})
 into
 (\ref{W2AY9845729}).
To update ${{\bf{ W}}_{k}}$ into ${{\bf{ W}}_{k+q}}$,
substitute
  (\ref{AbigByALLENNEA084945})  into (\ref{W2AY9845729}) (i.e., $\mathbf{W} = {{{{\mathbf{A}}}}^{\dagger}}{{\mathbf{Y}}}$)  to write
  \begin{align}
{{\bf{ W}}_{k+q}} &=\left[ \begin{matrix}
   {{\bf{ A}}_k^\dag}\mathbf{Y}+\mathbf{T}{{\mathbf{G}}^{T}}\left( {\bf{H}}^{T}\mathbf{Y}-{{\mathbf{ H}}^T}{{\mathbf{ A}}_k}{{\bf{ A}}_k^\dag}\mathbf{Y} \right)  \\
   \mathbf{G}{{\mathbf{G}}^{T}}\left( {\bf{H}}^{T}\mathbf{Y}-{{\mathbf{ H}}^T}{{\mathbf{ A}}_k}{{\bf{ A}}_k^\dag}\mathbf{Y} \right)  \\
\end{matrix} \right]  \notag \\
 & =\left[ \begin{matrix}
   {{\bf{ W}}_k}+\mathbf{T}{{\mathbf{G}}^{T}}\left( {\bf{H}}^{T}\mathbf{Y}-{{\mathbf{ H}}^T}{{\mathbf{ A}}_k}{{\bf{ W}}_k} \right)  \\
   \mathbf{G}{{\mathbf{G}}^{T}}\left( {\bf{H}}^{T}\mathbf{Y}-{{\mathbf{ H}}^T}{{\mathbf{ A}}_k}{{\bf{ W}}_k} \right)  \\
\end{matrix} \right].  \label{W2AMHEWNH3495}
\end{align}
When   the ridge solution is computed
 by (\ref{W2AMHEWNH3495initial}) or (\ref{W2AMHEWNH3495}),
 the ridge inverse ${{\bf{A}}_{k}^\dag}$ or  ${\mathbf{ A}_{k+q}^\dag}$
is not required. Accordingly,
 there is actually no need
  to compute  the ridge inverse
from  the inverse Cholesky factor by (\ref{A_gen_inv_def_by_L_112}) or
(\ref{AbigByALLENNEA084945}).

\subsection{A Numerically More Stable Implementation to Update the Inverse Cholesky Factor and the Ridge Solution (i.e., Chol-2)}


Let us
define
 \begin{equation}\label{DLLtE1335645}
\mathbf{D}={{\bf{F}}_k}{{\bf{F}}_k^{T}}{{\mathbf{ A}}_k^T}{{\mathbf{ H}}}
\end{equation}
and
\begin{equation}\label{C_matrix_def_111}
\mathbf{C}={{\bf{H}}}-{\mathbf{ A}_k}\mathbf{D}={{\bf{H}}}-{\mathbf{ A}_k}{{\bf{F}}_k}{{\bf{F}}_k^{T}}{{\bf{ A}}_k^T}{{\bf{H}}},
\end{equation}
which are similar to (\ref{C_def121numda2GetD}) and (\ref{C_matrix_def_111Orig}),
respectively.
Then we can apply  (\ref{C_matrix_def_111})  to
write (\ref{ZF_def_L_2_items3aJun2021})
as
 \begin{align}
{{\mathbf{G}}}{{\mathbf{G}}^{T}}&= \left( {\bf{H}}^T( {{\bf{H}}}-{\mathbf{ A}_k}{{\bf{F}}_k}{{\bf{F}}_k^T}{{\bf{ A}}_k^T}{{\bf{H}}}) + \lambda {\bf{I}} \right)^{-1}  \notag \\
&=  \left( {\bf{H}}^T \mathbf{C}  + \lambda {\bf{I}} \right)^{-1}.  \label{GG2HC943039fd234}
\end{align}

 To  write (\ref{GG2HC943039fd234}) in another form,
let us
substitute
${{\bf{H}}}=\mathbf{C}+{\mathbf{ A}_k}\mathbf{D}$ (deduced from (\ref{C_matrix_def_111}))
into
 ${\bf{H}}^{T}\mathbf{ C}$ (in (\ref{GG2HC943039fd234}))
 to obtain
\begin{equation}\label{BLimHCIDACCt321}
{{{\bf{H}}^{T}\mathbf{ C}={{\mathbf{ C}_{{}}^{T}\mathbf{ C}}}+{{\mathbf{ D}}^{T}}{{\bf{ A}}_k^T}\mathbf{ C}}},
\end{equation}
where  ${{\mathbf{ D}}^{T}}{{\bf{ A}}_k^T}\mathbf{ C}$   satisfies
\begin{equation}\label{DACdifferent23412Equal999}
{{\mathbf{ D}}^{T}}{{\bf{ A}}_k^T}\mathbf{ C}=\lambda {{\mathbf{ D}}^{T}}  \mathbf{D},
\end{equation}
as will be verified in Appendix B.
Then we
 substitute  (\ref{DACdifferent23412Equal999}) into (\ref{BLimHCIDACCt321}) to obtain
 \begin{equation}\label{HC2CCDD920034jdfjer23kds43}
{\bf{H}}^{T}\mathbf{C}={{\mathbf{C}_{{}}^{T}\mathbf{C}}}+\lambda \mathbf{D}^{T}\mathbf{D},
\end{equation}
which is substituted
into
(\ref{GG2HC943039fd234})
to rewrite it as
\begin{equation}\label{GG2HC943039fd234moreStable}
{{\mathbf{G}}}{{\mathbf{G}}^{T}}= \left( {{\mathbf{C}_{{}}^{T}\mathbf{C}}}+\lambda \mathbf{D}^{T}\mathbf{D}  + \lambda {\bf{I}} \right)^{-1}.
\end{equation}
It can easily be seen that
 \begin{multline}\label{xCCDDlamdax2930dsk23}
{\mathbf{x}}^T({{\mathbf{C}_{{}}^{T}\mathbf{C}}}+\lambda \mathbf{D}^{T}\mathbf{D}  + \lambda {\bf{I}}){\mathbf{x}}= \\
(\mathbf{C}{\mathbf{x}})^{T} (\mathbf{C}{\mathbf{x}}) +\lambda (\mathbf{D}{\mathbf{x}})^{T}(\mathbf{D}{\mathbf{x}})  + \lambda {\mathbf{x}}^T{\mathbf{x}}  >0
 \end{multline}
is satisfied for any vector ${\mathbf{x}} \ne {\mathbf{0}}$. Thus
 ${{\mathbf{C}_{{}}^{T}\mathbf{C}}}+\lambda \mathbf{D}^{T}\mathbf{D}  + \lambda {\bf{I}}$ in (\ref{GG2HC943039fd234moreStable})
is positive definite,
which partly explains the fact that (\ref{GG2HC943039fd234moreStable})
is numerically more stable than (\ref{GG2HC943039fd234}) in the simulations when the ridge parameter $\lambda$ is very small.

$\mathbf{D}$ computed by
(\ref{DLLtE1335645})
can be utilized to simplify
(\ref{ZF_def_L_2_items3bJun2021})
into
   \begin{equation}\label{T2DG32902394309}
\mathbf{T}=-\mathbf{D}\mathbf{G}.
 \end{equation}
Moreover, we  utilize (\ref{W2AMHEWNH3495initial}) and (\ref{C_matrix_def_111})
in turn
to
 obtain
\begin{align}
{\bf{H}}^{T}\mathbf{Y}-{{\mathbf{ H}}^T}{{\mathbf{ A}}_k}{{\bf{ W}}_k}&= {\bf{H}}^{T}\mathbf{Y}-{{\mathbf{ H}}^T}{{\mathbf{ A}}_k}{{\bf{F}}_k}{{\bf{F}}_k^{T}}\mathbf{A}_{{k}}^{T} {{\mathbf{Y}}}  \notag \\
&= ({\bf{H}}^{T}-{{\mathbf{ H}}^T}{{\mathbf{ A}}_k}{{\bf{F}}_k}{{\bf{F}}_k^{T}}\mathbf{A}_{{k}}^{T}){{\mathbf{Y}}}  \notag \\
&= {\bf{C}}^{T} {{\mathbf{Y}}},  \label{HYHAWHY932032kds23}
\end{align}
which is substituted into
(\ref{W2AMHEWNH3495}) to write it  as
\begin{equation}\label{W2WTGCeiwo32dsk}
{{\bf{ W}}_{k+q}} =\left[ \begin{matrix}
   {{\bf{ W}}_k} + \mathbf{T} {{\mathbf{G}}^{T}} {\bf{C}}^{T} {{\mathbf{Y}}}  \\
   \mathbf{G}{{\mathbf{G}}^{T}}{\bf{C}}^{T} {{\mathbf{Y}}}  \\
\end{matrix} \right].
 \end{equation}

\subsection{Construction Model and Learning Procedure of Proposed Chol-1 and Chol-2}

The proposed efficient ridge solution (i.e., \textbf{Chol-1}) and its numerically more
stable implementation (i.e., \textbf{Chol-2})
 follow the construction model and learning procedure summarized
in \textbf{Algorithm 1}, and utilize
\textbf{Algorithm 6}
to implement the steps in rows 12, 18 and 30 of \textbf{Algorithm 1}  (which compute the output weights).
The functions ${\psi _3}\left(\bullet \right)$ and ${{\psi}' _3}\left(\bullet \right)$
  in  \textbf{Algorithm 6}
are defined by \textbf{Algorithms 7} and \textbf{8}, respectively.
When the ridge parameter $\lambda$ is very small, we can
choose ${{\psi}' _3}\left(\bullet \right)$ for \textbf{Chol-2} instead of ${\psi _3}\left(\bullet \right)$ for \textbf{Chol-1}
    in  \textbf{Algorithm 6},
to obtain
  the numerically more stable implementation.
Notice that \textbf{Algorithm 6} includes (\ref{L_m_def12431RidgeInvJun30}) and
(\ref{W2AMHEWNH3495initial}),
  \textbf{Algorithm 7} includes
(\ref{L_big_BLK_def1}),
(\ref{ZF_def_L_2_items3abJun2021})
and
(\ref{W2AMHEWNH3495}),
 and \textbf{Algorithm 8} includes (\ref{DLLtE1335645}),
(\ref{C_matrix_def_111}),
(\ref{GG2HC943039fd234moreStable}),
(\ref{T2DG32902394309}),   (\ref{L_big_BLK_def1}) and  (\ref{W2WTGCeiwo32dsk}).

\begin{algorithm}[htb]
\caption{:~\bf The Proposed Efficient Ridge Solution for BLS / its Numerically More Stable Implementation
 (i.e., \textbf{Chol-1} / \textbf{Chol-2}):  Initialization and Update of  Inverse Cholesky Factor and Output Weights}
\begin{algorithmic}
\State \quad \quad  $\cdots \cdots$
\State {\footnotesize 12:}  Compute the upper-triangular ${{\mathbf{\ddot F}}_n^{m}}$  satisfying
\State \quad \;  ${{\mathbf{\ddot F}}_n^{m}}({{\mathbf{\ddot F}}_n^{m}})^{T}= {{\left( ({{\mathbf{\ddot A}}_n^{m}})^{T}{{\mathbf{\ddot A}}_n^{m}} +\lambda \mathbf{I} \right)}^{-1}}$,
\State \quad \;  and compute $\mathbf{\ddot W}_n^{m} ={{\mathbf{\ddot F}}_n^{m}}({{\mathbf{\ddot F}}_n^{m}})^{T} ({{\mathbf{\ddot A}}_n^{m}})^{T} \mathbf{Y}$;
\State \quad \quad  $\cdots \cdots$
\State {\footnotesize 18:}  Compute ${\mathbf{\ddot F}}_n^{m+1}$ and $\mathbf{\ddot W}_n^{m+1}$  by
\State \quad \;  ${\psi _3}\left({{{\mathbf{\ddot F}}_n^{m}}}, {{\mathbf{\ddot A}}_n^{m}}, \mathbf{\ddot H}_{m+1},  \mathbf{\ddot W}_n^{m}, {\mathbf{Y}}\right)$ or ${{\psi}' _3}\left( \cdots \right)$;
\State \quad \quad  $\cdots \cdots$
\State {\footnotesize 30:}  Compute $\mathbf{\ddot F}_{n+1}^{m}$ and $\mathbf{\ddot W}_{n+1}^{m}$  by
\State \quad \;  ${\psi _3}\left({{{\mathbf{\ddot F}}_n^{m}}}, {{\mathbf{\ddot A}}_n^{m}}, \left[{{\mathbf{\ddot Z}}_{n+1}}|{{\mathbf{\ddot H}}_{e{{x}_{m}}}} \right],  \mathbf{\ddot W}_n^{m}, {\mathbf{Y}}\right)$ or ${{\psi}' _3}\left( \cdots \right)$;
\State \quad \quad  $\cdots \cdots$
\end{algorithmic}
\end{algorithm}

\begin{algorithm}
\caption{The  Algorithm to Update ${{\mathbf{F}}_{k}}$ and ${{\mathbf{W}}_{k}}$ for \textbf{Chol-1}}\label{euclid}
\begin{algorithmic}[0]
\Function{${\psi _3}$}{${{\mathbf{ F}}_k}$, ${{\mathbf{A}}_k}$, ${{\mathbf{H}}}$,  ${{\mathbf{ W}}_k}$, ${\mathbf{Y}}$}
\State Compute $\mathbf{ G}$ by
  ${{\mathbf{G}}}{{\mathbf{G}}^{T}}= \left( \begin{array}{c}
{{{\bf{H}}^{T}{{\bf{H}}}+\lambda \mathbf{I}}-} \\
{{{\mathbf{ H}}^T}{{\mathbf{ A}}_k}{{\bf{F}}_k}{{\bf{F}}_k^{T}}{{{\mathbf{ A}}_k^T}{{\mathbf{ H}}}}}
\end{array} \right)^{-1}$
\State $\mathbf{T}=-{{\bf{F}}_k}{{\bf{F}}_k^{T}}{{{\mathbf{ A}}_k^T}{{\mathbf{ H}}}}\mathbf{G}$
\State ${{\mathbf{ F}}_{k+q}}=\left[ \begin{matrix}
   {{\mathbf{ F}}_k} & \mathbf{ T}  \\
   \mathbf{0} & \mathbf{ G}  \\
\end{matrix} \right]$
\State ${ {\mathbf{ W}}_{k+q} }=\left[ \begin{matrix}
   {{\mathbf{ W}}_k}+\mathbf{ T}{{\mathbf{ G}}^{T}}\left( {\mathbf{ H}}^{T}\mathbf{Y}-{{\mathbf{ H}}^T}{{\mathbf{ A}}_k}{{\mathbf{ W}}_k} \right)  \\
   \mathbf{ G}{{\mathbf{ G}}^{T}}\left( {\mathbf{ H}}^{T}\mathbf{Y}-{{\mathbf{ H}}^T}{{\mathbf{ A}}_k}{{\mathbf{ W}}_k} \right)  \\
\end{matrix} \right]$
\State \textbf{return} ${{\mathbf{ F}}_{k+q}}$, ${{\mathbf{ W}}_{k+q}}$
\EndFunction
\end{algorithmic}
\end{algorithm}

\begin{algorithm}
\caption{The  Numerically More Stable Implementation to Update ${{\mathbf{F}}_{k}}$ and ${{\mathbf{W}}_{k}}$ for \textbf{Chol-2}}\label{euclid}
\begin{algorithmic}[0]
\Function{${{\psi}' _3}$}{${{\mathbf{ F}}_k}$, ${{\mathbf{A}}_k}$, ${{\mathbf{H}}}$,  ${{\mathbf{ W}}_k}$, ${\mathbf{Y}}$}
\State $\mathbf{D}={{\bf{F}}_k}{{\bf{F}}_k^{T}}{{\mathbf{ A}}_k^T}{{\mathbf{ H}}}$
\State $\mathbf{C}={{\bf{H}}}-{\mathbf{ A}_k}\mathbf{D}$
\State Compute $\mathbf{ G}$ by
${{\mathbf{G}}}{{\mathbf{G}}^{T}}= \left( {{\mathbf{C}_{{}}^{T}\mathbf{C}}}+\lambda \mathbf{D}^{T}\mathbf{D}  + \lambda {\bf{I}} \right)^{-1}$
\State $\mathbf{T}=-\mathbf{D} \mathbf{G}$
\State ${{\mathbf{ F}}_{k+q}}=\left[ \begin{matrix}
   {{\mathbf{ F}}_k} & \mathbf{ T}  \\
   \mathbf{0} & \mathbf{ G}  \\
\end{matrix} \right]$
\State ${{\bf{ W}}_{k+q}} =\left[ \begin{matrix}
   {{\bf{ W}}_k} + \mathbf{T} {{\mathbf{G}}^{T}} {\bf{C}}^{T} {{\mathbf{Y}}}  \\
   \mathbf{G}{{\mathbf{G}}^{T}}{\bf{C}}^{T} {{\mathbf{Y}}}  \\
\end{matrix} \right]$
\State \textbf{return} ${{\mathbf{ F}}_{k+q}}$, ${{\mathbf{ W}}_{k+q}}$
\EndFunction
\end{algorithmic}
\end{algorithm}

\subsection{Comparison of Existing and Proposed Efficient BLS Algorithms by Inverse Cholesky Factorization}

  As we write (\ref{W2AMHEWNH3495}) as (\ref{W2WTGCeiwo32dsk}),
 we utilize
(\ref{W2AinvY989565forwk322Jun17fdef}) and (\ref{C2HAFFAH03290kdf43fd}) in turn
to obtain
${\mathbf{ H}}^{T}\mathbf{Y}-{{\mathbf{ H}}^T}{{\mathbf{ A}}_k}{{\mathbf{\bar W}}_k}={\mathbf{ H}}^{T}\mathbf{Y}-{{\mathbf{ H}}^T}  {{\mathbf{ A}}_k} {{\mathbf{\bar F}}_k} {{\mathbf{\bar F}}_k^{T}}{{\mathbf{A}}_k^T}\mathbf{Y}={\mathbf{\bar C}}\mathbf{Y}$, which is
substituted into
  (\ref{W2AMHEWNH3495pseudoInv329ak})
  to write it
as
 \begin{equation}\label{W2AMHEWNH3495pseudoInv329aksimple}{ {\mathbf{\bar W}}_{k+q} }=\left[ \begin{matrix}
   {{\mathbf{\bar W}}_k} + \mathbf{\bar T}  {{\mathbf{\bar G}}^{T}}  {\mathbf{\bar C}}\mathbf{Y}  \\
   \mathbf{\bar G}{{\mathbf{\bar G}}^{T}}   {\mathbf{\bar C}}\mathbf{Y}  \\
\end{matrix} \right].
\end{equation}
Then we can compare \textbf{Algorithm 8} to
 \textbf{Algorithm 5} with   (\ref{W2AMHEWNH3495pseudoInv329ak}) replaced by
(\ref{W2AMHEWNH3495pseudoInv329aksimple}),
to conclude that  the only difference
between the existing and proposed efficient BLS algorithms
 lies
between
(\ref{GG2HC943039fd234moreStable})
 and
 (\ref{ZF_def_L_2_items3a}).

Compared to (\ref{ZF_def_L_2_items3a}),
(\ref{GG2HC943039fd234moreStable}) only adds the entry $\lambda \mathbf{D}^{T}\mathbf{D}$.
Then (\ref{GG2HC943039fd234moreStable}) is equal to (\ref{ZF_def_L_2_items3a})
 when
  $\lambda \to 0$, while
    usually
    (\ref{GG2HC943039fd234moreStable}) is different from  (\ref{ZF_def_L_2_items3a})  when
 $\lambda \to 0$ is not satisfied.
Accordingly,  (\ref{ZF_def_L_2_items3a}) can be regarded as the special case  of (\ref{GG2HC943039fd234moreStable}) with $\lambda \to 0$,
while (\ref{GG2HC943039fd234moreStable})
 can be regarded as
  an extension of  (\ref{ZF_def_L_2_items3a}).

\section{Proposed Ridge Solution for BLS by Updating the Ridge Inverse}

The BLS algorithm proposed in the last section
utilizes the sub-matrices in the inverse Cholesky
factor ${{\bf{F}}_{k+q}}$ to update the output weights,
and updates the inverse Cholesky factor
efficiently. In this section, we will utilize a sub-matrix in the ridge inverse
${\mathbf{ A}_{k+q}^\dag}$
 to update the output weights,
 and we will also propose an algorithm to update the ridge inverse.
 The proposed  algorithm to update the ridge inverse,
 which actually computes the ridge inverse of a partitioned matrix,
 can be regarded as an extension of
 the well-known Greville's method~\cite{cite_general_inv_book}
  that can only compute the generalized inverse of a partitioned matrix.
  Specifically, we will show that
 the proposed  algorithm to update the ridge inverse is
 equivalent to
  the Greville's Method when the ridge parameter
  $\lambda \to 0$, i.e., it includes
  the Greville's method  as a special case.

\begin{table}[!t]
\footnotesize
\renewcommand{\arraystretch}{1.3}
\newcommand{\tabincell}[2]{\begin{tabular}{@{}#1@{}}#2\end{tabular}}
\caption{Comparison of Flops between the Existing and Proposed
Efficient BLS Algorithms by Inverse Cholesky
Factorization} \label{table_example} \centering
\begin{tabular}{|c|c|c|c|}
\multicolumn{2}{|c|}{{\bfseries Existing Alg. 5 (for Exg-E)}}   &  \multicolumn{2}{c|}{{\bfseries Proposed Alg. 7 (for Chol-1)}}  \\
\bfseries  \tabincell{c}{Eqn. }    & {\bfseries   \tabincell{c}{Flops}}  & \bfseries  \tabincell{c}{Eqn. }    &
{\bfseries   \tabincell{c}{Flops}}  \\
\hline
 \bfseries  (\ref{C_def121numda2GetD})  & $2qkl+2q k^2$           &  \multirow{2}{*}{${{\mathbf{ A}}_k^T}{{\mathbf{ H}}}$}     &  \multirow{2}{*}{$2qkl$}   \\
\cline{1-2}
  \bfseries  (\ref{C_matrix_def_111Orig})   & $2qkl$                  &  &   \\
\hline
\bfseries (\ref{ZF_def_L_2_items3a}) &  $ q^2 l +\frac{2}{3}q^3$                   & \bfseries (\ref{ZF_def_L_2_items3aJun2021}) & $q^2 l+q k^2 + q^2 k +\frac{2}{3}q^3$  \\
\hline
\bfseries (\ref{ZF_def_L_2_items3b}) & $q^2 k$                                 & \bfseries (\ref{ZF_def_L_2_items3bJun2021}) & $q k^2 + q^2 k $          \\
\hline
  \bfseries  (\ref{W2AMHEWNH3495pseudoInv329ak})  &   $2cql+4cqk +2 c q^2$                               & \bfseries (\ref{W2AMHEWNH3495}) &
    $2cql+4cqk +2 c q^2$  \\
\hline
   \bfseries Sum & \tabincell{c}{$(4qk+q^2+2cq)l+$ \\ $2q k^2+(q^2+4cq)k$ \\ $+\frac{2}{3}q^3+2c q^2$\\$\approx(4qk+q^2)l$ }
    & \bfseries Sum &   \tabincell{c}{$(2qk+q^2+2cq)l+$ \\ $2q k^2+(2q^2+4cq)k$ \\ $+\frac{2}{3}q^3+2c q^2$ \\$\approx (2qk+q^2)l$}
        \\
\hline
\end{tabular}
\end{table}

  \subsection{The Algorithm to Update the Ridge Inverse  and Solution without Using the Inverse Cholesky
Factor (i.e., R-Inv)}


The ridge inverse and solution  are updated in (\ref{AbigByALLENNEA084945}) and (\ref{W2AMHEWNH3495}), respectively,
by utilizing  the sub-matrices in the inverse Cholesky factor ${{\bf{F}}_{k+q}}$, i.e., $\mathbf{G}$ and
 $\mathbf{T}$
 in   (\ref{L_big_BLK_def1}).
To avoid using any sub-matrix in ${{\bf{F}}_{k+q}}$,
  we will propose the ridge solution by
   updating the ridge inverse  in this subsection,
  which is abbreviated as \textbf{R-Inv}.

 To update the ridge inverse,
 substitute
 (\ref{C_matrix_def_111})
 and
(\ref{T2DG32902394309})
  into (\ref{AbigByALLENN121}) to get
  \begin{equation}\label{A2ADGGC302ds23sd}
{\mathbf{ A}_{k+q}^\dag}=\left[ {\begin{array}{*{20}{c}}
{\begin{array}{l}
{{\bf{ A}}_k^\dag } -\mathbf{D}\mathbf{G} {{\bf{G}}^T} {\bf{C}}^T
\end{array}}\\
{{\bf{G}}{{\bf{G}}^T}{\bf{C}}^T}
\end{array}} \right].
 \end{equation}
Then we write (\ref{A2ADGGC302ds23sd}) as
  \begin{equation}\label{A_1col_inv_book1matrix}
 {\mathbf{ A}_{k+q}^\dag}={{\left[ {\mathbf{ A}_k}|{{\bf{H}}} \right]}^{\dagger}}=\left[ \begin{matrix}
   {{\bf{ A}}_k^\dag}-\mathbf{D}{{\mathbf{B}}^{T}}  \\
   {{\mathbf{B}}^{T}}  \\
\end{matrix} \right],
\end{equation}
where
$\mathbf{D}$
has been defined by  (\ref{DLLtE1335645}), and
\begin{equation}\label{B2NNHELLA144563}
{{\mathbf{B}}^{T}}=\mathbf{G}{{\mathbf{G}}^{T}} {{\mathbf{C}}^{T}}.
\end{equation}
To  update the ridge solution,
  substitute
(\ref{B2NNHELLA144563})
into
(\ref{W2WTGCeiwo32dsk})
to get
   \begin{equation}\label{W2WDBB94398}
{{\bf{ W}}_{k+q}} =\left[ {\begin{array}{*{20}{c}}
{{{\bf{ W}}_k} - {\bf{D}}{{\bf{B}}^T}{\bf{Y}}}\\
{{{\bf{B}}^T}{\bf{Y}}}
\end{array}} \right].
 \end{equation}
In (\ref{DLLtE1335645}) and (\ref{B2NNHELLA144563}),
 the sub-matrices in ${{\bf{F}}_{k+q}}$ are still utilized, e.g., ${{\bf{F}}_k}$ and  $\mathbf{G}$.
 So we substitute (\ref{A_gen_inv_def_by_L_112}) into (\ref{DLLtE1335645})
 to obtain
 \begin{equation}\label{D_matrix_def_111}
\mathbf{D} = {{\bf{ A}}_k^\dag}{{\bf{H}}},
\end{equation}
and  substitute
(\ref{GG2HC943039fd234moreStable})
into
(\ref{B2NNHELLA144563}) to obtain
 \begin{equation}\label{B2HCnumdaICt2021JunSmallLamda}
 {{\mathbf{B}}^{T}}={{\left({{\mathbf{C}_{{}}^{T}\mathbf{C}}}+\lambda \mathbf{D}^{T}\mathbf{D}+\lambda \mathbf{I} \right)}^{-1}}{{\mathbf{C}}^{T}}.
\end{equation}

We can also compute ${{\mathbf{B}}^{T}}$ by
\begin{equation}\label{B2HCnumdaICt}
{{\mathbf{B}}^{T}}={{\left({\bf{H}}^{T}{\mathbf{C}}+\lambda \mathbf{I} \right)}^{-1}}{{\mathbf{C}}^{T}},
\end{equation}
which is obtained by
substituting  (\ref{GG2HC943039fd234})
into (\ref{B2NNHELLA144563}).
However,
usually  (\ref{B2HCnumdaICt2021JunSmallLamda})
is numerically more stable than (\ref{B2HCnumdaICt}) in the simulations when the ridge parameter $\lambda$ is small.
 This may be explained by the fact that the theoretical
 positive definite
 ${\bf{H}}^{T}{\mathbf{C}}+\lambda \mathbf{I}$ in
 (\ref{B2HCnumdaICt}) may no longer be
 positive definite
due to numerical errors when  $\lambda$ is small,
while  ${{\mathbf{C}_{{}}^{T}\mathbf{C}}}+\lambda \mathbf{D}^{T}\mathbf{D}+\lambda \mathbf{I}$
 always results in a positive definite matrix, as shown in (\ref{xCCDDlamdax2930dsk23}).

\subsection{Construction Model and Learning Procedure of R-Inv}

 In the proposed ridge solution for BLS based on the ridge inverse,
  the construction model and learning procedure
in \textbf{Algorithm 1} are followed, and
\textbf{Algorithm 9} is utilized to
implement
the steps in rows 12, 18 and 30 of \textbf{Algorithm 1} (which compute the output weights).
The function ${\psi _4}\left(\bullet \right)$
  in  \textbf{Algorithm 9}
is defined by \textbf{Algorithm 10}.
Notice that
 \textbf{Algorithm 9} includes (\ref{Generelized_inv_def1}) and (\ref{W2AY9845729}),
while   \textbf{Algorithm 10} includes (\ref{D_matrix_def_111}), (\ref{C_matrix_def_111}),
(\ref{B2HCnumdaICt2021JunSmallLamda}),
 (\ref{A_1col_inv_book1matrix})
and
(\ref{W2WDBB94398}).

\begin{algorithm}[htb]
\caption{:~\bf  The Proposed Ridge Solution for BLS by
   Updating the Ridge Inverse (i.e., \textbf{R-Inv}):  Initialization and Update of  Ridge Inverse and Output Weights}
\begin{algorithmic}
\State \quad \quad  $\cdots \cdots$ 
\State {\footnotesize 12:}  Compute   $({{\mathbf{\ddot A}}_n^{m}})^{\dagger}= {{\left( ({{\mathbf{\ddot A}}_n^{m}})^{T}{{\mathbf{\ddot A}}_n^{m}} +\lambda \mathbf{I} \right)}^{-1}} ({{\mathbf{\ddot A}}_n^{m}})^{T}$
\State \quad \;  and  $\mathbf{\ddot W}_n^{m} =({{\mathbf{\ddot A}}_n^{m}})^{\dagger}\mathbf{Y}$;
\State \quad \quad  $\cdots \cdots$
\State {\footnotesize 18:}  Compute $({\mathbf{\ddot A}}_n^{m+1})^{\dagger}$ and $\mathbf{\ddot W}_n^{m+1}$  by
\State \quad \;  ${\psi _4}\left({({{\mathbf{\ddot A}}_n^{m}})^{\dagger}}, {{\mathbf{\ddot A}}_n^{m}}, \mathbf{\ddot H}_{m+1},  \mathbf{\ddot W}_n^{m}, {\mathbf{Y}}\right)$;
\State \quad \quad  $\cdots \cdots$
\State {\footnotesize 30:}  Compute $(\mathbf{\ddot A}_{n+1}^{m})^{\dagger}$ and $\mathbf{\ddot W}_{n+1}^{m}$  by
\State \quad \;  ${\psi _4}\left({({{\mathbf{\ddot A}}_n^{m}})^{\dagger}}, {{\mathbf{\ddot A}}_n^{m}}, \left[{{\mathbf{\ddot Z}}_{n+1}}|{{\mathbf{\ddot H}}_{e{{x}_{m}}}} \right],  \mathbf{\ddot W}_n^{m}, {\mathbf{Y}}\right)$;
\State \quad \quad  $\cdots \cdots$
\end{algorithmic}
\end{algorithm}

\begin{algorithm}
\caption{\small The Algorithm to Update ${{\bf{ A}}_k^\dag}$ and ${{\mathbf{W}}_{k}}$ for \textbf{R-Inv}}\label{euclid}
\begin{algorithmic}[0]
\Function{${\psi _4}$}{${{\mathbf{A}}_k^{\dagger}}$, ${{\mathbf{A}}_k}$, ${{\mathbf{H}}}$,  ${{\mathbf{ W}}_k}$, ${\mathbf{Y}}$}
\State $\mathbf{ D} = {{\mathbf{A}}_k^{\dagger}}{{\mathbf{H}}}$
\State $\mathbf{ C}={{\mathbf{H}}}-{{\mathbf{A}}_k}\mathbf{ D}$
\State $ {{\mathbf{B}}^{T}}={{\left({{\mathbf{C}_{{}}^{T}\mathbf{C}}}+\lambda \mathbf{D}^{T}\mathbf{D}+\lambda \mathbf{I} \right)}^{-1}}{{\mathbf{C}}^{T}}$
\State $ {{\mathbf{A}}_{k + q}^{\dagger}}=\left[ \begin{matrix}
   {{\mathbf{A}}_k^{\dagger}}-\mathbf{ D}{{\mathbf{ B}}^{T}}  \\
   {{\mathbf{ B}}^{T}}  \\
\end{matrix} \right]$
\State ${{\mathbf{ W}}_{k+q}}{{ = }}\left[ {\begin{array}{*{20}{c}}
{{{\mathbf{ W}}_k} - {\mathbf{ D}}{{\mathbf{ B}}^T}{\mathbf{Y}}}\\
{{{\mathbf{ B}}^T}{\mathbf{Y}}}
\end{array}} \right]$
\State \textbf{return} ${{\mathbf{A}}_{k + q}^{\dagger}}$, ${{\mathbf{ W}}_{k+q}}$
\EndFunction
\end{algorithmic}
\end{algorithm}


\subsection{Comparison of the Proposed Algorithm to Update the Ridge Inverse
and the Greville's Method to Update the Generalized Inverse}

By comparing
\textbf{Algorithm 3} including (\ref{A_1col_inv_book1matrixOrig})-(\ref{W2WDBB94398Orig})
with \textbf{Algorithm 10} including  (\ref{D_matrix_def_111}), (\ref{C_matrix_def_111}),
(\ref{B2HCnumdaICt2021JunSmallLamda}),
 (\ref{A_1col_inv_book1matrix})
and
(\ref{W2WDBB94398}),
it can be seen that
 the only difference
  lies between
(\ref{B_Matrix_def1abOrig}) and (\ref{B2HCnumdaICt2021JunSmallLamda}).

  Appendix C
   shows that
 (\ref{B2HCnumdaICt2021JunSmallLamda})
is equal to
(\ref{B_Matrix_def1abOrig}) when
  $\lambda \to 0$.
  Moreover,
  (\ref{B2HCnumdaICt2021JunSmallLamda})
  is utilized in \textbf{Algorithm 10} to  update the
 ridge inverse, while
 (\ref{B_Matrix_def1abOrig})  is
 utilized in \textbf{Algorithm 3} to  update  the generalized inverse
 by extending
 the Greville's method.
Then it can be
concluded
that
when $\lambda \to 0$,
 the
 proposed algorithm to
 update the
 ridge inverse
 is equivalent to
 the Greville's method
 utilized
  in \cite{BL_trans_paper} to update the generalized inverse.
Accordingly,  the Greville's method to update the generalized inverse can be regarded as the special case with $\lambda \to 0$ of the proposed algorithm
to update the ridge inverse.

When
  $\lambda \to 0$ is not satisfied,
usually $\lambda \mathbf{D}^{T}\mathbf{D} \ne  \mathbf{0}$  and then
it can easily be seen that
    (\ref{B2HCnumdaICt2021JunSmallLamda})
is different from
(\ref{B_Matrix_def1aOrig}) in the case with  ${\bf{\bar C}} \ne \mathbf{0}$.
So
    usually the proposed algorithm
    to
    update the
 ridge inverse
 is different from 
     the Greville's method when
 $\lambda \to 0$ is not satisfied.
Accordingly,
 the proposed algorithm, which actually computes the ridge inverse of a partitioned matrix $\left[ {\mathbf{ A}_k}|{{\bf{H}}} \right]$,
 can be regarded as
  an extension of
 the Greville's method
 that computes the generalized inverse of a partitioned matrix~\cite{cite_general_inv_book}. In other words,
 the proposed algorithm
 expands the application range of the Greville's method
  from the generalized inverse to the ridge inverse.

\begin{table*}[!t]
\scriptsize
\renewcommand{\arraystretch}{1.3}
\newcommand{\tabincell}[2]{\begin{tabular}{@{}#1@{}}#2\end{tabular}}
\caption{ {\footnotesize Snapshot Results of Testing Accuracy on MNIST Dataset with the Ridge Parameter $\lambda={{10}^{-8}}, {{10}^{-7}}, {{10}^{-6}}$}} \label{table_example} \centering
\begin{tabular}{c|c|c|c|c|c|c |c | c | c| c|c |c | c| c |c|}
\hline
\multicolumn{4}{c|}{{\bfseries \tabincell{c}{  Number of}}}  &\multirow{3}*{60}   & 60  &  70   &     80     &    90      &    100     &   110     &   120    &    130     &    140     &    150      &    160              \\
\multicolumn{4}{c|}{{\bfseries \tabincell{c}{ Feature }}}    &   &  $\downarrow$   &     $\downarrow$     &    $\downarrow$      &    $\downarrow$     &   $\downarrow$     &   $\downarrow$    &    $\downarrow$     &    $\downarrow$     &    $\downarrow$      &    $\downarrow$    &    $\downarrow$     \\
\multicolumn{4}{c|}{{\bfseries \tabincell{c}{   Nodes}}}     &   &  70   &     80     &    90      &    100     &   110     &   120    &    130     &    140     &    150      &    160    &    170     \\
 \hline
\multicolumn{4}{c|}{{\bfseries \tabincell{c}{  Number of}}}  &\multirow{3}*{3000}   & 3000  &  5000   &     7000     &    9000      &    11000     &   13000     &   15000    &    17000     &    19000     &    21000      &    23000              \\
\multicolumn{4}{c|}{{\bfseries \tabincell{c}{ Enhancement }}}    &   &  $\downarrow$   &     $\downarrow$     &    $\downarrow$      &    $\downarrow$     &   $\downarrow$     &   $\downarrow$    &    $\downarrow$     &    $\downarrow$     &    $\downarrow$      &    $\downarrow$    &    $\downarrow$     \\
\multicolumn{4}{c|}{{\bfseries \tabincell{c}{   Nodes}}}     &   &  5000   &     7000     &    9000      &    11000     &   13000     &   15000    &    17000     &    19000     &    21000      &    23000    &    25000     \\
\hline
\hline
 \multirow{12}*{\rotatebox{90}{{\bfseries   \tabincell{c}{Testing Accuracy ($\% $) for $\lambda= {{10}^{-8}}$}} }}  &  \multirow{4}*{\rotatebox{90}{Gen. Inv. Sol. }}    & \multirow{2}*{\textbf{Orig.}}    &Mean      &  98.30  &   98.62  &   98.76  &   98.85  &  \textbf{98.88}  &  98.88  &   98.86  &   98.81  &   98.73  &   98.59  &   98.39  &   98.10     \\
\cdashline{4-16}
   &     &    &  Std    &  0.060  &    0.058  &    0.059  &    0.054  &   \textbf{0.052}  &    0.059  &    0.065  &    0.071  &    0.072  &    0.076  &    0.098  &    0.135     \\
\cline{3-16}
  &      & \multirow{2}*{\textbf{Exg-E}}    &Mean      & 98.30  &   98.62  &   98.76  &   98.85  &   \textbf{98.89}  &   98.88  &   98.86  &   98.82  &   98.73  &   98.61  &   98.42  &   98.16      \\
\cdashline{4-16}
   &     &    &  Std    & 0.060  &    0.058  &    0.059  &    0.053  &    \textbf{0.054}  &    0.053  &    0.056  &    0.063  &    0.069  &    0.085  &    0.101  &    0.148     \\
\cline{2-16}
  &\multirow{8}*{\rotatebox{90}{Ridge Solutions}}       &\multirow{2}*{\textbf{Chol-1}}    &Mean      &  98.30  &   98.62  &   98.76  &   98.85  &   98.90  &   NA  &   NA  &   NA  &   NA  &   NA  &   NA  &   NA      \\
 \cdashline{4-16}
   &     &    &  Std    &  0.060  &    0.057  &    0.061  &    0.048  &    0.050  &   NA  &   NA  &   NA  &   NA  &   NA  &   NA  &   NA     \\
\cline{3-16}
  &      & \multirow{2}*{\textbf{Chol-2}}    &Mean      & 98.30  &   98.62  &   98.76  &   98.85  &   98.89  &   98.90  &   98.92  &   98.93  &   98.93  &   98.92  &   98.91  &   98.89      \\
 \cdashline{4-16}
   &     &    &  Std    &  0.060  &    0.057  &    0.060  &    0.047  &    0.047  &    0.052  &    0.059  &    0.050  &    0.050  &    0.057  &    0.055  &    0.059     \\
\cline{3-16}
  &      & \multirow{2}*{\textbf{R-Inv}}    &Mean      &  98.30  &   98.62  &   98.76  &   98.85  &   98.89  &   98.90  &   98.92  &   98.92  &   98.91  &   98.89  &   98.88  &   98.86      \\
 \cdashline{4-16}
   &     &    &  Std    &  0.060  &    0.057  &    0.059  &    0.047  &    0.048  &    0.054  &    0.053  &    0.058  &    0.061  &    0.071  &    0.075  &    0.073     \\
\cline{3-16}
  &      & \multirow{2}*{\textbf{D-Rdg}}    &Mean      &  98.30  &   98.62  &   98.76  &   98.85  &   98.90  &   98.91  &   98.93  &   98.94  &   98.95  &   98.95  &   98.95  &   98.95     \\
  \cdashline{4-16}
   &     &    &  Std    & 0.060  &    0.057  &    0.060  &    0.047  &    0.047  &    0.049  &    0.052  &    0.043  &    0.043  &    0.045  &    0.040  &    0.039      \\
\hline
\hline
 \multirow{12}*{\rotatebox{90}{{\bfseries   \tabincell{c}{Testing Accuracy ($\% $) for $\lambda= {{10}^{-7}}$}} }}  &  \multirow{4}*{\rotatebox{90}{Gen. Inv. Sol. }}    & \multirow{2}*{\textbf{Orig.}}    &Mean      &   98.30  &   98.58  &   98.72  &   98.79  &   98.81  &   98.77  &   98.69  &   98.52  &   98.27  &   97.91  &   97.38  &   96.69         \\
\cdashline{4-16}
   &     &    &  Std    &    0.072   &    0.067   &    0.042   &    0.050   &    0.067   &    0.094   &    0.142   &    0.209   &    0.311   &    0.425   &    0.588   &    0.764        \\
\cline{3-16}
  &      & \multirow{2}*{\textbf{Exg-E}}    &Mean      &  98.30  &   98.58  &   98.72  &   98.80  &   98.82  &   98.78  &   98.68  &   98.53  &   98.27  &   97.90  &   97.40  &   96.73         \\
\cdashline{4-16}
   &     &    &  Std    &   0.072   &    0.066   &    0.042   &    0.051   &    0.068   &    0.092   &    0.143   &    0.215   &    0.305   &    0.422   &    0.579   &    0.763       \\
\cline{2-16}
  &\multirow{8}*{\rotatebox{90}{Ridge Solutions}}       &\multirow{2}*{\textbf{Chol-1}}    &Mean      &     98.30  &   98.57  &   98.72  &   98.81  &   98.86  &   98.90  &   98.93  &  \textbf{98.95}  &  NA  &  NA  &  NA  &  NA        \\
 \cdashline{4-16}
   &     &    &  Std    &    0.072   &    0.062   &    0.051   &    0.053   &    0.053   &    0.050   &    0.044   &    \textbf{0.046}   &  NA  &  NA  &  NA  &  NA        \\
\cline{3-16}
  &      & \multirow{2}*{\textbf{Chol-2}}    &Mean      &    98.30  &   98.57  &   98.72  &   98.81  &   98.87  &   98.90  &   98.93  &  98.95  &   98.96  &   98.95  &   \textbf{98.96}  &   98.96         \\
 \cdashline{4-16}
   &     &    &  Std    &   0.072   &    0.062   &    0.050   &    0.054   &    0.055   &    0.049   &    0.051   &    0.047   &    0.049   &    0.056   &    \textbf{0.046}   &    0.046      \\
\cline{3-16}
  &      & \multirow{2}*{\textbf{R-Inv}}    &Mean      &   98.30  &   98.57  &   98.71  &   98.81  &   98.87  &   98.90  &   98.93  &   98.94  &  \textbf{98.95}  &   98.95  &   98.95  &   98.95      \\
 \cdashline{4-16}
   &     &    &  Std    &   0.072   &    0.062   &    0.050   &    0.052   &    0.053   &    0.050   &    0.050   &    0.051   &    \textbf{0.039}   &    0.049   &    0.044   &    0.041        \\
\cline{3-16}
  &      & \multirow{2}*{\textbf{D-Rdg}}    &Mean      &   98.30  &   98.57  &   98.72  &   98.81  &   98.86  &   98.90  &   98.93  &   98.95  &   98.96  &   98.97  &   98.97  &   98.98       \\
  \cdashline{4-16}
   &     &    &  Std    & 0.072   &    0.062   &    0.051   &    0.053   &    0.053   &    0.049   &    0.045   &    0.047   &    0.041   &    0.044   &    0.040   &    0.044       \\
\hline
\hline
 \multirow{12}*{\rotatebox{90}{{\bfseries   \tabincell{c}{Testing Accuracy ($\% $) for $\lambda= {{10}^{-6}}$}} }}  &  \multirow{4}*{\rotatebox{90}{Gen. Inv. Sol. }}    & \multirow{2}*{\textbf{Orig.}}    &Mean      &  98.22  &   98.50  &   98.65  &   98.70  &   98.65  &   98.50  &   98.22  &   97.79  &   97.15  &   96.32  &   95.19  &   93.82      \\
\cdashline{4-16}
   &     &    &  Std    &    0.071 &    0.060 &    0.055 &    0.069 &    0.090 &    0.149 &    0.260 &    0.421 &    0.659 &    0.940 &    1.240 &    1.602        \\
\cline{3-16}
  &      & \multirow{2}*{\textbf{Exg-E}}    &Mean      &    98.22  &   98.50  &   98.65  &   98.69  &   98.65  &   98.50  &   98.22  &   97.79  &   97.16  &   96.33  &   95.21  &   93.82        \\
\cdashline{4-16}
   &     &    &  Std    &   0.071 &    0.060 &    0.055 &    0.067 &    0.093 &    0.148 &    0.266 &    0.420 &    0.653 &    0.929 &    1.260 &    1.599        \\
\cline{2-16}
  &\multirow{8}*{\rotatebox{90}{Ridge Solutions}}       &\multirow{2}*{\textbf{Chol-1}}    &Mean      &  98.22  &   98.48  &   98.64  &   98.73  &   98.79  &   98.83  &   98.86  &   98.88  &   98.91  &   98.93  &   98.94  &   \textbf{98.96}       \\
 \cdashline{4-16}
   &     &    &  Std    &   0.071 &    0.061 &    0.060 &    0.053 &    0.050 &    0.052 &    0.041 &    0.045 &    0.041 &    0.041 &    0.037 &    \textbf{0.042}        \\
\cline{3-16}
  &      & \multirow{2}*{\textbf{Chol-2}}    &Mean      &  98.22  &   98.48  &   98.64  &   98.73  &   98.79  &   98.83  &   98.86  &   98.88  &   98.91  &   98.92  &   98.94  &   \textbf{98.95}      \\
 \cdashline{4-16}
   &     &    &  Std    &    0.071 &    0.060 &    0.060 &    0.053 &    0.049 &    0.052 &    0.043 &    0.044 &    0.037 &    0.040 &    0.034 &    \textbf{0.034}       \\
\cline{3-16}
  &      & \multirow{2}*{\textbf{R-Inv}}    &Mean      &   98.22  &   98.48  &   98.64  &   98.73  &   98.79  &   98.83  &   98.86  &   98.89  &   98.91  &   98.92  &   98.94  &   \textbf{98.95}        \\
 \cdashline{4-16}
   &     &    &  Std    &    0.071 &    0.060 &    0.060 &    0.052 &    0.049 &    0.049 &    0.039 &    0.046 &    0.040 &    0.043 &    0.038 &    \textbf{0.041}      \\
\cline{3-16}
  &      & \multirow{2}*{\textbf{D-Rdg}}    &Mean      &    98.22  &   98.48  &   98.64  &   98.73  &   98.79  &   98.83  &   98.86  &   98.88  &   98.91  &   98.93  &   98.94  &   98.96        \\
  \cdashline{4-16}
   &     &    &  Std    &    0.071 &    0.061 &    0.060 &    0.053 &    0.050 &    0.052 &    0.041 &    0.045 &    0.041 &    0.041 &    0.038 &    0.042        \\
\hline
\end{tabular}
\end{table*}

\begin{table*}[!t]
\scriptsize
\renewcommand{\arraystretch}{1.3}
\newcommand{\tabincell}[2]{\begin{tabular}{@{}#1@{}}#2\end{tabular}}
\caption{{\footnotesize Snapshot Results of Testing Accuracy on MNIST Dataset  with the Ridge Parameter $\lambda={{10}^{-5}}, {{10}^{-4}}, {{10}^{-3}},  {{10}^{-2}}, {{10}^{-1}}$}} \label{table_example} \centering
\begin{tabular}{c|c|c|c|c|c|c |c | c | c| c|c |c | c| c |c|}
\hline
\multicolumn{4}{c|}{{\bfseries \tabincell{c}{  Number of}}}  &\multirow{3}*{60}   & 60  &  70   &     80     &    90      &    100     &   110     &   120    &    130     &    140     &    150      &    160              \\
\multicolumn{4}{c|}{{\bfseries \tabincell{c}{ Feature }}}    &   &  $\downarrow$   &     $\downarrow$     &    $\downarrow$      &    $\downarrow$     &   $\downarrow$     &   $\downarrow$    &    $\downarrow$     &    $\downarrow$     &    $\downarrow$      &    $\downarrow$    &    $\downarrow$     \\
\multicolumn{4}{c|}{{\bfseries \tabincell{c}{   Nodes}}}     &   &  70   &     80     &    90      &    100     &   110     &   120    &    130     &    140     &    150      &    160    &    170     \\
 \hline
\multicolumn{4}{c|}{{\bfseries \tabincell{c}{  Number of}}}  &\multirow{3}*{3000}   & 3000  &  5000   &     7000     &    9000      &    11000     &   13000     &   15000    &    17000     &    19000     &    21000      &    23000              \\
\multicolumn{4}{c|}{{\bfseries \tabincell{c}{ Enhancement }}}    &   &  $\downarrow$   &     $\downarrow$     &    $\downarrow$      &    $\downarrow$     &   $\downarrow$     &   $\downarrow$    &    $\downarrow$     &    $\downarrow$     &    $\downarrow$      &    $\downarrow$    &    $\downarrow$     \\
\multicolumn{4}{c|}{{\bfseries \tabincell{c}{   Nodes}}}     &   &  5000   &     7000     &    9000      &    11000     &   13000     &   15000    &    17000     &    19000     &    21000      &    23000    &    25000     \\
\hline
\hline
 \multirow{10}*{\rotatebox{90}{{\bfseries   \tabincell{c}{Testing Accuracy ($\% $) \\ for $\lambda= {{10}^{-5}}$}} }}  &  \multirow{4}*{\rotatebox{90}{Gen. Inv. Sol. }}    & \multirow{2}*{\textbf{Orig.}}    &Mean      &  97.98    &   98.27    &   98.43    &   98.47    &   98.38    &   98.17    &   97.78    &   97.20    &   96.34    &   95.18    &   93.70    &   91.88     \\
\cdashline{4-16}
   &     &    &  Std    &      0.094    &    0.079    &    0.082    &    0.094    &    0.138    &    0.225    &    0.396    &    0.645    &    0.943    &    1.336    &    1.807    &    2.270    \\
\cline{3-16}
  &      & \multirow{2}*{\textbf{Exg-E}}    &Mean      &   97.98    &   98.27    &   98.43    &   98.47    &   98.38    &   98.17    &   97.78    &   97.20    &   96.34    &   95.17    &   93.70    &   91.88     \\
\cdashline{4-16}
   &     &    &  Std    &    0.094    &    0.079    &    0.082    &    0.094    &    0.137    &    0.221    &    0.393    &    0.649    &    0.949    &    1.341    &    1.814    &    2.274   \\
\cline{2-16}
  &\multirow{6}*{\rotatebox{90}{Ridge Solutions}}       &\multirow{2}*{\textbf{Chol-2}}    &Mean      &  97.98    &   98.24    &   98.38    &   98.47    &   98.56    &   98.63    &   98.67    &   98.71    &   98.74    &   98.77    &   98.79    &   98.81      \\
 \cdashline{4-16}
   &     &    &  Std    & 0.094    &    0.075    &    0.075    &    0.060    &    0.061    &    0.054    &    0.056    &    0.055    &    0.054    &    0.045    &    0.052    &    0.040       \\
\cline{3-16}
  &      & \multirow{2}*{\textbf{R-Inv}}    &Mean      &  97.98    &   98.24    &   98.38    &   98.48    &   98.56    &   98.63    &   98.66    &   98.71    &   98.74    &   98.77    &   98.79    &   98.81     \\
 \cdashline{4-16}
   &     &    &  Std    &     0.094    &    0.075    &    0.076    &    0.060    &    0.062    &    0.054    &    0.052    &    0.055    &    0.055    &    0.046    &    0.051    &    0.043    \\
\cline{3-16}
  &      & {\textbf{Chol-1} \&}    &Mean      &  97.98    &   98.24    &   98.38    &   98.47    &   98.56    &   98.63    &   98.66    &   98.71    &   98.74    &   98.77    &   98.79    &   98.81      \\
  \cdashline{4-16}
   &     & \textbf{D-Rdg}    &  Std    &   0.094    &    0.075    &    0.075    &    0.060    &    0.061    &    0.055    &    0.055    &    0.055    &    0.054    &    0.046    &    0.052    &    0.041   \\
\hline
\hline
 \multirow{8}*{\rotatebox{90}{{\bfseries   \tabincell{c}{Testing Accuracy ($\% $) \\ for $\lambda= {{10}^{-4}}$}} }}  &  \multirow{4}*{\rotatebox{90}{Gen. Inv. Sol. }}    & \multirow{2}*{\textbf{Orig.}}    &Mean      &   97.47       &   97.84       &   98.01       &   98.06       &   97.93       &   97.66       &   97.17       &   96.42       &   95.35       &   93.95       &   92.21       &   90.12       \\
\cdashline{4-16}
   &     &    &  Std    &   0.105       &    0.101       &    0.074       &    0.099       &    0.135       &    0.221       &    0.383       &    0.646       &    0.981       &    1.389       &    1.843       &    2.301       \\
\cline{3-16}
  &      & \multirow{2}*{\textbf{Exg-E}}    &Mean      &  97.47       &   97.84       &   98.01       &   98.06       &   97.93       &   97.66       &   97.17       &   96.42       &   95.35       &   93.95       &   92.22       &   90.12        \\
\cdashline{4-16}
   &     &    &  Std    &  0.105       &    0.101       &    0.074       &    0.099       &    0.135       &    0.221       &    0.384       &    0.646       &    0.981       &    1.392       &    1.839       &    2.301       \\
\cline{2-16}
  &\multirow{4}*{\rotatebox{90}{Ridge Sol.}}       &\multirow{2}*{\textbf{R-Inv}}    &Mean      &   97.47       &   97.80       &   97.98       &   98.10       &   98.19       &   98.25       &   98.32       &   98.36       &   98.40       &   98.44       &   98.48       &   98.50      \\
 \cdashline{4-16}
   &     &    &  Std    &     0.105       &    0.092       &    0.073       &    0.078       &    0.075       &    0.069       &    0.062       &    0.061       &    0.056       &    0.061       &    0.053       &    0.046       \\
\cline{3-16}
  &      & \textbf{Chol-1, Chol-2}     &Mean      &  97.47       &   97.80       &   97.98       &   98.10       &   98.19       &   98.25       &   98.32       &   98.36       &   98.40       &   98.45       &   98.48       &   98.50        \\
  \cdashline{4-16}
   &     & \& \textbf{D-Rdg}   &  Std    & 0.105       &    0.092       &    0.073       &    0.078       &    0.075       &    0.069       &    0.062       &    0.061       &    0.056       &    0.060       &    0.053       &    0.046       \\
\hline
\hline
 \multirow{6}*{\rotatebox{90}{{\bfseries   \tabincell{c}{Testing \\ Accuracy ($\% $) \\ for $\lambda= {{10}^{-3}}$}} }}  &  \multirow{4}*{\rotatebox{90}{Gen. Inv. Sol. }}    & \multirow{2}*{\textbf{Orig.}}    &Mean      &  96.65      &   97.13      &   97.35      &   97.40      &   97.24      &   96.84      &   96.18      &   95.16      &   93.77      &   92.07      &   90.06      &   87.78      \\
\cdashline{4-16}
   &     &    &  Std    &  0.186      &    0.128      &    0.127      &    0.163      &    0.272      &    0.504      &    0.885      &    1.442      &    2.096      &    2.797      &    3.539      &    4.184         \\
\cline{3-16}
  &      & \multirow{2}*{\textbf{Exg-E}}    &Mean      &  96.65      &   97.13      &   97.35      &   97.40      &   97.24      &   96.84      &   96.18      &   95.16      &   93.77      &   92.07      &   90.06      &   87.78        \\
\cdashline{4-16}
   &     &    &  Std    &   0.186      &    0.128      &    0.127      &    0.163      &    0.272      &    0.504      &    0.885      &    1.442      &    2.096      &    2.797      &    3.539      &    4.184       \\
\cline{2-16}
&  \multirow{2}*{\rotatebox{90}{Rdg.}}  &\textbf{Chol-1, Chol-2,}    &Mean      &   96.65      &   97.06      &   97.30      &   97.48      &   97.62      &   97.71      &   97.78      &   97.85      &   97.90      &   97.96      &   98.00      &   98.04       \\
 \cdashline{4-16}
   &   & \textbf{R-Inv} \& \textbf{D-Rdg}     &  Std    &  0.186      &    0.136      &    0.125      &    0.105      &    0.082      &    0.082      &    0.070      &    0.070      &    0.063      &    0.060      &    0.059      &    0.059          \\
\hline
\hline
\rule{0pt}{10pt}
 \multirow{4}*{\rotatebox{90}{{\bfseries   \tabincell{c}{Testing \\ Accuracy ($\% $) \\ for $\lambda= {{10}^{-2}}$}} }}  & \multirow{2}*{\rotatebox{90}{Gen.}} & \multirow{2}*{{ \textbf{Orig.} \&   \textbf{Exg-E}   }}     &Mean
 &   95.44      &   96.03      &   96.29      &   96.32      &   96.05      &   95.51      &   94.61      &   93.29      &   91.64      &   89.58      &   87.24      &   84.66         \\
\cdashline{4-16}
\rule{0pt}{10pt}
  & &      &  Std    &    0.263      &    0.168      &    0.170      &    0.231      &    0.397      &    0.684      &    1.119      &    1.717      &    2.371      &    3.094      &    3.788      &    4.328      \\
\cline{2-16}
\rule{0pt}{10pt}
  &\multirow{2}*{\rotatebox{90}{Rdg.}} & \textbf{Chol-1}, \textbf{Chol-2},            &Mean      &    95.44      &   95.98      &   96.28      &   96.50      &   96.65      &   96.79      &   96.89      &   96.98      &   97.07      &   97.13      &   97.19      &   97.23        \\
 \cdashline{4-16}
 \rule{0pt}{10pt}
 &  & \textbf{R-Inv} \& \textbf{D-Rdg}         &  Std    &   0.263      &    0.152      &    0.127      &    0.120      &    0.117      &    0.105      &    0.095      &    0.083      &    0.077      &    0.084      &    0.086      &    0.076      \\
\hline
\hline
\rule{0pt}{10pt}
 \multirow{4}*{\rotatebox{90}{{\bfseries   \tabincell{c}{Testing \\ Accuracy ($\% $) \\ for $\lambda= {{10}^{-1}}$}} }}  & \multirow{2}*{\rotatebox{90}{Gen.}} & \multirow{2}*{{ \textbf{Orig.} \&   \textbf{Exg-E}   }}    &Mean      &   93.64      &   94.45      &   94.82      &   94.89      &   94.63      &   94.03      &   93.05      &   91.64      &   89.91      &   87.83      &   85.50      &   83.02            \\
\cdashline{4-16}
\rule{0pt}{10pt}
   &  &    &  Std    &    0.370      &    0.252      &    0.179      &    0.266      &    0.492      &    0.888      &    1.419      &    2.126      &    2.804      &    3.504      &    4.122      &    4.667       \\
\cline{2-16}
\rule{0pt}{10pt}
  & \multirow{2}*{\rotatebox{90}{Rdg.}}  &\textbf{Chol-1}, \textbf{Chol-2},          &Mean      &    93.64      &   94.41      &   94.82      &   95.11      &   95.32      &   95.48      &   95.63      &   95.73      &   95.83      &   95.92      &   95.99      &   96.07        \\
 \cdashline{4-16}
 \rule{0pt}{10pt}
   &  & \textbf{R-Inv} \& \textbf{D-Rdg}         &  Std    &   0.370      &    0.236      &    0.151      &    0.122      &    0.116      &    0.118      &    0.114      &    0.097      &    0.100      &    0.101      &    0.087      &    0.096      \\
\hline
\end{tabular}
\end{table*}

\begin{table*}[!t]
\scriptsize
\renewcommand{\arraystretch}{1.3}
\newcommand{\tabincell}[2]{\begin{tabular}{@{}#1@{}}#2\end{tabular}}
\caption{{\footnotesize Snapshot Results of Testing Accuracy on NORB Dataset with the Ridge Parameter $\lambda= {10}^{-8},{10}^{-7},{10}^{-6},{10}^{-5}$}} \label{table_example} \centering
\begin{tabular}{c|c|c|c|c|c|c |c | c | c| c|c |c |}
 \hline
\multicolumn{4}{c|}{{\bfseries \tabincell{c}{  Number of}}}  &\multirow{3}*{2000}   & 2000  &  3500   &     5000     &    6500      &    8000     &   9500     &   11000    &    12500                   \\
\multicolumn{4}{c|}{{\bfseries \tabincell{c}{ Enhancement }}}    &   &  $\downarrow$   &     $\downarrow$     &    $\downarrow$      &    $\downarrow$     &   $\downarrow$     &   $\downarrow$    &    $\downarrow$     &    $\downarrow$        \\
\multicolumn{4}{c|}{{\bfseries \tabincell{c}{   Nodes}}}     &   &  3500   &     5000     &    6500      &    8000     &   9500     &   11000    &    12500     &    14000         \\
\hline
\hline
 \multirow{12}*{\rotatebox{90}{{\bfseries   \tabincell{c}{Testing Accuracy ($\% $) for $\lambda= {{10}^{-8}}$}} }}  &  \multirow{4}*{\rotatebox{90}{Gen. Inv. Sol. }}    & \multirow{2}*{\textbf{Orig.}}    &Mean      &  88.55    &   88.20    &   87.88    &   87.70    &   87.44    &   86.91    &   86.35    &   85.24    &   83.53    \\
\cdashline{4-13}
   &     &    &  Std    &  0.570    &    0.500    &    0.563    &    0.595    &    0.599    &    0.545    &    0.650    &    0.733    &    0.827        \\
\cline{3-13}
  &      & \multirow{2}*{\textbf{Exg-E}}    &Mean       & 88.55    &   88.20    &   87.88    &   87.70    &   87.43    &   86.92    &   86.33    &   85.28    &   83.61      \\
\cdashline{4-13}
   &     &    &  Std     &     0.570    &    0.500    &    0.565    &    0.594    &    0.597    &    0.545    &    0.634    &    0.691    &    0.764      \\
\cline{2-13}
  &\multirow{8}*{\rotatebox{90}{Ridge Solutions}}       &\multirow{2}*{\textbf{Chol-1}}    &Mean      &   88.55    &   88.21    &   87.90    &   87.73    &   87.50    &   87.08    &   86.78    &   86.39    &   85.84      \\
 \cdashline{4-13}
   &     &    &  Std     &     0.570    &    0.497    &    0.563    &    0.607    &    0.584    &    0.495    &    0.509    &    0.523    &    0.547     \\
\cline{3-13}
  &      & \multirow{2}*{\textbf{Chol-2}}    &Mean      &  88.55    &   88.21    &   87.90    &   87.72    &   87.51    &   87.09    &   86.78    &   86.39    &   85.84    \\
 \cdashline{4-13}
   &     &    &  Std    &    0.570    &    0.497    &    0.563    &    0.608    &    0.584    &    0.495    &    0.512    &    0.524    &    0.545      \\
\cline{3-13}
  &      & \multirow{2}*{\textbf{R-Inv}}    &Mean      &  88.55    &   88.21    &   87.90    &   87.72    &   87.50    &   87.08    &   86.76    &   86.33    &   85.68       \\
 \cdashline{4-13}
   &     &    &  Std    &    0.570    &    0.497    &    0.560    &    0.612    &    0.587    &    0.490    &    0.538    &    0.572    &    0.579    \\
\cline{3-13}
  &      & \multirow{2}*{\textbf{D-Rdg}}    &Mean      &    88.55    &   88.21    &   87.90    &   87.73    &   87.50    &   87.08    &   86.78    &   86.39    &   85.84     \\
  \cdashline{4-13}
   &     &    &  Std    &   0.570    &    0.498    &    0.563    &    0.607    &    0.584    &    0.495    &    0.510    &    0.523    &    0.544      \\
\hline
\hline
 \multirow{12}*{\rotatebox{90}{{\bfseries   \tabincell{c}{Testing Accuracy ($\% $) for $\lambda= {{10}^{-7}}$}} }}  &  \multirow{4}*{\rotatebox{90}{Gen. Inv. Sol. }}    & \multirow{2}*{\textbf{Orig.}}    &Mean      &   88.62        &   88.22        &   87.86        &   87.56        &   87.14        &   86.59        &   85.84        &   84.67        &   82.66     \\
\cdashline{4-13}
   &     &    &  Std    &     0.568        &    0.632        &    0.628        &    0.606        &    0.582        &    0.653        &    0.635        &    0.753        &    0.966   \\
\cline{3-13}
  &      & \multirow{2}*{\textbf{Exg-E}}    &Mean       &  88.62        &   88.22        &   87.86        &   87.55        &   87.14        &   86.60        &   85.84        &   84.69        &   82.69      \\
\cdashline{4-13}
   &     &    &  Std     &   0.568        &    0.632        &    0.627        &    0.608        &    0.581        &    0.652        &    0.631        &    0.737        &    0.943     \\
\cline{2-13}
  &\multirow{8}*{\rotatebox{90}{Ridge Solutions}}       &\multirow{2}*{\textbf{Chol-1}}    &Mean      &    88.62        &   88.24        &   87.88        &   87.64        &   87.32        &   87.05        &   86.83        &   86.67        &   86.67     \\
 \cdashline{4-13}
   &     &    &  Std     &   0.568        &    0.628        &    0.619        &    0.603        &    0.614        &    0.670        &    0.580        &    0.577        &    0.648       \\
\cline{3-13}
  &      & \multirow{2}*{\textbf{Chol-2}}    &Mean      &   88.62        &   88.24        &   87.88        &   87.64        &   87.32        &   87.05        &   86.83        &   86.67        &   86.67       \\
 \cdashline{4-13}
   &     &    &  Std    &   0.568        &    0.627        &    0.619        &    0.602        &    0.614        &    0.671        &    0.580        &    0.578        &    0.649   \\
\cline{3-13}
  &      & \multirow{2}*{\textbf{R-Inv}}    &Mean      & 88.62        &   88.24        &   87.88        &   87.64        &   87.32        &   87.05        &   86.83        &   86.66        &   86.63      \\
 \cdashline{4-13}
   &     &    &  Std    &  0.568        &    0.627        &    0.618        &    0.603        &    0.613        &    0.666        &    0.571        &    0.571        &    0.664      \\
\cline{3-13}
  &      & \multirow{2}*{\textbf{D-Rdg}}    &Mean      & 88.62        &   88.24        &   87.88        &   87.64        &   87.32        &   87.05        &   86.83        &   86.66        &   86.67       \\
  \cdashline{4-13}
   &     &    &  Std    &0.568        &    0.627        &    0.619        &    0.602        &    0.614        &    0.671        &    0.580        &    0.578        &    0.649
     \\
\hline
\hline
 \multirow{12}*{\rotatebox{90}{{\bfseries   \tabincell{c}{Testing Accuracy ($\% $) for $\lambda= {{10}^{-6}}$}} }}  &  \multirow{4}*{\rotatebox{90}{Gen. Inv. Sol. }}    & \multirow{2}*{\textbf{Orig.}}    &Mean      &   88.57    &   88.41    &   88.19    &   87.82    &   87.42    &   86.73    &   85.35    &   83.08    &   79.74     \\
\cdashline{4-13}
   &     &    &  Std    &    0.660    &    0.555    &    0.549    &    0.537    &    0.611    &    0.720    &    1.125    &    1.765    &    2.496      \\
\cline{3-13}
  &      & \multirow{2}*{\textbf{Exg-E}}    &Mean       &    88.57    &   88.41    &   88.19    &   87.82    &   87.42    &   86.72    &   85.36    &   83.09    &   79.74       \\
\cdashline{4-13}
   &     &    &  Std     &   0.660    &    0.555    &    0.548    &    0.536    &    0.608    &    0.723    &    1.123    &    1.776    &    2.499      \\
\cline{2-13}
  &\multirow{8}*{\rotatebox{90}{Ridge Solutions}}       &\multirow{2}*{\textbf{Chol-1}}    &Mean      &   88.57    &   88.49    &   88.36    &   88.18    &   88.14    &   88.18    &   88.21    &   88.31    &   88.43      \\
 \cdashline{4-13}
   &     &    &  Std     &  0.660    &    0.549    &    0.539    &    0.494    &    0.530    &    0.515    &    0.496    &    0.475    &    0.482      \\
\cline{3-13}
  &      & \multirow{2}*{\textbf{Chol-2}}    &Mean      &   88.57    &   88.49    &   88.36    &   88.18    &   88.14    &   88.18    &   88.21    &   88.31    &   88.43      \\
 \cdashline{4-13}
   &     &    &  Std    &  0.660    &    0.549    &    0.539    &    0.494    &    0.530    &    0.515    &    0.496    &    0.474    &    0.482       \\
\cline{3-13}
  &      & \multirow{2}*{\textbf{R-Inv}}    &Mean      &   88.57    &   88.49    &   88.36    &   88.18    &   88.14    &   88.18    &   88.21    &   88.30    &   88.41      \\
 \cdashline{4-13}
   &     &    &  Std    &   0.660    &    0.549    &    0.538    &    0.494    &    0.531    &    0.515    &    0.498    &    0.473    &    0.492    \\
\cline{3-13}
  &      & \multirow{2}*{\textbf{D-Rdg}}    &Mean      &  88.57    &   88.49    &   88.36    &   88.18    &   88.14    &   88.18    &   88.21    &   88.31    &   88.43       \\
  \cdashline{4-13}
   &     &    &  Std    &    0.660    &    0.549    &    0.539    &    0.494    &    0.530    &    0.515    &    0.496    &    0.475    &    0.482     \\
\hline
\hline
 \multirow{12}*{\rotatebox{90}{{\bfseries   \tabincell{c}{Testing Accuracy ($\% $) for $\lambda= {{10}^{-5}}$}} }}  &  \multirow{4}*{\rotatebox{90}{Gen. Inv. Sol. }}    & \multirow{2}*{\textbf{Orig.}}    &Mean      &  88.84    &   88.73    &   88.49    &   87.83    &   86.48    &   84.16    &   80.75    &   76.34    &   71.49     \\
\cdashline{4-13}
   &     &    &  Std    &     0.637    &    0.567    &    0.621    &    0.773    &    1.112    &    1.831    &    2.746    &    3.509    &    4.041     \\
\cline{3-13}
  &      & \multirow{2}*{\textbf{Exg-E}}    &Mean       &   88.84    &   88.73    &   88.49    &   87.83    &   86.48    &   84.16    &   80.75    &   76.33    &   71.49    \\
\cdashline{4-13}
   &     &    &  Std     &   0.637    &    0.567    &    0.621    &    0.773    &    1.112    &    1.830    &    2.745    &    3.507    &    4.041       \\
\cline{2-13}
  &\multirow{8}*{\rotatebox{90}{Ridge Solutions}}       &\multirow{2}*{\textbf{Chol-1}}    &Mean      &    88.84    &   88.98    &   89.10    &   89.10    &   89.14    &   89.22    &   89.29    &   89.38    &   89.46      \\
 \cdashline{4-13}
   &     &    &  Std     &  0.637    &    0.542    &    0.536    &    0.528    &    0.486    &    0.399    &    0.424    &    0.385    &    0.373      \\
\cline{3-13}
  &      & \multirow{2}*{\textbf{Chol-2}}    &Mean      &     88.84    &   88.98    &   89.10    &   89.10    &   89.14    &   89.22    &   89.29    &   89.38    &   89.46       \\
 \cdashline{4-13}
   &     &    &  Std    &  0.637    &    0.543    &    0.536    &    0.528    &    0.486    &    0.399    &    0.424    &    0.385    &    0.373       \\
\cline{3-13}
  &      & \multirow{2}*{\textbf{R-Inv}}    &Mean      &   88.84    &   88.98    &   89.10    &   89.10    &   89.14    &   89.22    &   89.29    &   89.38    &   89.46       \\
 \cdashline{4-13}
   &     &    &  Std    &   0.637    &    0.543    &    0.537    &    0.528    &    0.485    &    0.401    &    0.426    &    0.385    &    0.373      \\
\cline{3-13}
  &      & \multirow{2}*{\textbf{D-Rdg}}    &Mean      &    88.84    &   88.98    &   89.10    &   89.10    &   89.13    &   89.22    &   89.29    &   89.38    &   89.46       \\
  \cdashline{4-13}
   &     &    &  Std    &    0.637    &    0.543    &    0.536    &    0.528    &    0.486    &    0.399    &    0.424    &    0.385    &    0.373     \\
\hline
\end{tabular}
\end{table*}

\section{Complexity Comparison,  Numerical Experiments, and Implementation Aspects}

In this section, we compare the expected flops (floating-point operations) of the presented BLS algorithms. Then
to evaluate the accuracy and training time
of the presented BLS algorithms, we
perform classification experiments
on the Modified National Institute of Standards and Technology (MNIST)
dataset~\cite{61_dataSet} and the New York University object recognition
benchmark (NORB) dataset~\cite{Norb_dataSet}.
 Our simulations are carried out
 on MATLAB software platform under a Microsoft-Windows Server with  $128$ GB of RAM.
The tansig function is adopted for the
enhancement nodes, where
the weights
    ${{\mathbf{W}}_{{{h}_{j}}}}$ and
   the biases
    ${{\mathbf{\beta }}_{{{h}_{j}}}}$ ($j=1,2,\cdots, m$)
    are drawn from the
standard uniform distributions on the interval $\left[ {\begin{array}{*{20}{c}}
{{\rm{ - }}1}&1
\end{array}} \right]$.



\subsection{Complexity Comparison}

When counting the flops, we utilize the well-known fact that
  $l q (2 k - 1) \approx 2 l  k q$ flops are required to multiply
 a $l \times k$ matrix by a $k \times q$ matrix,  and  $l  k=0(l  k q) $ flops are required  to sum two $l \times k$ matrices.
 Moreover, notice that for a $k \times k$ Hermitian matrix,
  $k^3$ flops are required to compute the inverse
   by
  the  inv function in Matlab~\cite{BLSpaper2021zhf1}, while  $\frac{2}{3}k^3$ flops are required to
   compute the inverse Cholesky factorization~\cite{my_inv_chol_paper}.


The original BLS~\cite{BL_trans_paper}
 requires
 $(6qk+3q^2+2cq)l+2cqk+q^3$  flops~\cite{BLSpaper2021zhf1}
   in each update for \textbf{Algorithm 3},
where $c$ denotes the number of output nodes.
As mentioned in  Subsection C of Section \Rmnum{4},
the only difference
 between the
 proposed BLS based on the ridge inverse
 and
 the original BLS
  lies between
  (\ref{B2HCnumdaICt2021JunSmallLamda})
and (\ref{B_Matrix_def1abOrig}).
Compared to~\footnote{When comparing flops, we only consider the usual case where  ${\bf{\bar C}} \ne \mathbf{0}$.}
(\ref{B_Matrix_def1aOrig}),
 (\ref{B2HCnumdaICt2021JunSmallLamda}) requires
 the extra $q^2 k$ flops to compute
$\lambda \mathbf{D}^{T}\mathbf{D}$.
Then the
 proposed BLS based on the ridge inverse
 requires
 $(6qk+3q^2+2cq)l+2cqk+q^3+q^2 k$  flops
 in each update for \textbf{Algorithm 10}.

Table \Rmnum{1} compares
the flops  of \textbf{Algorithms 5} and  \textbf{7}, which update the intermediate results and the output weights
for the existing efficient BLS (i.e., \textbf{Exg-E}) and the proposed efficient BLS (i.e., \textbf{Chol-1}), respectively.
%
We list the flops for each equation,
and give the corresponding total flops.
 To calculate the flops,
 notice that
 ${{\bf{F}}_k^{T}}{{{\mathbf{ A}}_k^T}{{\mathbf{ H}}}}$ computed in  (\ref{ZF_def_L_2_items3aJun2021})
is utilized in
  (\ref{ZF_def_L_2_items3bJun2021}). Moreover, both  ${{\bf{F}}_k}$ and $\mathbf{G}$ are triangular, and  only about half entries need to be computed for the Hermitian matrices.
 It can be seen that (\ref{W2AMHEWNH3495pseudoInv329ak}) and
 (\ref{W2AMHEWNH3495})  have the same form and require the same flops,
 while  the flops for (\ref{W2AMHEWNH3495pseudoInv329ak}) claimed in \cite{BLSpaper2021zhf1} need to be
  revised,  to include the flops of $2cql$, $2cqk$, $ c q^2$, $ c q^2$
 and $2cqk$ to compute ${{\bf{H}}^T} \times {\bf{Y}},({{\bf{H}}^T}{{\bf{A}}_k}) \times {{\bf{\bar W}}_k},{{\bf{\bar G}}^T} \times \left(  \cdots  \right),{\bf{\bar G}} \times {{\bf{\bar G}}^T}\left(  \cdots  \right)$ and ${\bf{\bar T}} \times {{\bf{\bar G}}^T}\left(  \cdots  \right)$, respectively.


To compare
the dominant flops,
we assume
\begin{subnumcases}{\label{q20lc20qISAO312}}
 q=0(l) &  \label{q20lc20qISAO312q}\\
c=0(q),  &  \label{q20lc20qISAO312c}
\end{subnumcases}
as in \cite{BLSpaper2021zhf1},
since usually
the new
added hidden nodes are much less than the training samples,
and we focus on the situation where  the new
added hidden nodes are much more than the output nodes.
Then
 the original BLS~\cite{BL_trans_paper},
the existing efficient  BLS~\cite{BLSpaper2021zhf1}
and the proposed  efficient  BLS
require the dominant flops  of
$(6qk+3q^2)l$,
 $(4qk+q^2)l$ and
 $(2qk+q^2)l$, respectively, for \textbf{Algorithms 3},  \textbf{5} and \textbf{7}.
Thus in each update,
compared to the proposed  efficient BLS,
     the original BLS requires
\begin{equation}\label{Flops3Times32990k32}
\frac{(6qk+3q^2)l}{(2qk+q^2)l}=3
 \end{equation}
   times of
     flops,
     while
       the existing efficient BLS, which spends the extra $2qkl$ flops
   to
get
 $\mathbf{\bar C}={{\bf{H}}}-{\mathbf{ A}_k}\mathbf{\bar D}$
by (\ref{C_matrix_def_111Orig}),
 requires
     \begin{equation}\label{FlopRatioMyChol2OldChol0219as}
\frac{(4qk+q^2)l}{(2qk+q^2)l}=  2-   \frac{1}{1+2(k/q)} >   \frac{5}{3}  \approx 1.67
 \end{equation}
 times~\footnote{Here we only consider the usual case with $k > q$, i.e.,
 the existing nodes are more than the nodes inserted in an update.} of flops.
 Moreover,
 it can be seen that
  \textbf{Algorithms 8}  and \textbf{5}
  require the same dominant flops~\footnote{\textbf{Algorithm 8}
requires extra $q^2 k=0(qkl)$ flops to compute
$\lambda \mathbf{D}^{T}\mathbf{D}$ in (\ref{GG2HC943039fd234moreStable}),
compared to \textbf{Algorithm 5} with   (\ref{W2AMHEWNH3495pseudoInv329ak}) replaced by
(\ref{W2AMHEWNH3495pseudoInv329aksimple}), while only the
$2cqk=0(qkl)$ flops to compute $({{\bf{H}}^T}{{\bf{A}}_k}) \times {{\bf{\bar W}}_k}$
in (\ref{W2AMHEWNH3495pseudoInv329ak})   is saved in (\ref{W2AMHEWNH3495pseudoInv329aksimple}).} in each update,
and so do the numerically more stable implementation of the proposed efficient BLS
and the existing efficient BLS.

To compute the initial ${{\mathbf{A}}_k^{\dagger}}$ and  $\mathbf{W}_k$ by (\ref{Generelized_inv_def1}) and (\ref{W2AY9845729}), respectively,
the proposed BLS based on the ridge inverse
requires
 the flops of
 $k^2 l+k^3$, $2k^2 l$ and $2klc$ to compute $({{\bf{A}}_k^T} \times {{\bf{A}}_k}+ \lambda {\bf{I}})^{-1}$,
    $( \cdots )^{-1} \times {{\bf{A}}_k^T} $   and  ${{\mathbf{A}}_k^{\dagger}} \times {{\bf{Y}}}$, respectively,
    and the corresponding total flops are~\footnote{In (\ref{FlopsMyInverse3209d})
 and the following (\ref{FlopsMyChol3292}), we assume $c=0(k)$, which  holds  in most cases since
$c=0(q)$   (i.e., (\ref{q20lc20qISAO312c})) and usually $q \le k$.}
       \begin{equation}\label{FlopsMyInverse3209d}
3k^2 l+k^3+2klc \approx 3k^2 l+k^3.
 \end{equation}
 As a comparison,
to compute the initial
${{\mathbf{F}}_k}$ and   $\mathbf{W}_k$ by   (\ref{L_m_def12431RidgeInvJun30}) and
(\ref{W2AMHEWNH3495initial}), respectively,  
the proposed efficient BLS based on the inverse Cholesky factor
requires the flops of $k^2 l+\frac{2}{3}k^3$, $2klc$ and $2k^2 c$
  to compute 
  the inverse Cholesky factorization ${{\bf{F}}_k}{{\bf{F}}_k^{T}}= ({{\bf{A}}_k^T} \times {{\bf{A}}_k}+ \lambda {\bf{I}})^{-1}$,
  $\mathbf{A}_{{k}}^{T} \times {{\mathbf{Y}}}$ and  ${{\bf{F}}_k} \times ({{\bf{F}}_k^{T}} \times \mathbf{A}_{{k}}^{T} {{\mathbf{Y}}})$, respectively, and
  the corresponding total flops are
     \begin{equation}\label{FlopsMyChol3292}
k^2 l+\frac{2}{3}k^3 + 2klc + 2k^2 c \approx k^2 l+\frac{2}{3}k^3.
 \end{equation}

From (\ref{FlopsMyInverse3209d})
 and
(\ref{FlopsMyChol3292}), we can conclude that with respect to the proposed efficient BLS based on the inverse Cholesky factor,
the proposed BLS based on the ridge inverse
  requires
  \begin{equation}\label{FlopRatioMyChol2MyInvPewodk234}
\frac{3k^2 l+k^3}{k^2 l+\frac{2}{3}k^3}= 3 - \frac{3}{2+3(l/k)}>\frac{12}{5}
 \end{equation}
times~\footnote{Here we consider the usual case with $l>k$.} of flops.
 Moreover, it can be seen that
 the flops of (\ref{FlopsMyInverse3209d}) and (\ref{FlopsMyChol3292}) are also required by
the original BLS~\cite{BL_trans_paper}  and the existing efficient BLS~\cite{BLSpaper2021zhf1}, respectively,
 to compute the initial intermediate result  and output weights.

\begin{table*}[!t]
\scriptsize
\renewcommand{\arraystretch}{1.3}
\newcommand{\tabincell}[2]{\begin{tabular}{@{}#1@{}}#2\end{tabular}}
\caption{{\footnotesize Snapshot Results of Testing Accuracy on NORB Dataset with the Ridge Parameter $\lambda= {10}^{-4},{10}^{-3},{10}^{-2},{10}^{-1}$}} \label{table_example} \centering
\begin{tabular}{c|c|c|c|c|c|c |c | c | c| c|c |c |}
 \hline
\multicolumn{4}{c|}{{\bfseries \tabincell{c}{  Number of}}}  &\multirow{3}*{2000}   & 2000  &  3500   &     5000     &    6500      &    8000     &   9500     &   11000    &    12500                   \\
\multicolumn{4}{c|}{{\bfseries \tabincell{c}{ Enhancement }}}    &   &  $\downarrow$   &     $\downarrow$     &    $\downarrow$      &    $\downarrow$     &   $\downarrow$     &   $\downarrow$    &    $\downarrow$     &    $\downarrow$        \\
\multicolumn{4}{c|}{{\bfseries \tabincell{c}{   Nodes}}}     &   &  3500   &     5000     &    6500      &    8000     &   9500     &   11000    &    12500     &    14000         \\
\hline
\hline
\multirow{8}*{\rotatebox{90}{{\bfseries   \tabincell{c}{Testing Accuracy ($\% $) \\ for $\lambda= {{10}^{-4}}$}} }}  &  \multirow{4}*{\rotatebox{90}{Gen. Inv. Sol. }}    & \multirow{2}*{\textbf{Orig.}}    &Mean      &     89.42    &   89.35    &   88.83    &   87.79    &   85.87    &   82.98    &   79.19    &   74.89    &   70.44         \\
\cdashline{4-13}
   &     &    &  Std    &      0.419    &    0.473    &    0.847    &    1.280    &    2.192    &    3.425    &    4.554    &    5.382    &    5.853         \\
\cline{3-13}
  &      & \multirow{2}*{\textbf{Exg-E}}    &Mean      &    89.42    &   89.35    &   88.83    &   87.79    &   85.87    &   82.98    &   79.19    &   74.89    &   70.44          \\
\cdashline{4-13}
   &     &    &  Std    &   0.419    &    0.473    &    0.847    &    1.280    &    2.192    &    3.425    &    4.553    &    5.381    &    5.853         \\
\cline{2-13}
  &\multirow{4}*{\rotatebox{90}{Ridge Sol.}}       &\multirow{2}*{\textbf{R-Inv}}    &Mean      &   89.42    &   89.76    &   89.81    &   89.89    &   89.96    &   90.03    &   90.06    &   90.07    &   90.10      \\
 \cdashline{4-13}
   &     &    &  Std    &     0.419    &    0.408    &    0.464    &    0.403    &    0.341    &    0.297    &    0.327    &    0.340    &    0.323       \\
\cline{3-13}
  &      & \textbf{Chol-1, Chol-2}     &Mean      &     89.42    &   89.76    &   89.81    &   89.89    &   89.96    &   90.03    &   90.06    &   90.07    &   90.10        \\
  \cdashline{4-13}
   &     & \& \textbf{D-Rdg}   &  Std    &     0.419    &    0.408    &    0.464    &    0.403    &    0.341    &    0.298    &    0.327    &    0.340    &    0.323        \\
\hline
\hline
\multirow{8}*{\rotatebox{90}{{\bfseries   \tabincell{c}{Testing Accuracy ($\% $) \\ for $\lambda= {{10}^{-3}}$}} }}  &  \multirow{4}*{\rotatebox{90}{Gen. Inv. Sol. }}    & \multirow{2}*{\textbf{Orig.}}    &Mean      &      89.49    &   \textbf{89.51}    &   89.46    &   88.84    &   87.76    &   86.00    &   83.62    &   80.79    &   77.66       \\
\cdashline{4-13}
   &     &    &  Std    &     0.380    &    0.433    &    0.562    &    1.013    &    1.901    &    3.146    &    4.545    &    5.928    &    7.168        \\
\cline{3-13}
  &      & \multirow{2}*{\textbf{Exg-E}}    &Mean      &      89.49    &   \textbf{89.51}    &   89.46    &   88.84    &   87.76    &   86.00    &   83.62    &   80.79    &   77.66          \\
\cdashline{4-13}
   &     &    &  Std    &    0.380    &    0.433    &    0.562    &    1.013    &    1.901    &    3.146    &    4.545    &    5.928    &    7.168       \\
\cline{2-13}
  &\multirow{4}*{\rotatebox{90}{Ridge Sol.}}       &\multirow{2}*{\textbf{R-Inv}}    &Mean      &     89.49    &   89.89    &   90.13    &   90.25    &   90.32    &   90.33    &   90.33    &   90.34    &   \textbf{90.38}      \\
 \cdashline{4-13}
   &     &    &  Std    &      0.380    &    0.314    &    0.288    &    0.289    &    0.292    &    0.290    &    0.268    &    0.254    &    0.238     \\
\cline{3-13}
  &      & \textbf{Chol-1, Chol-2}     &Mean      &      89.49    &   89.89    &   90.13    &   90.25    &   90.32    &   90.33    &   90.33    &   90.34    &   \textbf{90.38}        \\
  \cdashline{4-13}
   &     & \& \textbf{D-Rdg}   &  Std    &    0.380    &    0.314    &    0.288    &    0.289    &    0.292    &    0.290    &    0.268    &    0.254    &    0.238      \\
\hline
\hline
\rule{0pt}{10pt}
 \multirow{4}*{\rotatebox{90}{{\bfseries   \tabincell{c}{Testing \\ Accuracy ($\% $) \\ for $\lambda= {{10}^{-2}}$}} }}  & \multirow{2}*{\rotatebox{90}{Gen.}} & \multirow{2}*{{ \textbf{Orig.} \&   \textbf{Exg-E}   }}    &Mean      &   88.87    &   89.22    &   89.42    &    89.49    &   89.37    &   89.03    &   88.39    &   87.52    &   86.39        \\
\cdashline{4-13}
\rule{0pt}{10pt}
   &  &    &  Std    &  0.413    &    0.363    &    0.340    &    0.431    &    0.723    &    1.279    &    2.153    &    3.180    &    4.322      \\
\cline{2-13}
\rule{0pt}{10pt}
  & \multirow{2}*{\rotatebox{90}{Rdg.}}  &\textbf{Chol-1}, \textbf{Chol-2},          &Mean      &    88.87    &   89.34    &   89.59    &   89.80    &   89.93    &   90.00    &   90.07    &   90.13    &    90.18    \\
 \cdashline{4-13}
 \rule{0pt}{10pt}
   &  & \textbf{R-Inv} \& \textbf{D-Rdg}         &  Std    &   0.413    &    0.353    &    0.281    &    0.279    &    0.257    &    0.239    &    0.242    &    0.225    &    0.210   \\
\hline
\hline
\rule{0pt}{10pt}
 \multirow{4}*{\rotatebox{90}{{\bfseries   \tabincell{c}{Testing \\ Accuracy ($\% $) \\ for $\lambda= {{10}^{-1}}$}} }}  & \multirow{2}*{\rotatebox{90}{Gen.}} & \multirow{2}*{{ \textbf{Orig.} \&   \textbf{Exg-E}   }}    &Mean      &   86.73    &   88.05    &   88.56    &   88.86    &   89.01    &   89.09    &    89.11    &   89.01    &   88.86       \\
\cdashline{4-13}
\rule{0pt}{10pt}
   &  &    &  Std    &     0.819    &    0.469    &    0.313    &    0.288    &    0.313    &    0.461    &    0.706    &    1.095    &    1.657    \\
\cline{2-13}
\rule{0pt}{10pt}
  & \multirow{2}*{\rotatebox{90}{Rdg.}}  &\textbf{Chol-1}, \textbf{Chol-2},          &Mean      &    86.73    &   87.94    &   88.48    &   88.79    &   89.00    &   89.15    &   89.26    &   89.34    &    89.41    \\
 \cdashline{4-13}
 \rule{0pt}{10pt}
   &  & \textbf{R-Inv} \& \textbf{D-Rdg}         &  Std    &    0.819    &    0.476    &    0.320    &    0.279    &    0.250    &    0.247    &    0.221    &    0.197    &    0.190     \\
\hline
\end{tabular}
\end{table*}

\subsection{Numerical Experiments on  MNIST and NORB Datasets}
The MNIST dataset
 consists of $70000$ handwritten digital images partitioned into training and test sets of  $60000$ and $10000$  images, respectively.
As the simulations for Table  \Rmnum{4} in \cite{BL_trans_paper}, we set the initial network as $10 \times 6$
feature nodes and $3000$ enhancement nodes. Then in each update, we add  $10$ feature nodes,
 $750$ enhancement nodes only corresponding to the added feature nodes,
and  $1250$ additional enhancement nodes,
till we reach $170$ feature nodes and $25000$ enhancement nodes after 11 updates~\footnote{Notice that only 4 updates are included in Table  \Rmnum{4} of \cite{BL_trans_paper}.}.

On the other hand, the NORB dataset~\cite{Norb_dataSet}
 consists of $48600$ images with $2 \times 32 \times 32$ pixels each.
 The images
  belong to five distinct categories: (1) animals; (2) humans; (3) airplanes; (4) trucks; and (5) cars.
 Of these images, half are selected into the training set, and the other half form
 the test set.
  We
   set the initial network as $100 \times 10$
feature nodes, as the simulations for Table  \Rmnum{7} in \cite{BL_trans_paper}.
Starting from the initial network of
$2000$ enhancement nodes,
  we add
$1500$ enhancement nodes in each update, till
we reach $14000$ enhancement nodes after 8 updates.
\begin{table*}[!t]
\scriptsize
\renewcommand{\arraystretch}{1.3}
\setlength\tabcolsep{3.5pt}
\newcommand{\tabincell}[2]{\begin{tabular}{@{}#1@{}}#2\end{tabular}}
\caption{Snapshot Results of Training Time on MNIST Dataset  and the Corresponding Speedups} \label{table_example} \centering
\begin{tabular}{|c|c||c c c c c| c c |c c c c c|c c|}
\hline
{\bfseries   No. of }         & {\bfseries   No. of}
      &\multicolumn{7}{c|}{{\bfseries   Additional  Training Time (seconds)}}
       &\multicolumn{7}{c|}{{\bfseries   Accumulative  Training Time (seconds) }} \\
{\bfseries   Feature  }         & {\bfseries   Enhancement}
      &\multicolumn{5}{c|}{{\bfseries   \tabincell{c}{Time for Each Update}}}
      &\multicolumn{2}{c|}{{\bfseries   \tabincell{c}{Speedups  of \textbf{Chol-1}}}}
       &\multicolumn{5}{c|}{{\bfseries   \tabincell{c}{Time for Each Update}}}
      &\multicolumn{2}{c|}{{\bfseries   \tabincell{c}{Speedups of \textbf{Chol-1}}}}  \\
{\bfseries  Nodes }         & {\bfseries   Nodes }      &\textbf{Orig.}  &\textbf{R-Inv} &\textbf{Exg-E}  &\textbf{Chol-2}  &\textbf{Chol-1}       &to \textbf{Orig.}   &to \textbf{Exg-E}     &\textbf{Orig.}  &\textbf{R-Inv} &\textbf{Exg-E}  &\textbf{Chol-2}  &\textbf{Chol-1}  &to \textbf{Orig.}   &to \textbf{Exg-E}      \\
\hline
\bfseries 60  & \bfseries 3000   &   9.56    &    9.58    &    4.93    &    4.93    &    4.92    &    1.94    &        &    9.56    &    9.58    &    4.93    &    4.93    &    4.92    &    1.94    &       \\
\hdashline
\bfseries  $\to$ 70  & \bfseries  $\to$ 5000   &  15.33    &   15.47    &    8.77    &    8.89    &    5.53    &    2.77    &    1.59    &   24.89    &   25.05    &   13.70    &   13.81    &   10.45    &    2.38    &    1.31  \\
\hdashline
\bfseries  $\to$ 80  & \bfseries  $\to$ 7000   & 22.62    &   22.72    &   13.85    &   13.97    &    8.52    &    2.65    &    1.63    &   47.51    &   47.77    &   27.55    &   27.79    &   18.97    &    2.50    &    1.45  \\
\hdashline
\bfseries  $\to$ 90  & \bfseries  $\to$  9000  &  29.69    &   29.80    &   19.39    &   19.57    &   11.71    &    2.54    &    1.66    &   77.20    &   77.57    &   46.94    &   47.36    &   30.68    &    2.52    &    1.53   \\
\hdashline
\bfseries  $\to$ 100  & \bfseries  $\to$ 11000   &  36.91    &   37.09    &   25.43    &   25.59    &   15.50    &    2.38    &    1.64    &  114.12    &  114.66    &   72.37    &   72.95    &   46.18    &    2.47    &    1.57  \\
\hdashline
\bfseries  $\to$ 110  & \bfseries  $\to$ 13000   &    43.81    &   44.19    &   31.47    &   31.65    &   19.54    &    2.24    &    1.61    &  157.93    &  158.85    &  103.84    &  104.60    &   65.72    &    2.40    &    1.58  \\
\hdashline
\bfseries  $\to$ 120  & \bfseries  $\to$  15000  &  51.04    &   51.49    &   37.99    &   38.25    &   24.03    &    2.12    &    1.58    &  208.97    &  210.34    &  141.82    &  142.85    &   89.75    &    2.33    &    1.58   \\
\hdashline
\bfseries  $\to$ 130  & \bfseries  $\to$ 17000   &  57.73    &   58.24    &   44.53    &   44.75    &   28.49    &    2.03    &    1.56    &  266.70    &  268.58    &  186.36    &  187.59    &  118.24    &    2.26    &    1.58  \\
\hdashline
\bfseries  $\to$ 140  & \bfseries  $\to$ 19000   &  65.27    &   65.75    &   51.87    &   52.26    &   33.81    &    1.93    &    1.53    &  331.96    &  334.34    &  238.23    &  239.86    &  152.05    &    2.18    &    1.57   \\
\hdashline
\bfseries  $\to$ 150  & \bfseries  $\to$ 21000   &  71.70    &   72.51    &   59.03    &   59.39    &   38.85    &    1.85    &    1.52    &  403.66    &  406.84    &  297.25    &  299.24    &  190.90    &    2.11    &    1.56  \\
\hdashline
\bfseries  $\to$ 160  & \bfseries  $\to$ 23000   &   79.13    &   80.00    &   67.12    &   67.33    &   44.83    &    1.77    &    1.50    &  482.79    &  486.84    &  364.37    &  366.57    &  235.74    &    2.05    &    1.55  \\
\hdashline
\bfseries  $\to$ 170  & \bfseries  $\to$ 25000   &   85.82    &   87.00    &   74.83    &   75.07    &   50.44    &    1.70    &    1.48    &  568.61    &  573.84    &  439.21    &  441.64    &  286.17    &    1.99    &    1.53  \\
\hline
\end{tabular}
\end{table*}

\begin{figure*}[!t]
    \centering
    \subfloat[MNIST
dataset.]{
        \hspace*{-.1in}
        \includegraphics[scale=0.56]{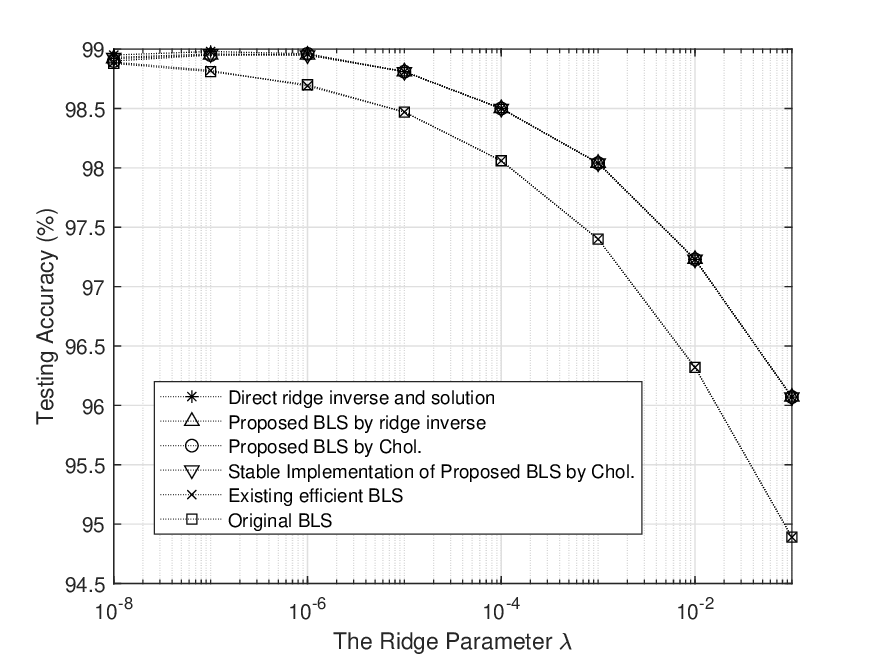}
        \label{p_D_req_100}
    }
        \subfloat[NORB dataset.]{
        \hspace*{-.1in}
        \includegraphics[scale=0.56]{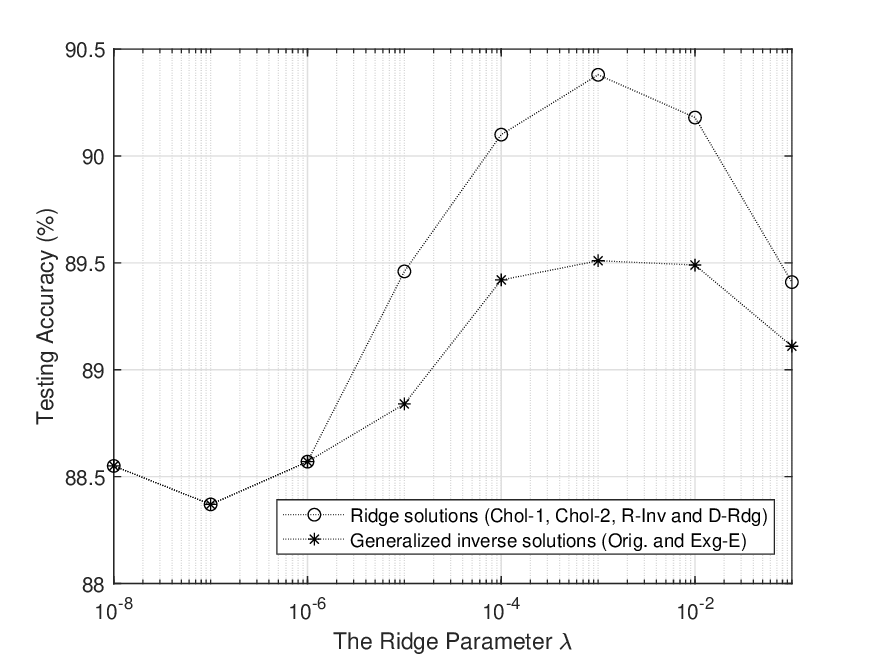}
        \label{p_D_req_100}
    }
\caption{The maximum testing accuracies achieved under different ridge parameters on MNIST and NORB datasets.}
\label{zhftry1}
\end{figure*}

For each update,
 the testing accuracies
of the presented BLS
algorithms on the MNIST and NORB datasets  are shown in Tables \Rmnum{2}, \Rmnum{3} and Tables \Rmnum{4},  \Rmnum{5},
respectively, where we list the mean and standard deviation of 100 simulations. Moreover,
Table \Rmnum{6} lists the training time of the presented BLS
algorithms on the MNIST dataset, which is the average value of  $500$ simulations.
In Tables \Rmnum{2}-\Rmnum{6} and
the rest of this section,
  \textbf{D-Rdg} denotes 
 the direct ridge inverse and solution (by (\ref{Generelized_inv_def1}) and
(\ref{W2AY9845729})), while the abbreviations \textbf{Orig.}, \textbf{Exg-E},
\textbf{Chol-1}, \textbf{Chol-2} and \textbf{R-Inv}
have been defined in the previous sections.

When
 the ridge parameter $\lambda \le {{10}^{-6}}$
in Table \Rmnum{2} and
 $\lambda \le {{10}^{-5}}$
in Table \Rmnum{4},
the
generalized inverse solutions
\textbf{Orig.} and  \textbf{Exg-E}
sometimes achieve different testing accuracies due to numerical errors,
and so do  the ridge solutions
\textbf{Chol-1},
\textbf{Chol-2},
\textbf{R-Inv}
and
\textbf{D-Rdg}.
Then Tables \Rmnum{2} and \Rmnum{4}  lists the results for each BLS algorithm.
 Moreover,
 the ridge parameter
 satisfies
 ${{10}^{-5}}  \le \lambda \le {{10}^{-1}}$ and ${{10}^{-4}}  \le \lambda \le {{10}^{-1}}$ in Tables \Rmnum{3} and \Rmnum{5}, respectively.
Our simulations for Table \Rmnum{3} show that
 \textbf{Chol-1},
\textbf{Chol-2}
and
\textbf{R-Inv}
always achieve the same  testing accuracy~\footnote{
A sufficiently big $\lambda$ can assure the positive definiteness of
${\bf{H}}^{T}{{\bf{H}}}+\lambda \mathbf{I} - {{\mathbf{ H}}^T}{{\mathbf{ A}}_k}{{\bf{F}}_k}{{\bf{F}}_k^{T}}{{{\mathbf{ A}}_k^T}{{\mathbf{ H}}}}$
in
(\ref{ZF_def_L_2_items3aJun2021})  and
${{\mathbf{C}_{{}}^{T}\mathbf{C}}}+\lambda \mathbf{D}^{T}\mathbf{D}+\lambda \mathbf{I}$
in (\ref{GG2HC943039fd234moreStable}) and (\ref{B2HCnumdaICt2021JunSmallLamda}),
 which can reduces the numerical errors caused by
 the corresponding inverse Cholesky factorization or inverse.} as \textbf{D-Rdg} when $\lambda$
 is not less than
 ${{10}^{-5}}$,  ${{10}^{-4}}$ and ${{10}^{-3}}$, respectively,
 and  \textbf{Orig.} always achieve the same  testing accuracy as   \textbf{Exg-E}
 when $\lambda \ge {{10}^{-2}}$.
 On the other hand, our simulations for Table \Rmnum{5} show that
 \textbf{Chol-1} and
\textbf{Chol-2}
always achieve the same  testing accuracy as \textbf{D-Rdg} when $\lambda \ge {{10}^{-4}}$,
 and when $\lambda \ge {{10}^{-2}}$, both the generalized inverse solutions achieve the same testing accuracy,
 and so do all the ridge solutions.
  Then when multiple different BLS algorithms get the same testing accuracy, the testing accuracy is listed only once in Tables \Rmnum{3} and \Rmnum{5}
 for simplicity.

 When the ridge parameter $\lambda$ is too small, our simulations on the MNIST dataset show that accumulation of numerical errors may  make
 ${{\bf{H}}^{T}{{\bf{H}}}+\lambda \mathbf{I}}- {{\mathbf{ H}}^T}{{\mathbf{ A}}_k}{{\bf{F}}_k}{{\bf{F}}_k^{T}}{{{\mathbf{ A}}_k^T}{{\mathbf{ H}}}}$ in (\ref{ZF_def_L_2_items3aJun2021})
 no longer positive definite, resulting in the unavailable inverse Cholesky factor ${{\mathbf{G}}}$.
Accordingly,
 Table \Rmnum{2} shows that
the testing accuracy is not available (i.e., NA) for
\textbf{Chol-1}
 from the $5^{th}$ and  $8^{th}$
 update when $\lambda$ is ${{10}^{-8}}$ and ${{10}^{-7}}$, respectively.



Table \Rmnum{6} lists the training time of \textbf{Orig.},   \textbf{Exg-E},
\textbf{Chol-1},   \textbf{Chol-2} and  \textbf{R-Inv},
and shows the speedups in training time of
the proposed  efficient  ridge solution (i.e., \textbf{Chol-1})
to the original BLS~\cite{BL_trans_paper} (i.e., \textbf{Orig.})
 and the existing efficient BLS~\cite{BLSpaper2021zhf1} (i.e., \textbf{Exg-E}),
 respectively. The speedup in Table \Rmnum{6} is
the  ratio between the training time of the existing BLS and that of the proposed efficient ridge solution.
As observed from Table \Rmnum{4},
 the proposed ridge solution based on the ridge
inverse (i.e., \textbf{R-Inv}) and the numerically more stable implementation of the proposed efficient ridge solution
(i.e., \textbf{Chol-2})
require nearly the same training time as
the original BLS (i.e., \textbf{Orig.})
and the existing efficient BLS  (i.e., \textbf{Exg-E}), respectively,
while the speedups in total training time
  of the proposed efficient ridge solution (i.e., \textbf{Chol-1}) to the original BLS
   and the existing efficient BLS
  are $1.99 \sim 2.52$
    and $1.31 \sim 1.58$,  respectively.
  It can be seen that the above comparisons of training time
  are  consistent with the
   theoretical
calculation of flops  in the last subsection.

\subsection{Analysis of the Maximum Testing Accuracies Achieved under Different Ridge Parameters}

Based on the  testing accuracies in Tables \Rmnum{2}-\Rmnum{5},
we show the maximum testing accuracies achieved under different ridge parameters on MNIST and NORB datasets in Fig. 1.
On the NORB dataset, both the generalized inverse solutions achieve the same maximum testing accuracies, and so do all the ridge solutions.
Then for simplicity, Fig. 1.b only shows the maximum testing accuracies for the generalized inverse solutions and those for the  ridge solutions.

From Tables \Rmnum{2},  \Rmnum{3} and Fig. 1.a, it can be seen that
 on the MNIST dataset,
 the existing generalized inverse solutions  \textbf{Orig.} and   \textbf{Exg-E}
achieve the maximum testing accuracy of $98.88\% \sim 98.89\%$ when $\lambda = {{10}^{-8}}$,
while the proposed ridge solutions \textbf{Chol-1},   \textbf{Chol-2} and  \textbf{R-Inv}
achieve the maximum testing accuracy of
$98.95\% \sim 98.96\%$ when $\lambda = {{10}^{-7}}, {{10}^{-6}}$.
On the other hand,  it can be seen from Tables \Rmnum{4},  \Rmnum{5} and Fig. 1.b
that on the NORB dataset,
 the existing generalized inverse solutions  (i.e., \textbf{Orig.} and   \textbf{Exg-E})
and the proposed ridge solutions (i.e., \textbf{Chol-1},   \textbf{Chol-2} and  \textbf{R-Inv})
achieve the maximum testing accuracies of $89.51\%$ and $90.38\%$, respectively,  when $\lambda = {{10}^{-3}}$.
Accordingly, it can be concluded that
the proposed ridge solutions
achieve better testing accuracies than
the existing generalized inverse solutions.


\subsection{Implementation Aspects of  the Efficient Inverse
Cholesky Factorization: Parallelization in Distributed Systems, Memory Saving, Square-Root Free and Division Free}

The efficient inverse Cholesky factorization~\cite{my_inv_chol_paper}
updates the inverse Cholesky factor of  an $(i-1) \times (i-1)$ Hermitian matrix  ${{\bf{R}}_{i-1}}$ into that of
the $i \times i$ Hermitian matrix ${{\bf{R}}_{i}}$, which includes ${{\bf{R}}_{i-1}}$  as the $(i-1) \times (i-1)$ leading principal submatrix.
Then to develop an efficient  solution for the incremental Broad
Learning System (BLS) on added nodes~\cite{BL_trans_paper},
  the efficient inverse Cholesky factorization is extended  in this paper to
update the inverse Cholesky factor of  an $i \times i$ Hermitian matrix  ${{\bf{R}}_{i}}$ into
that of the $k \times k$ Hermitian matrix  ${{\bf{R}}_{k}}$ through only $1$ iteration,
where $k>i$, and  ${{\bf{R}}_{k}}$
includes ${{\bf{R}}_{i}}$  as the $i \times i$ leading principal submatrix.
In this subsection, we  introduce the implementation aspects of  the above-described efficient inverse
Cholesky factorization. In Appendix D, we
develop the memory-saving parallel implementation of the inverse Cholesky factorization for
MIMD (Multiple Instruction and Multiple Data Stream) computers~\cite{ParallelInverseMatrix320sd3s}
with multiple processors.

In the inverse Cholesky factorization, there are square-root and division operations
 which
are time-consuming
 for conventional multiply-add digital computers
\cite{CORDICandGivens},
and require high bit precision and quite a few clock
cycles in fixed-point implementations~\cite{WCNCcholesky,kjLiuGIVENS}.
Accordingly, we develop the memory-saving parallel implementation of
the inverse LDL$^T$ factorization in Appendix E, which can avoid the square-root  operations.
Furthermore, we also develop the memory-saving parallel implementation of
the division free inverse LDL$^T$ factorization in Appendix F.




In Appendix E,
 we give some analysis about the parallel performance on MIMD
  computers~\cite{ParallelInverseMatrix320sd3s} of the proposed parallel implementation of
  the inverse LDL$^T$ factorization, and
 we  describe the application of
 the inverse LDL$^T$ factorization to the distributed BLS.
 Notice that only for the inverse LDL$^T$ factorization,
 we give the analysis of the parallel performance
 and describe  the application to the distributed BLS.

   The above-described memory-saving parallel implementations of the inverse Cholesky factorization,
   the inverse LDL$^T$ factorization and the division free inverse LDL$^T$ factorization
     compute the inverse Cholesky factor,  the inverse LDL$^T$ factors and
   the division free inverse LDL$^T$ factors of a Hermitian matrix ${{\bf{R}}_k} \in {\Re ^{k \times k}}$
   by $k$ iterations.
In the theoretical situation where the
parallel algorithm runs on $1+2+ \cdots + k=\frac{(k+1)k}{2}$ processors, i.e., each entry in the upper-triangular part of ${{\bf{R}}_k}$ is stored in an exclusive processor,
the proposed parallel implementations update
 all entries in columns $i+1$ to $k$ of the upper-triangular part of ${{\bf{R}}_k}$ except row $i$
 simultaneously in the $i^{th}$ ($i=1,2, \cdots, k$) iteration, and each entry
 is updated with
a multiplication and an addition executed serially.
In the situation where the
parallel algorithm runs on $k$ processors, and each column of the upper-triangular part  of  ${{\bf{R}}_k}$ is stored in an exclusive processor,
${\bf{R}}{{(1:i,i)}}, {\bf{R}}(i,i+1), {\bf{R}}(i,i+2),\cdots,{\bf{R}}(i,j-1)$
    need to be transmitted in the $i^{th}$ ($i=1,2, \cdots, k$) iteration to  the processor storing the $j^{th}$ column of  ${\bf{R}}$
    from the processors storing the $i^{th}, (i+1)^{th}, (i+2)^{th}, \cdots, (j-1)^{th}$ columns of ${\bf{R}}$,
    respectively,  where $j>i$.
In the general case where the columns of ${\bf{R}}$ are stored in $\tau \le k$ processors,
and processors $1,2,\cdots,\tau$
store
    ${\upsilon _1}, {\upsilon _2}, \cdots,  {\upsilon _\tau}$
    columns of ${\bf{R}}$, respectively,
 ${\bf{R}}{{(1:i,i)}}$ and ${\bf{R}}(i,i+1:{{\tilde \upsilon} _\mu})$  in  processor $\mu$ need to be transmitted to processors $\mu+1,\mu+2,\cdots,\tau$,
    while
   ${\bf{R}}(i,{{\tilde \upsilon} _{\varsigma-1}}+1:{{\tilde \upsilon} _{\varsigma}})$ in  processor $\varsigma$ needs to be  transmitted to processors $\varsigma+1,\varsigma+2,\cdots,\tau$,
   where $\varsigma=\mu+1,\mu+2,\cdots,\tau-1$.
   The above-mentioned general case of the proposed parallel implementations can be applied in the
distributed BLS with model-parallelism.

\section{Conclusion}

In the original BLS algorithm~\cite{BL_trans_paper},
the generalized
inverse solution for the output weights is computed
from
the generalized
inverse,
which is updated efficiently
by
     the Greville's method~\cite{cite_general_inv_book},
and  approximated by ridge regression, i.e.,
the ridge inverse with the ridge parameter $\lambda \to 0$ assumed.
Then an efficient implementation of
 the original BLS on added nodes
 was proposed in \cite{BLSpaper2021zhf1}, where
 $\lambda \to 0$ is still assumed.
 In this paper,
   we propose two ridge solutions
   for the BLS
   on added nodes,
        where  the assumption of $\lambda \to 0$  is no longer required, and  $\lambda$ can be  any positive real number.

     In  the proposed ridge solution based on the ridge inverse,
     a sub-matrix in the ridge inverse is utilized  to update the
     output weights, and the ridge inverse is updated
     by extending the Greville's method that computes the generalized inverse of a partitioned matrix.
 The proposed algorithm to
 update the
 ridge inverse
 and the Greville's method
 are equivalent when $\lambda \to 0$,
  but are usually different when $\lambda \to 0$ is not satisfied.
  So  the proposed algorithm to update the ridge inverse
  includes the Greville's method as a special case,
  and
   expands the application range of the Greville's method
     from the generalized inverse to the ridge inverse.
   On the other hand,
the proposed efficient ridge solution based on the inverse Cholesky factor
utilizes the sub-matrices in the inverse Cholesky factor to update the
     output weights, and updates the inverse Cholesky factor by
     an efficient inverse Cholesky factorization for a Hermitian matrix partitioned into $2 \times 2$ blocks,
     which extends
      the corresponding algorithm in \cite{my_inv_chol_paper}.
      Moreover, we also develop another implementation of  the proposed efficient ridge solution, which is numerically more stable when the ridge parameter $\lambda$
      is very small.


   The proposed ridge solution based on the ridge inverse and the numerically more stable implementation of the proposed efficient ridge solution
     require the same dominant flops as the original BLS algorithm and the existing efficient generalized inverse solution, respectively.
 Compared to the proposed efficient ridge solution
  based on the inverse Cholesky factor,
   in each update the original BLS algorithm
       requires about $3$
   times of
     flops,
     and  the existing efficient generalized inverse solution usually  requires more than  $\frac{5}{3}$
times of flops. In the numerical experiments,
    the speedups
  of the proposed efficient ridge solution to the original BLS
   and the existing efficient
   generalized
inverse solution
  are $1.99 \sim 2.52$
    and $1.31 \sim 1.58$,  respectively, when the total training time is considered.

    The numerical experiments on MNIST and NORB datasets
    show that   both the proposed ridge solutions
   for  BLS achieve better testing accuracies than the original BLS and the existing efficient generalized inverse solution for BLS.
  The  proposed ridge solution based on the ridge inverse, the proposed efficient ridge solution and its numerically more stable implementation
 all achieve the maximum testing accuracy of  $98.95\% \sim 98.96\%$ on the MNIST dataset when $\lambda = {{10}^{-7}}, {{10}^{-6}}$,
 and achieve the maximum testing accuracy of $90.38\%$
  on the NORB dataset   when $\lambda = {{10}^{-3}}$.
   As a comparison,  the original BLS and the existing efficient generalized inverse solution
   both achieve the maximum testing accuracy of $98.88\% \sim 98.89\%$ on the MNIST dataset when $\lambda = {{10}^{-8}}$,
   and achieve the maximum testing accuracy of $89.51\%$  on the NORB dataset when $\lambda = {{10}^{-3}}$.

\appendices
\section{The Derivation of
(\ref{ZF_def_L_2_items3abJun2021})}
From (\ref{L_big_BLK_def1}) we obtain
\begin{small}
\begin{equation}\label{LinvDef12356}{{\bf{F}}_{k+q}^{-1}}=\left[ \begin{matrix}
   {{\bf{F}}_k^{-1}} & -{{\bf{F}}_k^{-1}}\mathbf{T}{{\mathbf{G}}^{-1}}  \\
   \mathbf{0} & {{\mathbf{G}}^{-1}}  \\
\end{matrix} \right],
\end{equation}
\end{small}
which is substituted into (\ref{L_m_def12431}) to obtain
\begin{small}
 \begin{gather}
{{{{\bf{R}}_{k+q}}}} =
{{\left[ \begin{matrix}
   {{\bf{F}}_k^{-1}} & -{{\bf{F}}_k^{-1}}\mathbf{T}{{\mathbf{G}}^{-1}}  \\
   \mathbf{0} & {{\mathbf{G}}^{-1}}  \\
\end{matrix} \right]}^{T}}
\left[ \begin{matrix}
   {{\bf{F}}_k^{-1}} & -{{\bf{F}}_k^{-1}}\mathbf{T}{{\mathbf{G}}^{-1}}  \\
   \mathbf{0} & {{\mathbf{G}}^{-1}}  \\
\end{matrix} \right] \notag \\
\left[ {\begin{array}{*{20}{c}}
{{{\bf{F}}_k^{ - T}}
{{\bf{F}}_k^{ - 1}}}&{ \begin{array}{l}
 - {{\bf{F}}_k^{ - T}}
{{\bf{F}}_k^{ - 1}}{\mathbf{T}}{{\bf{G}}^{ - 1}}
\end{array} }\\
{\left( \begin{array}{l}
 - {{\bf{G}}^{ - T}}{{\mathbf{T}}^T}  \\
\times {{\bf{F}}_k^{ - T}}{{\bf{F}}_k^{ - 1}}
\end{array} \right)}&{\left( \begin{array}{l}
{{\bf{G}}^{ - T}}{{\mathbf{T}}^T}{{\bf{F}}_k^{ - T}}
  {{\bf{F}}_k^{ - 1}} \times \\
 {\mathbf{T}}{{\bf{G}}^{ - 1}} + {{\bf{G}}^{ - T}}{{\bf{G}}^{ - 1}}
\end{array} \right)}
\end{array}} \right]. \label{Rbig2FFTG8392832a}
\end{gather}
\end{small}
 Then (\ref{Rbig2FFTG8392832a}) can be compared
 with
(\ref{R_def_perhaps_No_R1})
to deduce
\begin{small}
\begin{subnumcases}{\label{110_111}}
-{{\bf{F}}_k^{-T}}{{\bf{F}}_k^{-1}}\mathbf{T}{{\mathbf{G}}^{-1}}
={{{\mathbf{ A}}_k^T}{{\mathbf{ H}}}} &  \label{110}\\
{{\bf{G}}^{ - T}}{{\mathbf{T}}^T}{{\bf{F}}_k^{ - T}}{{\bf{F}}_k^{ - 1}}{\mathbf{T}}{{\bf{G}}^{ - 1}}
 + {\bf{G}}^{ - T}{\bf{G}}^{ - 1} = {\bf{H}}^{T}{{\bf{H}}}+\lambda \mathbf{I}. & \label{111}
\end{subnumcases}
\end{small}

We can deduce
 (\ref{ZF_def_L_2_items3bJun2021})
  from (\ref{110}).
   On the other hand, 
let us substitute
 (\ref{ZF_def_L_2_items3bJun2021})
 into (\ref{111}) to obtain
\begin{small}
\begin{gather}
{{\mathbf{G}}^{-T}}{{\mathbf{G}}^{-1}} = {{\bf{H}}^{T}{{\bf{H}}}+\lambda \mathbf{I}} - {{\mathbf{G}}^{-T}}{{({{\bf{F}}_k}{{\bf{F}}_k^{T}}{{{\mathbf{ A}}_k^T}{{\mathbf{ H}}}}\mathbf{G})}^{T}}
{{\bf{F}}_k^{-T}}{{\bf{F}}_k^{-1}}
 \times \notag \\
{{\bf{F}}_k}
{{\bf{F}}_k^{T}} {{\mathbf{ A}}_k^T}{{\mathbf{ H}}}
\mathbf{G}{{\mathbf{G}}^{-1}} ={{\bf{H}}^{T}{{\bf{H}}}+\lambda \mathbf{I}}- {{\mathbf{ H}}^T}{{\mathbf{ A}}_k}{{\bf{F}}_k}{{\bf{F}}_k^{T}}{{{\mathbf{ A}}_k^T}{{\mathbf{ H}}}},  \notag
\end{gather}
\end{small}
from which
we can
deduce
(\ref{ZF_def_L_2_items3aJun2021}).

\section{The Derivation of (\ref{DACdifferent23412Equal999})}
 Substitute
 (\ref{C_matrix_def_111}) and (\ref{D_matrix_def_111})
   into
      the left side of  (\ref{DACdifferent23412Equal999})
  successively
  to obtain
\begin{align}
{{\mathbf{ D}}^{T}}{{\bf{ A}}_k^T}\mathbf{ C} & ={{\mathbf{ D}}^{T}}{{\bf{ A}}_k^T}({\bf{H}}-{\mathbf{ A}_k}\mathbf{ D})  \notag \\
& ={{\mathbf{D}}^{T}}{{\bf{ A}}_k^T}({\bf{H}}-{\mathbf{ A}_k}{{\bf{ A}}_k^\dag}{{\bf{H}}}).  \label{HAAHminusAAH7656}
\end{align}
Also substitute~\cite{InverseSumofMatrix8312}[Eqn. (20)]
  \begin{equation}\label{AAtNumdaAtANumdaEqu12}
  {{\left( {{{{\mathbf{A}}}}^{T}}{{\mathbf{A}}}+\lambda \mathbf{I} \right)}^{-1}}{{{{\mathbf{A}}}}^{T}}={{{{\mathbf{A}}}}^{T}}{{\left( {{\mathbf{A}}}{{{{\mathbf{A}}}}^{T}}+\lambda \mathbf{I} \right)}^{-1}}
 \end{equation}
     into
     (\ref{Generelized_inv_def1})
      to get
        another form of
        (\ref{Generelized_inv_def1}),
   i.e.,
      \begin{equation}\label{AinvLimNumda0AAiA1221extend3920new211}
\mathbf{A}^{\dagger}= {{{{\mathbf{A}}}}^{T}}{{\left( {{\mathbf{A}}}{{{{\mathbf{A}}}}^{T}}+\lambda \mathbf{I} \right)}^{-1}},
\end{equation}
which is then 
 substituted into  (\ref{HAAHminusAAH7656})  to obtain
 \begin{gather}
{{\mathbf{ D}}^{T}}{{\bf{ A}}_k^T}\mathbf{ C}={{\mathbf{D}}^{T}}{{\bf{ A}}_k^T}
\left( {\bf{H}}-{\mathbf{ A}_k}{{\bf{ A}}_k^T}{{\left( {\mathbf{ A}_k}{{\bf{ A}}_k^T}+\lambda \mathbf{I} \right)}^{-1}}\bf{H} \right)  \notag \\
={{\mathbf{D}}^{T}}{{\bf{ A}}_k^T}
\left( {\bf{H}}-\left( \mathbf{I}-\lambda {{\left( {\mathbf{ A}_k}{{\bf{ A}}_k^T}+\lambda \mathbf{I} \right)}^{-1}} \right)\bf{H} \right) \notag \\
=\lambda {{\mathbf{D}}^{T}}{{\bf{ A}}_k^T}{{\left( {\mathbf{ A}_k}{{\bf{ A}}_k^T}+\lambda \mathbf{I} \right)}^{-1}}\bf{H}. \label{LamdaAHAAAI83209}
\end{gather}
To deduce
(\ref{DACdifferent23412Equal999}) finally, we substitute (\ref{AinvLimNumda0AAiA1221extend3920new211}) and  (\ref{D_matrix_def_111})
into (\ref{LamdaAHAAAI83209}) successively to
 obtain
${{\mathbf{ D}}^{T}}{{\bf{ A}}_k^T}\mathbf{ C}=\lambda {{\mathbf{D}}^{T}} {{\bf{ A}}_k^\dag}{\bf{H}}=\lambda {{\mathbf{ D}}^{T}} {\mathbf{D}}$.

\section{To Verify that  (\ref{B2HCnumdaICt2021JunSmallLamda})
is Equal to
  (\ref{B_Matrix_def1abOrig}) when
  $\lambda \to 0$}


When  ${\bf{\bar C}} \ne \mathbf{0}$,
we need to verify that  ${{\mathbf{B}}^{T}}$ in  (\ref{B2HCnumdaICt2021JunSmallLamda})
is equal to
${{\mathbf{\bar B}}^{T}}$ in
 (\ref{B_Matrix_def1aOrig}) if   $\lambda \to 0$.
We can let  $\lambda \to 0$ in
  (\ref{B2HCnumdaICt2021JunSmallLamda}) 
  to write
 \begin{multline}\label{Lamda0BtASiewkd0321932}
\underset{\lambda \to 0}{\mathop{\lim }}\,{{{\mathbf{B}}^{T}}} = \underset{\lambda \to 0}{\mathop{\lim }}\,{{{{\left({{\mathbf{C}_{{}}^{T}\mathbf{C}}}+\lambda \mathbf{D}^{T}\mathbf{D}+\lambda \mathbf{I} \right)}^{-1}}{{\mathbf{C}}^{T}}}} \\ = {{{{\left({{\mathbf{\bar C}_{{}}^{T}\mathbf{\bar C}}} \right)}^{-1}}{{\mathbf{\bar C}}^{T}}}}= {{{\mathbf{\bar B}}^{T}}},
\end{multline}
where
$\underset{\lambda \to 0}{\mathop{\lim }}\,{\mathbf{C}}={\bf{\bar C}}$
 and (\ref{B_Matrix_def1aOrig}) are utilized successively.

When  ${\bf{\bar C}} = \mathbf{0}$,
we need to verify that  ${{\mathbf{B}}^{T}}$ in  (\ref{B2HCnumdaICt2021JunSmallLamda})
is equal to
${{\mathbf{\bar B}}^{T}}$
in  (\ref{B_Matrix_def1bOrig}) if   $\lambda \to 0$.
Firstly, we use
 ${\bf{\bar C}} = \mathbf{0}$ to write  ${\bf{C}}$  defined by (\ref{C_matrix_def_111})
in another form.
Let ${\bf{\bar C}} = \mathbf{0}$  in (\ref{C_matrix_def_111Orig}) to obtain
${{\bf{H}}}={\mathbf{ A}_k}\mathbf{\bar D}$,
which is
substituted into (\ref{C_matrix_def_111}) to obtain
\begin{align}
\mathbf{C}&={\mathbf{ A}_k}\mathbf{\bar D}-{\mathbf{ A}_k}\mathbf{D} ={\mathbf{ A}_k}{{\bf{ A}}_k^+}{{\bf{H}}}-{\mathbf{ A}_k}{{\bf{ A}}_k^\dag}{{\bf{H}}}   \notag \\
&={\mathbf{ A}_k}{{(\mathbf{A}_{{k}}^{T}{\mathbf{ A}_k})}^{-1}}\mathbf{A}_{{k}}^{T}{{\bf{H}}}-{\mathbf{ A}_k}{{(\mathbf{A}_{{k}}^{T}{\mathbf{ A}_k}+\lambda \mathbf{I})}^{-1}}\mathbf{A}_{{k}}^{T}{{\bf{H}}}  \notag \\
&={\mathbf{ A}_k}\left( {{(\mathbf{A}_{{k}}^{T}{\mathbf{ A}_k})}^{-1}} -{{(\mathbf{A}_{{k}}^{T}{\mathbf{ A}_k}+\lambda \mathbf{I})}^{-1}}  \right)  \mathbf{A}_{{k}}^{T}{{\bf{H}}},  \label{C2dseqf32vdasa32}
\end{align}
where (\ref{D_matrix_def_111Orig}),
(\ref{D_matrix_def_111}),
(\ref{Ainv2AtAinvAt9096}) and
(\ref{Generelized_inv_def1})
are utilized.
To  simplify (\ref{C2dseqf32vdasa32}),
 we utilize a special case of the Woodbury matrix identity~\cite[equation (68)]{LectureNotesWoodbury}, i.e.,
  \begin{equation}\label{WoodburySpecial320kds320as}
{(\pmb{\Psi} + \pmb{\Lambda})^{ - 1}} = {\pmb{\Psi}^{ - 1}} - {\pmb{\Psi}^{ - 1}}{({\pmb{\Psi}^{ - 1}} + {\pmb{\Lambda}^{ - 1}})^{ - 1}}{\pmb{\Psi}^{ - 1}},
 \end{equation}
to write  ${{(\mathbf{A}^{T}{\mathbf{ A}}+\lambda \mathbf{I})}^{-1}}$
in (\ref{C2dseqf32vdasa32})  as
\begin{align}
&{{(\mathbf{A}_{{}}^{T}\mathbf{A}+\lambda \mathbf{I})}^{-1}}  \notag \\
&= (\mathbf{A}_{{}}^{T}\mathbf{A})^{ - 1}
 - (\mathbf{A}_{{}}^{T}\mathbf{A})^{ - 1}{\left((\mathbf{A}_{{}}^{T}\mathbf{A})^{ - 1} + {\lambda}^{ - 1} \mathbf{I}\right)^{ - 1}}(\mathbf{A}_{{}}^{T}\mathbf{A})^{ - 1}   \notag \\
&=(\mathbf{A}_{{}}^{T}\mathbf{A})^{ - 1}
 - \lambda (\mathbf{A}_{{}}^{T}\mathbf{A})^{ - 1}{ \left( { \mathbf{I}} +  \lambda(\mathbf{A}_{{}}^{T}\mathbf{A})^{ - 1}   \right)^{ - 1}}(\mathbf{A}_{{}}^{T}\mathbf{A})^{- 1}  \notag \\
&=(\mathbf{A}_{{}}^{T}\mathbf{A})^{ - 1} - \lambda (\mathbf{A}_{{}}^{T}\mathbf{A})^{ - 1}
 \pmb{\Phi}
 (\mathbf{A}_{{}}^{T}\mathbf{A})^{ - 1},  \label{AtALamdaAtAPhiASDko32032}
\end{align}
where for simplicity,
  we define
  \begin{equation}\label{Phi2IAA3923232ksd23fd43kf}
\pmb{\Phi} = { \left( { \mathbf{I}} +  \lambda(\mathbf{A}_{{}}^{T}\mathbf{A})^{ - 1}   \right)^{ - 1}}
 \end{equation}
 satisfying
    \begin{equation}\label{PhiZeroLimit30kdsdsadsa32}
\underset{\lambda \to 0}{\mathop{\lim }}\,{ \pmb{\Phi}} =\underset{\lambda \to 0}{\mathop{\lim }}\,{ { \left( { \mathbf{I}} +  \lambda(\mathbf{A}_{{}}^{T}\mathbf{A})^{ - 1}   \right)^{ - 1}}}
 = {\mathbf{I}}.
 \end{equation}
Then we substitute (\ref{AtALamdaAtAPhiASDko32032})  into (\ref{C2dseqf32vdasa32})
 to simplify (\ref{C2dseqf32vdasa32}) into
  \begin{align}
\mathbf{C}&={\mathbf{ A}_k}{\left( \begin{array}{l}
(\mathbf{A}_{{k}}^{T}{\mathbf{ A}_k})^{ - 1} - (\mathbf{A}_{{k}}^{T}{\mathbf{ A}_k})^{ - 1} +  \\
\lambda (\mathbf{A}_{{k}}^{T}{\mathbf{ A}_k})^{ - 1}\pmb{\Phi}(\mathbf{A}_{{k}}^{T}{\mathbf{ A}_k})^{ - 1}  \end{array} \right)^{-1}}   \mathbf{A}_{{k}}^{T}{{\bf{H}}}  \notag \\
&=\lambda {\mathbf{ A}_k} (\mathbf{A}_{{k}}^{T}{\mathbf{ A}_k})^{ - 1}\pmb{\Phi}(\mathbf{A}_{{k}}^{T}{\mathbf{ A}_k})^{ - 1}\mathbf{A}_{{k}}^{T}{{\bf{H}}} \notag \\
&=\lambda ({{\bf{ A}}_k^+})^{T} \pmb{\Phi}
 {{\bf{ A}}_k^+}
 {{\bf{H}}}  \notag \\
&=\lambda ({{\bf{ A}}_k^+})^{T} \pmb{\Phi}
 \mathbf{\bar D},   \label{C2AA230sda1sd}
\end{align}
where  (\ref{Ainv2AtAinvAt9096})  and (\ref{D_matrix_def_111Orig}) are utilized successively.

Finally,  let us utilize (\ref{C2AA230sda1sd}) to verify that (\ref{B2HCnumdaICt2021JunSmallLamda}) is equal to  (\ref{B_Matrix_def1bOrig}) when   $\lambda \to 0$.
We substitute (\ref{C2AA230sda1sd}) into (\ref{B2HCnumdaICt2021JunSmallLamda}) to obtain
\begin{align}
&{{\mathbf{B}}^{T}}  \notag \\
& ={{\left(
{{  \lambda {\mathbf{\bar D}}^T  \pmb{\Phi} {{\bf{ A}}_k^+}
    \lambda ({{\bf{ A}}_k^+})^{T}
     }} \pmb{\Phi} \mathbf{\bar D} +\lambda \mathbf{D}^{T}\mathbf{D}+\lambda \mathbf{I}
 \right)^{-1}}} \lambda {\mathbf{\bar D}}^T  \pmb{\Phi} {{\bf{ A}}_k^+}  \notag \\
&=\frac{1}{\lambda}
 {\left(
\mathbf{D}^{T}\mathbf{D}+ \mathbf{I} +  \lambda {\mathbf{\bar D}}^T
  \pmb{\Phi} {{\bf{ A}}_k^+}
     ({{\bf{ A}}_k^+})^{T}  \pmb{\Phi}
 \mathbf{\bar D} \right)^{-1}}\lambda {\mathbf{\bar D}}^T  \pmb{\Phi} {{\bf{ A}}_k^+}   \notag \\
& = {\left(
\mathbf{D}^{T}\mathbf{D}+ \mathbf{I} +  \lambda {\mathbf{\bar D}}^T
  \pmb{\Phi} {{\bf{ A}}_k^+}
     ({{\bf{ A}}_k^+})^{T}  \pmb{\Phi}
 \mathbf{\bar D} \right)^{-1}} {\mathbf{\bar D}}^T  \pmb{\Phi} {{\bf{ A}}_k^+},  \label{Bt2HA329kdsdsaasdv32a}
\end{align}
 and then let $\lambda \to 0$ in (\ref{Bt2HA329kdsdsaasdv32a}) to obtain
\begin{align}
&\underset{\lambda \to 0}{\mathop{\lim }}\,{ {{\mathbf{B}}^{T}}   }   \notag \\
&=\underset{\lambda \to 0}{\mathop{\lim }}\, {\left(
\mathbf{D}^{T}\mathbf{D}+ \mathbf{I} +  \lambda {\mathbf{\bar D}}^T
  \pmb{\Phi} {{\bf{ A}}_k^+}
     ({{\bf{ A}}_k^+})^{T}  \pmb{\Phi}
 \mathbf{\bar D} \right)^{-1}}   {\mathbf{\bar D}}^T  \pmb{\Phi} {{\bf{ A}}_k^+}  \notag \\
&={{ \left( {\mathbf{\bar D}^{T}\mathbf{\bar D}+ \mathbf{I} }\right)^{-1} }} {\mathbf{\bar D}}^T  {\mathbf{I}}  {{\bf{ A}}_k^+} ={{ \left( {\mathbf{\bar D}^{T}\mathbf{\bar D}+ \mathbf{I} }\right)^{-1} }} {{\bf{\bar D}}^T} {{\bf{ A}}_k^+} = {{\bf{\bar B}}^T},  \notag
\end{align}
 where  (\ref{PhiZeroLimit30kdsdsadsa32})
  and  (\ref{B_Matrix_def1bOrig}) are applied in turn.


\section{The Efficient  Inverse
	Cholesky Factorization  and the Corresponding Parallelization}

In this section, we will summarize the
efficient  inverse
Cholesky factorization
firstly,
and then develop
the memory-saving parallel implementation of the inverse Cholesky factorization.

\subsection{The Existing
	Efficient Inverse Cholesky Factorization}

In \cite{my_inv_chol_paper},
the inverse Cholesky factor
of a Hermitian matrix ${{\bf{R}}_i} \in {\Re ^{i \times i}}$
is the upper-triangular ${{\bf{F}}_i}$
satisfying
\begin{equation}\label{L_m_def12431RidgeInvJun30JunY24}
	{{\bf{F}}_i}{{\bf{F}}_i^{T}}={{\bf{R}}_i^{-1}}= ({{\bf{A}}_i^T}{{\bf{A}}_i}+ \lambda {\bf{I}})^{-1},
\end{equation}
where ${{\bf{A}}_i}$ is with $i$ columns,
the ridge parameter $\lambda$ is a positive real number,
and ${{\bf{ R}}_i}$ is defined by
\begin{equation}\label{R_define12321numdaRidge302923}
	{{\bf{ R}}_i}={{\bf{A}}_i^T}{{\bf{A}}_i}+ \lambda {\bf{I}}.
\end{equation}
From (\ref{L_m_def12431RidgeInvJun30JunY24}),  we can deduce
\begin{equation}\label{L_m_def12431Apdx2390sd}
{\bf{F}}_i^{-T}{{\bf{F}}_i^{-1}}={{{{\bf{R}}_i}}},
\end{equation}
which shows that the lower-triangular ${\bf{F}}_i^{-T}$ is the
conventional
Cholesky factor~\cite[Theorem 4.2.5]{Matrix_Computations_book} of ${{{{\bf{R}}_i}}}$.

${{\bf{A}}_i}$ can be partitioned into
\begin{equation}\label{Ai2Aiai320ds3ds4fs}
{{\bf{A}}_{i}} = \left[ {\begin{array}{*{20}{c}}
		{{\bf{A}}_{i - 1}} & {{\bf{a}}_i}
\end{array}} \right],
\end{equation}
where ${{\bf{A}}_{i - 1}}$ and ${{\bf{a}}_i}$ denotes the first $i-1$    columns and the $i^{th}$ column of  $\mathbf{A}_{i}$, respectively.
Then  (\ref{Ai2Aiai320ds3ds4fs}) can be substituted into (\ref{R_define12321numdaRidge302923}) to
partition ${{\bf{R}}_{i}}$   into 
\begin{equation}\label{R_def_perhaps_No_R1zhfPaperJunY24crc}
{{\bf{R}}_{i}}=\left[ \begin{matrix}
	{{\bf{A}}_{i-1}^T}{{\bf{A}}_{i-1}}+ \lambda {\bf{I}} & {{\bf{A}}_{i-1}^T}{{\bf{a}}_i}  \\
	{{\bf{a}}_i^T}{\bf{A}}_{i-1} & {{\bf{a}}_i^T}{{\bf{a}}_i}+ \lambda  \\
\end{matrix} \right]=\left[ \begin{matrix}
	{{\bf{R}}_{i-1}} & {\mathbf{p}}  \\
	\mathbf{p}^{T} & u  \\
\end{matrix} \right],
\end{equation}
where ${\mathbf{p}}$ and $u$  are a column vector and a scalar, respectively, and ${\bf{R}}_{i-1}$,
the $(i-1) \times (i-1)$ leading principal submatrix
of ${\bf{R}}_{i}$,  is also
defined
from ${{\bf{A}}_{i-1}}$
by
(\ref{R_define12321numdaRidge302923}).
The efficient inverse Cholesky factorization~\cite{my_inv_chol_paper}
updates
the inverse Cholesky factor of  ${{\bf{R}}_{i-1}}$ into that of ${{\bf{R}}_{i}}$
by
\begin{equation}\label{Fi2ffiiT3902}
{{{\bf{F}}_{i}}}=\left[ \begin{matrix}
	{{\bf{F}}_{i-1}} & {\bf{\bar f}}_{i}  \\
	{{\bf{0}}_{i - 1}^T} & f_{ii}  \\
\end{matrix} \right]=\left[ {\begin{array}{*{20}{c}}
		{\left( {\begin{array}{*{20}{c}}
					{{{\bf{F}}_{i - 1}}}\\
					{{\bf{0}}_{i - 1}^T}
			\end{array}} \right)}&{{\bf{f}}_i^{}}
\end{array}} \right],
\end{equation}
where ${\bf{\bar f}}_{i}$ and $f_{ii}$ are computed by
\begin{subnumcases}{\label{ZF_def_L_2_items3abzhfPaperJunY24ovsbn}}
{\bf{\bar f}}_{i} = 1/\sqrt {u - {{\bf{p}}^T}{{\bf{F}}_{i}}{\bf{F}}_{i}^T{\bf{p}}}  &  \label{ZF_def_L_2_items3azhfPaper}\\
f_{ii} =  - g{{\bf{F}}_{i}}{\bf{F}}_{i}^T{\bf{p}}. & \label{ZF_def_L_2_items3bzhfPaper}
\end{subnumcases}
In (\ref{Fi2ffiiT3902})
and
(\ref{ZF_def_L_2_items3abzhfPaperJunY24ovsbn}),
${{\bf{f}}_{i}}$ and $f_{ii}$ denote the $i^{th}$ column and diagonal entry of ${{\bf{F}}_{i}}$,
respectively, while ${\bf{\bar f}}_{i}$ denotes
${{\bf{f}}_{i}}$ with the last entry $f_{ii}$ removed.


In this paper,
(\ref{Ai2Aiai320ds3ds4fs}) is extended into
a column-partitioned matrix
\begin{equation}\label{A2iki3209923}
\mathbf{A}_{k}^{{}}=\left[ \begin{matrix}
	\mathbf{A}_{i}^{{}} & \mathbf{\underline{A}}_{{k-i}}^{{}}  \\
\end{matrix} \right],
\end{equation}
where $\mathbf{A}_{i}$ and ${{\mathbf{\underline{A}}}_{k-i}}$ are the first $i$ and  last $k-i$ columns of $\mathbf{A}_{k}^{{}}$ with $k$ columns, respectively.
Then  (\ref{A2iki3209923}) can be substituted into (\ref{R_define12321numdaRidge302923}) to
partition
${{\bf{ R}}_{k}}$
into
the $2 \times 2$ block
matrix, i.e.,
\begin{equation}\label{R_def_perhaps_No_R1apdx9ds23}
	{{\bf{R}}_{k}}=\left[ \begin{matrix}
		{{\bf{R}}_i} & {{\mathbf{ A}}_i^T}  {{\mathbf{\underline{A}}}_{k-i}}  \\
		{{\mathbf{\underline{A}}}_{k-i}^T}  {{\mathbf{ A}}_i} & {{\mathbf{\underline{A}}}_{k-i}^T}{{\mathbf{\underline{A}}}_{k-i}} +\lambda \mathbf{I}  \\
	\end{matrix} \right]=\left[ \begin{matrix}
		{{\bf{R}}_i} & {\mathbf{P}}  \\
		\mathbf{P}^{T} & {\mathbf{U}}  \\
	\end{matrix} \right],
\end{equation}
where
\begin{subnumcases}{\label{UP2AA23ds23Both}}
	\mathbf{U} = {{\mathbf{\underline{A}}}_{k-i}^T}{{\mathbf{\underline{A}}}_{k-i}} +\lambda \mathbf{I}   \in {\Re ^{(k-i) \times (k-i)}} &  \label{UP2AA23ds23BothUU}\\
	\mathbf{P}={{\mathbf{ A}}_i^T}{{\mathbf{\underline{A}}}_{k-i}} \in {\Re ^{i \times (k-i)}}.  &  \label{UP2AA23ds23BothPP}
\end{subnumcases}
The inverse Cholesky factor of  ${{\bf{R}}_i}$ is updated
into that of ${{\bf{R}}_{k}}$
by
\begin{equation}\label{L_big_BLK_def1Apdx329sd}
	{{\bf{F}}_{k}}=\left[ \begin{matrix}
		{{\bf{F}}_i} & \mathbf{T}  \\
		\mathbf{0} & \mathbf{G}  \\
	\end{matrix} \right]
\end{equation}
	where
	$\mathbf{T}\in {\Re ^{i \times (k-i)}}$ and
	$\mathbf{G} \in {\Re ^{(k-i) \times (k-i)}}$
	are computed by
	\begin{footnotesize}
		\begin{subnumcases}{\label{ZF_def_L_2_items3abJun2021appendix}}
			{{\mathbf{G}}}{{\mathbf{G}}^{T}}=\left({ {{\mathbf{\underline{A}}}_{k-i}^T}{{\mathbf{\underline{A}}}_{k-i}} +\lambda \mathbf{I}}- {{\mathbf{\underline{A}}}_{k-i}^T}{{\mathbf{ A}}_i}{{\bf{F}}_i}{{\bf{F}}_i^{T}}{{{\mathbf{ A}}_i^T}  {{\mathbf{\underline{A}}}_{k-i}}  }\right)^{-1} &  \label{ZF_def_L_2_items3aApdx239sd}\\
			\mathbf{T}=-{{\bf{F}}_i}{{\bf{F}}_i^{T}}{{{\mathbf{ A}}_i^T}{  {{\mathbf{\underline{A}}}_{k-i}}  }}\mathbf{G}. & \label{ZF_def_L_2_items3bJun2021apdxkg50d}
		\end{subnumcases}
	\end{footnotesize}
	
	In   this paper,
	the inverse Cholesky factor of  ${{\bf{ R}}_{i}}$  is updated into that of ${{\bf{ R}}_{k}}$ by
	the above-described block inverse Cholesky factorization,
	where
	${{\bf{ F}}_i}$ is updated into ${{\bf{ F}}_{k}}$
	by
	(\ref{ZF_def_L_2_items3abJun2021appendix})
	and (\ref{L_big_BLK_def1Apdx329sd}),
	and  the initial ${{\bf{ F}}_{i}}$ is computed
	by (\ref{L_m_def12431RidgeInvJun30JunY24}).
	$\mathbf{G}$ in
	(\ref{ZF_def_L_2_items3aApdx239sd})
	is also
	the upper-triangular
	inverse Cholesky factor,
	which
	can be
	computed
	by the efficient inverse Cholesky factorization in \cite{my_inv_chol_paper}
	or
	by inverting and transposing the lower-triangular Cholesky
	factor.

	\subsection{The Memory-Saving Parallel Implementation of the Inverse Cholesky Factorization}

	In (\ref{ZF_def_L_2_items3abJun2021appendix}) for  the block inverse Cholesky factorization,
	$-{{\mathbf{F}}_{i}}\mathbf{F}_{i}^{T}\mathbf{A}_{i}^{T}\mathbf{\underline{A}}_{k-i}$ and
	$\mathbf{\underline{A}}_{k-i}^{T}\mathbf{\underline{A}}_{k-i}^{{}} +\lambda \mathbf{I} - \mathbf{\underline{A}}_{k-i}^{T}\mathbf{A}_{i}^{{}}{{\mathbf{F}}_{i}}\mathbf{F}_{i}^{T}\mathbf{A}_{i}^{T}\mathbf{\underline{A}}_{k-i}^{{}}$
	can be written	as  
	\begin{small}
		\begin{subnumcases}{\label{PiXiboth9023sdk23}}
			{{\mathbf{\Pi }}_{i}}=-{{\mathbf{F}}_{i}}\mathbf{F}_{i}^{T}\mathbf{A}_{i}^{T}\mathbf{\underline{A}}_{k-i}^{{}} &  \label{pi2FFAA2390923}\\
			{{\mathbf{\Xi }}_{k-i}}= \mathbf{\underline{A}}_{k-i}^{T}\mathbf{\underline{A}}_{k-i}^{{}} +\lambda \mathbf{I} - \mathbf{\underline{A}}_{k-i}^{T}\mathbf{A}_{i}^{{}}{{\mathbf{F}}_{i}}\mathbf{F}_{i}^{T}\mathbf{A}_{i}^{T}\mathbf{\underline{A}}_{k-i}^{{}},  &  \label{Xi2AAFFAA23094d3}
		\end{subnumcases}
	\end{small}%
	where $i=1,2,...,k-1$, ${{\mathbf{\Pi }}_{i}} \in {\Re ^{i \times (k-i)}}$, and  ${{\mathbf{\Xi }}_{k-i}} \in {\Re ^{(k-i) \times (k-i)}}$.  																																																																			On the other hand, it can be seen from (\ref{ZF_def_L_2_items3abzhfPaperJunY24ovsbn}) and																																																																			(\ref{Fi2ffiiT3902})																																																																		for the efficient inverse Cholesky factorization~\cite{my_inv_chol_paper} 																																																																			that																																																																		in the $i^{th}$ iteration,																																																																			the $i^{th}$ column of ${{\bf{F}}_{i}}$ is computed by (\ref{ZF_def_L_2_items3abzhfPaperJunY24ovsbn}),																																																																			and then merged with ${{\bf{F}}_{i-1}}$																																																																			to obtain																																																																			${{\bf{F}}_{i}}$																																																																			by (\ref{Fi2ffiiT3902}).																																																																																																																																				Then in the $i^{th}$ ($i=1,2,...,k-1$) iteration																																																																			of the efficient inverse Cholesky factorization~\cite{my_inv_chol_paper},																																																																			we can use ${{\mathbf{F}}_{i}}$ to																																																																																																																																				compute 																																																															${{\mathbf{\Pi }}_{i}}$ and  ${{\mathbf{\Xi }}_{k-i}}$																																																																			by  (\ref{pi2FFAA2390923})																																																																			and		(\ref{Xi2AAFFAA23094d3}),																																																											respectively,  in order to improve the parallelization.
	
	To reduce the computational complexity,	we	utilize	${\mathbf{\Xi }}_{k-i+1}$ and ${\bf{\Pi }}_{i - 1}$ to compute																																																																		${{\mathbf{\Xi }}_{k-i}}$ and ${{\mathbf{\Pi }}_{i}}$																																																																			by					\begin{footnotesize}																																																																				\begin{equation}\label{XiXi2390A032dsModify}																																																																					{{\mathbf{\Xi }}_{k-i}}={{\mathbf{\Xi }}_{k-i+1}^{[-1,-1]}} - f_{ii}^2 {{\bf{\Xi }}_{k - i + 1}}(1,2:end)^T  {{\bf{\Xi }}_{k - i + 1}}(1,2:end)																																																																				\end{equation}							
	\end{footnotesize}%
	and								\begin{footnotesize}																																																																					\begin{multline}\label{FinalPi19sd3SimplifiedMore}																																																					{{\bf{\Pi }}_i} = \\																																																																						\left[ {\begin{array}{*{20}{c}}																																																																								{  {\bf{\Pi }}_{i - 1}(:,2:end) - {{\bf{\Pi }}_{i - 1}}(:,1) \cdot {f_{ii}^2}{{{\bf{\Xi }}}_{k - i + 1}}(1,2:end)}\\																																																																								- {{f_{ii}^2}{{{\bf{\Xi }}}_{k - i + 1}}(1,2:end)}																																																																						\end{array}} \right],																																																																				\end{multline}							
	\end{footnotesize}%
	respectively,	instead of (\ref{pi2FFAA2390923})																																																																				and					(\ref{Xi2AAFFAA23094d3}),																																																																				and compute the $i^{th}$ column of ${{\bf{F}}_{i}}$																																																																				(i.e., ${\bf{\bar f}}_{i}$ and $f_{ii}$ in  (\ref{Fi2ffiiT3902}))																																																																				by				\begin{subnumcases}{\label{fiifli2Pi3902sdBoth}}																																																																					f_{ii} = 1/\sqrt {{{{\bf{\Xi }}}_{k - i + 1}}(1,1)} &  \label{fii2sqrt2390sd23}\\																																																																					{\bf{\bar f}}_{i} =   f_{ii} {{\bf{\Pi }}_{i - 1}}(:,1).  &  \label{f1i1i2gPi39d234d3}																																																																				\end{subnumcases}
	In the next subsection,																																																																			we will derive the above (\ref{XiXi2390A032dsModify}),																																																																				(\ref{FinalPi19sd3SimplifiedMore}) and																																																																				(\ref{fiifli2Pi3902sdBoth}),																																																																				where  ${{\mathbf{\Xi }}_{k-i+1}^{[-1,-1]}}$ is ${{\mathbf{\Xi }}_{k-i+1}}$ with the first row and column removed,																																																																				and the notations $ {{{\bf{\Xi }}}_{k - i + 1}}(1,2:end)$,																																																																																																																																						${\bf{\Pi }}_{i - 1}(:,2:end)$,   ${{\bf{\Pi }}_{i - 1}}(:,1)$ and ${{{\bf{\Xi }}}_{k - i + 1}}(1,1)$																																																																				follow the MATLAB standard. Notice that in the remainder of this paper,																																																																				the notations will also follow the MATLAB standard.
	
	To save memories, we will give an implementation that cover the upper-triangular part of
	the Hermitian   matrix ${{\bf{R}}_k}$ with ${{\bf{F}}_k}$, the inverse Cholesky factor of ${{\bf{R}}_k}$.
	It can be seen from
	(\ref{R_def_perhaps_No_R1apdx9ds23})
	that
	\begin{subnumcases}{\label{}}
		{\bf{R}}(1:i,i + 1:k)={\mathbf{P}}=\mathbf{A}_{i}^{T}\mathbf{\underline{A}}_{k-i} &  \label{}\\
		{\bf{R}}(i + 1:k,i + 1:k)={\mathbf{U}}=\mathbf{\underline{A}}_{k-i}^{T}\mathbf{\underline{A}}_{k-i}^{{}} +\lambda \mathbf{I},  &  \label{}
	\end{subnumcases}
	which are substituted into (\ref{pi2FFAA2390923}) and
	(\ref{Xi2AAFFAA23094d3}) 																																																																																																															to obtain
	\begin{subnumcases}{\label{Pi2FFRXiR932ds23}}
		{{\bf{\Pi }}_i} = -{{\bf{F}}_i}{\bf{F}}_i^T{\bf{R}}(1:i,i + 1:k) &  \label{Pi2FFRXiR932ds23aaa}\\
		{\begin{split}
				{{\bf{\Xi }}_{k - i}} = {\bf{R}}(i + 1:k,i + 1:k) - \quad \quad \quad \quad \quad \quad \quad \\
				\quad \quad {\bf{R}}(1:i,i + 1:k)_{}^T {{\bf{F}}_i}{\bf{F}}_i^T{\bf{R}}(1:i,i + 1:k).
		\end{split}}   &  \label{Pi2FFRXiR932ds23bbb}
	\end{subnumcases}
	Let $k=1$ in (\ref{L_m_def12431RidgeInvJun30JunY24}) to obtain
	\begin{equation}\label{InitialF1aabb329sd}
		{{\mathbf{F}}_{1}}\mathbf{F}_{1}^{T}=1/\mathbf{R}(1,1),
	\end{equation}
	which is substituted
	into
	(\ref{Pi2FFRXiR932ds23}) with $i=1$, i.e.,
	\begin{footnotesize}
		\begin{subnumcases}{\label{InitialPi2FFRXiR932ds23NoUse}}
			{{\bf{\Pi }}_1} =-{{\bf{F}}_1}{\bf{F}}_1^T{\bf{R}}(1,2:k) &   \label{InitialPi2FFRXiR932ds23aaaNoUse}\\
			{{\bf{\Xi }}_{k - 1}} = {\bf{R}}(2:k,2:k) - {\bf{R}}(1,2:k)_{}^T {{\bf{F}}_1}{\bf{F}}_1^T{\bf{R}}(1,2:k),   &  \label{InitialPi2FFRXiR932ds23bbbNoUse}
		\end{subnumcases}
	\end{footnotesize}%
	to compute the initial ${{\bf{\Pi }}_1}$ and ${{\bf{\Xi }}_{k - 1}}$
	by
	\begin{small}
		\begin{subnumcases}{\label{InitialPi2FFRXiR932ds23}}
			{{\mathbf{\Pi }}_{1}}=-\frac{\mathbf{R}(1,2:k)}{\mathbf{R}(1,1)} &    \label{InitialPi2FFRXiR932ds23aaa}\\
			{{\mathbf{\Xi }}_{k-1}}=\mathbf{R}(2:k,2:k)-\frac{\mathbf{R}(1,2:k)_{{}}^{T}\mathbf{R}(1,2:k)}{\mathbf{R}(1,1)}.   &  \label{InitialPi2FFRXiR932ds23bbb}
		\end{subnumcases}
	\end{small}%
	Moreover, we compute the  initial ${{\bf{ F}}_{1}}$ by writing (\ref{InitialF1aabb329sd}) as
	\begin{equation}\label{InitialF1aabb329sdUse93s3d}
		{{\mathbf{F}}_{1}}= \sqrt {1/{\bf{R}}(1,1)}.
	\end{equation}
	
	When $i \ge 2$,
	none of the entries of $\mathbf{R}$ are required in  (\ref{XiXi2390A032dsModify}),
	(\ref{FinalPi19sd3SimplifiedMore}),
	(\ref{fii2sqrt2390sd23})
	and
	(\ref{f1i1i2gPi39d234d3}),
	which are utilized to compute
	${{\mathbf{\Xi }}_{k-i}}$,  ${{\mathbf{\Pi }}_{i}}$,
	$f_{ii}$ and
	${\bf{\bar f}}_{i}$, respectively.
	Then
	to save memories,
	we
	store ${{\bf{\Xi }}_{k - 1}}$,
	${{\mathbf{\Pi }}_{1}}$    and ${\mathbf{F}}_{1}$ in ${\bf{R}}(2:k,2:k)$,
	$\mathbf{R}(1,2:k)$  and ${\bf{R}}(1,1)$, respectively,
	to
	write
	(\ref{InitialPi2FFRXiR932ds23bbb}),
	(\ref{InitialPi2FFRXiR932ds23aaa})
	and
	(\ref{InitialF1aabb329sdUse93s3d})  as
	\begin{small}
		\begin{subnumcases}{\label{i1Initial2SaveMem}}
			{\begin{split}
					{\bf{R}}(2:k,2:k) ={\bf{R}}(2:k,2:k)  \quad \quad \quad \quad \quad \quad \quad \quad \quad  \\
					\quad \quad \quad \quad \quad \quad \quad \quad \quad   - \frac{{{\bf{R}}{{(1,2:k)}^T}{\bf{R}}(1,2:k)}}{{{\bf{R}}(1,1)}}
			\end{split}}   &  \label{i1Initial2SaveMem1}  \\
			\mathbf{R}(1,2:k)= - \frac{\mathbf{R}(1,2:k)}{\mathbf{R}(1,1)} &    \label{i1Initial2SaveMem2}\\
			{\bf{R}}(1,1) = \sqrt {1/{\bf{R}}(1,1)}.   &  \label{i1Initial2SaveMem3}
		\end{subnumcases}
	\end{small}%
	In the $i^{th}$ ($i \ge 2$) iteration,
	we can also save memories by storing
	${{\mathbf{\Xi }}_{k-i}}$,  ${{\mathbf{\Pi }}_{i}}$,
	$f_{ii}$ and
	${\bf{\bar f}}_{i}$
	in ${\bf{R}}(i + 1:k,i + 1:k)$,
	${\bf{R}}(1:i,i + 1:k)$,
	$\mathbf{R}(i,i)$ and ${\bf{R}}(1:i - 1,i)$,
	respectively, to write
	(\ref{XiXi2390A032dsModify}),
	(\ref{FinalPi19sd3SimplifiedMore}),
	(\ref{fii2sqrt2390sd23})
	and
	(\ref{f1i1i2gPi39d234d3})
	as
	\begin{small}
		\begin{subnumcases}{\label{InvCholSaveMem3290}}
			{\begin{split}
					{\bf{R}}(i + 1:k,i + 1:k) ={\bf{R}}(i + 1:k,i + 1:k)  \quad \quad \quad  \\
					\quad \quad \quad \quad \quad \quad \quad  - \frac{{{\bf{R}}{{(i,i + 1:k)}^T}{\bf{R}}(i,i + 1:k)}}{{{\bf{R}}(i,i)}}
			\end{split}}   &  \label{InvCholSaveMem3290aaa}  \\
			{\begin{split}
					{\bf{R}}(1:i,i + 1:k) = \quad \quad \quad \quad \quad \quad \quad \quad \quad \quad \quad \quad \quad \quad \\
					\quad \left[ {\begin{array}{*{20}{c}}
							{{\bf{R}}(1:i - 1,i + 1:k) - \frac{{{\bf{R}}(1:i - 1,i){\bf{R}}(i,i + 1:k)}}{{{\bf{R}}(i,i)}}}\\
							- {\frac{{{\bf{R}}(i,i + 1:k)}}{{{\bf{R}}(i,i)}}}
					\end{array}} \right]
			\end{split}}   &  \label{InvCholSaveMem3290bbb}  \\
			{\bf{R}}(i,i) = \sqrt {1/{\bf{R}}(i,i)}  &  \label{InvCholSaveMem3290ccc}\\
			{\bf{R}}(1:i - 1,i) =   {\bf{R}}(i,i){\bf{R}}(1:i - 1,i).  &  \label{InvCholSaveMem3290ddd}
		\end{subnumcases}
	\end{small}%
	Notice that to deduce (\ref{InvCholSaveMem3290}), we utilize the fact that
	${{\mathbf{\Xi }}_{k-(i-1)}}={{\mathbf{\Xi }}_{k-i+1}}$ and  ${{\mathbf{\Pi }}_{i-1}}$
	have been stored
	in ${\bf{R}}\left((i-1) + 1:k,(i-1) + 1:k\right)={\bf{R}}(i:k,i:k)$ and
	${\bf{R}}(1:i-1,(i-1) + 1:k)={\bf{R}}(1:i-1,i:k)$,
	respectively,
	where $i \ge 2$.
	
	It can be seen that (\ref{i1Initial2SaveMem}) is equivalent to (\ref{InvCholSaveMem3290}) with $i=1$.
	Then we can compute (\ref{InvCholSaveMem3290}) iteratively for $i=1,2,\cdots,k$, to
	cover the upper-triangular part of  ${\bf{R}}_k$  with  the inverse Cholesky factor ${\bf{F}}_k$ satisfying ${\bf{F}}_k{\bf{F}}_k^{T}={\bf{R}}_k^{-1}$.
	The corresponding implementation is summarized in \textbf{Algorithm 11}.
	
	\begin{algorithm}
		\caption{:~\bf The parallel implementation of the inverse Cholesky factorization}
		\begin{algorithmic}[1]
			\Require The upper-triangular part of a  Hermitian matrix ${\bf{R}}$
			\Ensure The inverse Cholesky factor of ${\bf{R}}$ in the upper-triangular part of  ${\bf{R}}$
			\For{$i=1:k$ ($k$ is the size of ${\bf{R}}$)}
			\State  $\begin{array}{l}
				{\bf{R}}(i + 1:k,i + 1:k) = \\
				{\bf{R}}(i + 1:k,i + 1:k) - \frac{{{\bf{R}}{{(i,i + 1:k)}^T}{\bf{R}}(i,i + 1:k)}}{{{\bf{R}}(i,i)}};
			\end{array}$
			\State $\begin{array}{l}
				{\bf{R}}(1:i,i + 1:k) = \\
				\left[ {\begin{array}{*{20}{c}}
						{{\bf{R}}(1:i - 1,i + 1:k) - \frac{{{\bf{R}}(1:i - 1,i){\bf{R}}(i,i + 1:k)}}{{{\bf{R}}(i,i)}}}\\
						-{\frac{{{\bf{R}}(i,i + 1:k)}}{{{\bf{R}}(i,i)}}}
				\end{array}} \right];
			\end{array}$
			\State ${\bf{R}}(i,i) = \sqrt {1/{\bf{R}}(i,i)} $;
			\State ${\bf{R}}(1:i - 1,i) =   {\bf{R}}(i,i){\bf{R}}(1:i - 1,i)$;
			\EndFor
			\State Now the upper-triangular part of  ${\bf{R}}$  becomes  the inverse Cholesky factor ${{\bf{F}}}$ satisfying ${{\bf{F}}}{{\bf{F}}^{T}}={\bf{R}}^{-1}$;
		\end{algorithmic}
	\end{algorithm}

	\subsection{The Derivation of (\ref{XiXi2390A032dsModify}),
		(\ref{FinalPi19sd3SimplifiedMore}) and (\ref{fiifli2Pi3902sdBoth})}
	In this subsection, let us deduce (\ref{XiXi2390A032dsModify}),
	(\ref{FinalPi19sd3SimplifiedMore}) and (\ref{fiifli2Pi3902sdBoth}) utilized in the last subsection.
	
	Obviously,
	${{\mathbf{\underline{A}}}_{k-i}}$  in
	(\ref{A2iki3209923})
	satisfies
	\begin{equation}\label{Ak2aA389323}
		{\bf{\underline{A}}}_{k - (i - 1)}^{}={\bf{\underline{A}}}_{k - i + 1}^{}=\left[ {\begin{array}{*{20}{c}}
				{{\bf{a}}_i^{}}&{{\bf{\underline{A}}}_{k - i}^{}}
		\end{array}} \right],
	\end{equation}
	where  $\mathbf{a}_{i}^{{}}$  is  the $i^{th}$ column of $\mathbf{A}_{k}^{{}}$.
	Then  use
	(\ref{Ak2aA389323})
	to write
	\begin{gather}
		{\bf{F}}_{i - 1}^T{\bf{A}}_{i - 1}^T{\bf{\underline{A}}}_{k - (i - 1)}^{} = {\bf{F}}_{i - 1}^T{\bf{A}}_{i - 1}^T\left[ {\begin{array}{*{20}{c}}
				{{\bf{a}}_i^{}}&{{\bf{\underline{A}}}_{k - i}^{}}
		\end{array}} \right]  \notag \\
		= \left[ {\begin{array}{*{20}{c}}
				{{\bf{F}}_{i - 1}^T{\bf{A}}_{i - 1}^T{\bf{a}}_i^{}}&{{\bf{F}}_{i - 1}^T{\bf{A}}_{i - 1}^T{\bf{\underline{A}}}_{k - i}^{}}
		\end{array}} \right].  \label{FAA2FAaAFA23233}
	\end{gather}
	From (\ref{R_def_perhaps_No_R1zhfPaperJunY24crc}), we can write
	$\left[ {\begin{array}{*{20}{c}}
			{\mathbf{p}}\\
			u
	\end{array}} \right] = \left[ {\begin{array}{*{20}{c}}
			{\bf{A}}_{i - 1}^T{\bf{a}}_i \\
			{\bf{a}}_i^T {\bf{a}}_i+\lambda
	\end{array}} \right]$,																																																																																								
	which is substituted   into
	(\ref{ZF_def_L_2_items3abzhfPaperJunY24ovsbn}) to obtain
	\begin{subnumcases}{\label{colNewWhole32099d3}}
		f_{ii} = 1/\sqrt {{\bf{a}}_i^T {\bf{a}}_i+\lambda - {\bf{a}}_i^T {\bf{A}}_{i - 1}{{\bf{F}}_{i-1}}{\bf{F}}_{i-1}^T{\bf{A}}_{i - 1}^T{\bf{a}}_i}  &  \label{colNewWhole32099d3aaa}\\
		{\bf{\bar f}}_{i} =  - f_{ii} {{\bf{F}}_{i-1}}{\bf{F}}_{i-1}^T {\bf{A}}_{i - 1}^T{\bf{a}}_i. & \label{colNewWhole32099d3bbb}
	\end{subnumcases}

	To update ${\bf{\Pi }}_{i - 1}$ into ${{\bf{\Pi }}_i}$ efficiently,
	substitute (\ref{FAA2FAaAFA23233})  into  (\ref{pi2FFAA2390923}) with $i=i-1$  to obtain
	\begin{multline}\label{Pi2FiFiaA39ds23d32}
		{{\mathbf{\Pi }}_{i-1}}=-{{\mathbf{F}}_{i-1}}\left[ {\begin{array}{*{20}{c}}
				{{\bf{F}}_{i - 1}^T{\bf{A}}_{i - 1}^T{\bf{a}}_i^{}}&{{\bf{F}}_{i - 1}^T{\bf{A}}_{i - 1}^T{\bf{\underline{A}}}_{k - i}^{}}
		\end{array}} \right] \\
		=-\left[ \begin{matrix}
			{{\mathbf{F}}_{i-1}}\mathbf{F}_{i-1}^{T}\mathbf{A}_{i-1}^{T}\mathbf{a}_{i}^{{}} & {{\mathbf{F}}_{i-1}}\mathbf{F}_{i-1}^{T}\mathbf{A}_{i-1}^{T}\mathbf{\underline{A}}_{k-i}^{{}}  \\
		\end{matrix} \right],
	\end{multline}
	which can be written as
	\begin{subnumcases}{\label{NoUsePi2parts329}}
		{{\bf{\Pi }}_{i - 1}}(:,1)=-{{\mathbf{F}}_{i-1}}\mathbf{F}_{i-1}^{T}\mathbf{A}_{i-1}^{T}\mathbf{a}_{i}^{{}} &  \label{Pi1FFAa2390ds3}\\
		{\bf{\Pi }}_{i - 1}(:,2:end)  =-{{\bf{F}}_{i - 1}}{\bf{F}}_{i - 1}^T{\bf{A}}_{i - 1}^T{\bf{\underline{A}}}_{k - i}^{}.  &  \label{Pi2TOend329090sd233}
	\end{subnumcases}
	Also substitute  (\ref{Fi2ffiiT3902}) and
	(\ref{Ai2Aiai320ds3ds4fs}) into ${\bf{F}}_i^T{\bf{A}}_i^T{\bf{\underline{A}}}_{k - i}$																																																																																																																																																																																							to obtain
	\begin{gather}
		{\bf{F}}_i^T{\bf{A}}_i^T{\bf{\underline{A}}}_{k - i}^{} = \left[ {\begin{array}{*{20}{c}}
				{\left( {\begin{array}{*{20}{c}}
							{{\bf{F}}_{i - 1}^T}&{{\bf{0}}_{i - 1}^{}}
					\end{array}} \right)}\\
				{{\bf{f}}_i^T}
		\end{array}} \right]\left[ {\begin{array}{*{20}{c}}
				{{\bf{A}}_{i - 1}^T}\\
				{{\bf{a}}_i^T}
		\end{array}} \right]{\bf{\underline{A}}}_{k - i}^{}   \notag \\
		= \left[ {\begin{array}{*{20}{c}}
				{{\bf{F}}_{i - 1}^T{\bf{A}}_{i - 1}^T{\bf{\underline{A}}}_{k - i}^{}}\\
				{{\bf{f}}_i^T{{\left[ {\begin{array}{*{20}{c}}
									{{\bf{A}}_{i - 1}^{}}&{{\bf{a}}_i^{}}
							\end{array}} \right]}^T}{\bf{\underline{A}}}_{k - i}^{}}
		\end{array}} \right]  \notag \\
		= \left[ {\begin{array}{*{20}{c}}
				{{\bf{F}}_{i - 1}^T{\bf{A}}_{i - 1}^T{\bf{\underline{A}}}_{k - i}^{}}\\
				{{\bf{f}}_i^T{\bf{A}}_i^T{\bf{\underline{A}}}_{k - i}^{}}
		\end{array}} \right],  \label{FAA2FAa1290s}
	\end{gather}																																																													and
	then
	substitute (\ref{FAA2FAa1290s}) and
	(\ref{Fi2ffiiT3902})
	into
	(\ref{pi2FFAA2390923})
	to obtain
	\begin{gather}
		{{\bf{\Pi }}_i} = - \left[ {\begin{array}{*{20}{c}}
				{\left( {\begin{array}{*{20}{c}}
							{{{\bf{F}}_{i - 1}}}\\
							{{\bf{0}}_{i - 1}^T}
					\end{array}} \right)}&{{\bf{f}}_i^{}}
		\end{array}} \right]
		\left[ {\begin{array}{*{20}{c}}
				{{\bf{F}}_{i - 1}^T{\bf{A}}_{i - 1}^T{\bf{\underline{A}}}_{k - i}^{}}\\
				{{\bf{f}}_i^T{\bf{A}}_i^T{\bf{\underline{A}}}_{k - i}^{}}
		\end{array}} \right]  \notag \\
		= \left( {\begin{array}{*{20}{c}}
				-{{{\bf{F}}_{i - 1}}{\bf{F}}_{i - 1}^T{\bf{A}}_{i - 1}^T{\bf{\underline{A}}}_{k - i}^{}}\\
				{{\bf{0}}_{k - i}^T}
		\end{array}} \right) - {\bf{f}}_i^{}{\bf{f}}_i^T{\bf{A}}_i^T{\bf{\underline{A}}}_{k - i}^{}. \label{PiFFAf3298sd23nh}
	\end{gather}
	Finally, substitute (\ref{Pi2TOend329090sd233})
	into  (\ref{PiFFAf3298sd23nh})  to obtain
	\begin{equation}\label{FinalUpdatePi19sd3}
		{{\bf{\Pi }}_i} = \left( {\begin{array}{*{20}{c}}
				{\bf{\Pi }}_{i - 1}(:,2:end) \\
				{{\bf{0}}_{k - i}^T}
		\end{array}} \right) - {\bf{f}}_i^{}{\bf{f}}_i^T{\bf{A}}_i^T{\bf{\underline{A}}}_{k - i}^{}.
	\end{equation}

	To update ${{\mathbf{\Xi }}_{k-(i-1)}}$ into ${{\mathbf{\Xi }}_{k-i}}$ efficiently,
	substitute  (\ref{FAA2FAa1290s}) into (\ref{Xi2AAFFAA23094d3})  to obtain
	\begin{footnotesize}
		\begin{align}
			&{{\mathbf{\Xi }}_{k-i}}={\bf{\underline{A}}}_{k - i}^T{\bf{\underline{A}}}_{k - i}^{} + \lambda {\bf{I}} - {\bf{\underline{A}}}_{k - i}^T{\bf{A}}_i^{}{{\bf{F}}_i}{\bf{F}}_i^T{\bf{A}}_i^T{\bf{\underline{A}}}_{k - i}^{}   \notag \\
			&= {\bf{\underline{A}}}_{k - i}^T{\bf{\underline{A}}}_{k - i}^{} + \lambda {\bf{I}} -
			{\left[ {\begin{array}{*{20}{c}}
						{{\bf{F}}_{i - 1}^T{\bf{A}}_{i - 1}^T{\bf{\underline{A}}}_{k - i}^{}}\\
						{{\bf{f}}_i^T{\bf{A}}_i^T{\bf{\underline{A}}}_{k - i}^{}}
				\end{array}} \right]^T}\left[ {\begin{array}{*{20}{c}}
					{{\bf{F}}_{i - 1}^T{\bf{A}}_{i - 1}^T{\bf{\underline{A}}}_{k - i}^{}}\\
					{{\bf{f}}_i^T{\bf{A}}_i^T{\bf{\underline{A}}}_{k - i}^{}}
			\end{array}} \right]   \notag \\
			&= {\bf{\underline{A}}}_{k - i}^T{\bf{\underline{A}}}_{k - i}^{} + \lambda {\bf{I}} - {\bf{\underline{A}}}_{k - i}^T{\bf{A}}_{i - 1}^{}{\bf{F}}_{i - 1}^{}{\bf{F}}_{i - 1}^T{\bf{A}}_{i - 1}^T{\bf{\underline{A}}}_{k - i}^{}  \notag \\
			&\quad \quad  \quad  \quad \quad  \quad  \quad \quad \quad \quad  \quad  \quad \quad  \quad  \quad  - {\bf{\underline{A}}}_{k - i}^T{\bf{A}}_i^{}{\bf{f}}_i^{}{\bf{f}}_i^T{\bf{A}}_i^T{\bf{\underline{A}}}_{k - i}^{},  \label{F3490dfkDFK4ds23c23a}
		\end{align}
	\end{footnotesize}%
	and substitute (\ref{Ak2aA389323}) and  (\ref{FAA2FAaAFA23233}) into (\ref{Xi2AAFFAA23094d3}) with  $i=i-1$  to obtain
	
	\begin{footnotesize}
		\begin{align}
			&{{\mathbf{\Xi }}_{k-i+1}}= {\bf{\underline{A}}}_{k - i + 1}^T{\bf{\underline{A}}}_{k - i + 1}^{} + \lambda {\bf{I}}   \notag \\
			&\quad \quad  \quad  \quad \quad  \quad  \quad \quad     -{\bf{\underline{A}}}_{k - i + 1}^T{\bf{A}}_{i - 1}^{}{{\bf{F}}_{i - 1}}{\bf{F}}_{i - 1}^T{\bf{A}}_{i - 1}^T{\bf{\underline{A}}}_{k - i + 1}^{}  \notag \\
			&= {\left[ {\begin{array}{*{20}{c}}
						{{\bf{a}}_i^{}}&{{\bf{\underline{A}}}_{k - i}^{}}
				\end{array}} \right]^T}\left[ {\begin{array}{*{20}{c}}
					{{\bf{a}}_i^{}}&{{\bf{\underline{A}}}_{k - i}^{}}
			\end{array}} \right] + \lambda {\bf{I}}   \notag \\
			&\quad \quad  \quad  \quad \quad  \quad  \quad \quad   -  {\left[ {\begin{array}{*{20}{c}}
						{{\bf{F}}_{i - 1}^T{\bf{A}}_{i - 1}^T{\bf{a}}_i^{}}&{{\bf{F}}_{i - 1}^T{\bf{A}}_{i - 1}^T{\bf{\underline{A}}}_{k - i}^{}}
				\end{array}} \right]^T}   \notag \\
			&\quad \quad  \quad  \quad \quad  \quad  \quad \quad   \times \left[ {\begin{array}{*{20}{c}}
					{{\bf{F}}_{i - 1}^T{\bf{A}}_{i - 1}^T{\bf{a}}_i^{}}&{{\bf{F}}_{i - 1}^T{\bf{A}}_{i - 1}^T{\bf{\underline{A}}}_{k - i}^{}}
			\end{array}} \right]   \notag \\
			& = \left[ {\begin{array}{*{20}{c}}
					{{\bf{a}}_i^T{\bf{a}}_i^{} + \lambda }&{{\bf{a}}_i^T{\bf{\underline{A}}}_{k - i}^{}}\\
					{{\bf{\underline{A}}}_{k - i}^T{\bf{a}}_i^{}}&{{\bf{\underline{A}}}_{k - i}^T{\bf{\underline{A}}}_{k - i}^{} + \lambda {\bf{I}}}
			\end{array}} \right] -   \notag \\
			&\quad \left[ {\begin{array}{*{20}{c}}
					{\left( \begin{array}{l}
							{\bf{a}}_i^T{\bf{A}}_{i - 1}^{}{\bf{F}}_{i - 1}^{} \times \\
							{\bf{F}}_{i - 1}^T{\bf{A}}_{i - 1}^T{\bf{a}}_i^{}
						\end{array} \right)}&{\left( \begin{array}{l}
							{\bf{a}}_i^T{\bf{A}}_{i - 1}^{}{\bf{F}}_{i - 1}^{} \times \\
							{\bf{F}}_{i - 1}^T{\bf{A}}_{i - 1}^T{\bf{\underline{A}}}_{k - i}^{}
						\end{array} \right)}\\
					{\left( \begin{array}{l}
							{\bf{\underline{A}}}_{k - i}^T{\bf{A}}_{i - 1}^{}{\bf{F}}_{i - 1}^{} \times \\
							{\bf{F}}_{i - 1}^T{\bf{A}}_{i - 1}^T{\bf{a}}_i^{}
						\end{array} \right)}&{\left( \begin{array}{l}
							{\bf{\underline{A}}}_{k - i}^T{\bf{A}}_{i - 1}^{}{\bf{F}}_{i - 1}^{} \times \\
							{\bf{F}}_{i - 1}^T{\bf{A}}_{i - 1}^T{\bf{\underline{A}}}_{k - i}^{}
						\end{array} \right)}
			\end{array}} \right] \notag \\
			&=\left[ {\begin{array}{*{20}{c}}
					{\left( \begin{array}{l}
							{\bf{a}}_i^T{\bf{a}}_i^{} + \lambda  - \\
							{\bf{a}}_i^T{\bf{A}}_{i - 1}^{}{\bf{F}}_{i - 1}^{} \times \\
							{\bf{F}}_{i - 1}^T{\bf{A}}_{i - 1}^T{\bf{a}}_i^{}
						\end{array} \right)}&{\left( \begin{array}{l}
							{\bf{a}}_i^T{\bf{\underline{A}}}_{k - i}^{} - \\
							{\bf{a}}_i^T{\bf{A}}_{i - 1}^{}{\bf{F}}_{i - 1}^{} \times \\
							{\bf{F}}_{i - 1}^T{\bf{A}}_{i - 1}^T{\bf{\underline{A}}}_{k - i}^{}
						\end{array} \right)}\\
					{\left( \begin{array}{l}
							{\bf{\underline{A}}}_{k - i}^T{\bf{a}}_i^{} - \\
							{\bf{\underline{A}}}_{k - i}^T{\bf{A}}_{i - 1}^{}{\bf{F}}_{i - 1}^{}\\
							\times {\bf{F}}_{i - 1}^T{\bf{A}}_{i - 1}^T{\bf{a}}_i^{}
						\end{array} \right)}&{\left( \begin{array}{l}
							{\bf{\underline{A}}}_{k - i}^T{\bf{\underline{A}}}_{k - i}^{} + \lambda {\bf{I}}\\
							- {\bf{\underline{A}}}_{k - i}^T{\bf{A}}_{i - 1}^{}{\bf{F}}_{i - 1}^{}\\
							\times {\bf{F}}_{i - 1}^T{\bf{A}}_{i - 1}^T{\bf{\underline{A}}}_{k - i}^{}
						\end{array} \right)}
			\end{array}} \right].  \label{AAFFAA2390kd4d34}
		\end{align}
	\end{footnotesize}%
	It can be seen from (\ref{AAFFAA2390kd4d34}) that	
	${{\mathbf{\Xi }}_{k-i+1}}$ with the first row and column removed is
	\begin{footnotesize}
		\begin{equation*}\label{XikikiAAiFFAA2383}
			{{\mathbf{\Xi }}_{k-i+1}^{[-1,-1]}}= {\bf{\underline{A}}}_{k - i}^T{\bf{\underline{A}}}_{k - i}^{} + \lambda {\bf{I}} -
			{\bf{\underline{A}}}_{k - i}^T{\bf{A}}_{i - 1}^{}{\bf{F}}_{i - 1}^{}{\bf{F}}_{i - 1}^T{\bf{A}}_{i - 1}^T{\bf{\underline{A}}}_{k - i}^{},
		\end{equation*}
	\end{footnotesize}%
	which is substituted into (\ref{F3490dfkDFK4ds23c23a}) to compute ${{\mathbf{\Xi }}_{k-i}}$ efficiently by
	\begin{equation}\label{XiXi2390A032ds}
		{{\mathbf{\Xi }}_{k-i}}={{\mathbf{\Xi }}_{k-i+1}^{[-1,-1]}} - {\bf{\underline{A}}}_{k - i}^T{\bf{A}}_i^{}{\bf{f}}_i^{}{\bf{f}}_i^T{\bf{A}}_i^T{\bf{\underline{A}}}_{k - i}^{}.
	\end{equation}
	
	Firstly, let us deduce (\ref{fiifli2Pi3902sdBoth}).
	We can obtain
	(\ref{f1i1i2gPi39d234d3}) by substituting  (\ref{Pi1FFAa2390ds3}) into (\ref{colNewWhole32099d3bbb}).
	On the other hand,
	we  deduce																																																																																																																	
	\begin{equation}\label{Xi2aLambdaAFFA2390s3}
		{{\bf{\Xi }}_{k - i + 1}}(1,1) = {\bf{a}}_i^T{\bf{a}}_i^{} + \lambda  - {\bf{a}}_i^T{\bf{A}}_{i - 1}^{}{\bf{F}}_{i - 1}^{}{\bf{F}}_{i - 1}^T{\bf{A}}_{i - 1}^T{\bf{a}}_i^{}
	\end{equation}
	from (\ref{AAFFAA2390kd4d34}),																																																																																																																		
	and then substitute (\ref{Xi2aLambdaAFFA2390s3})
	into 
	(\ref{colNewWhole32099d3aaa})
	to obtain (\ref{fii2sqrt2390sd23}).																																																																																																												
	Lastly, let us deduce (\ref{XiXi2390A032dsModify}) and  (\ref{FinalPi19sd3SimplifiedMore}).
	From (\ref{Fi2ffiiT3902}), we  write
	\begin{equation}\label{Fi2ffiiT3902new934ds}
		{{\bf{f}}_{i}}=\left[ {\begin{array}{*{20}{c}}
				{\bf{\bar f}}_{i}^T & f_{ii}
		\end{array}} \right]^T,
	\end{equation}
	into which
	substitute (\ref{colNewWhole32099d3bbb})
	to obtain
	\begin{equation}\label{tgfff239sd23New32a3a}
		{{\bf{f}}_{i}}=\left[ {\begin{array}{*{20}{c}}
				- f_{ii} ({{\bf{F}}_{i-1}}{\bf{F}}_{i-1}^T {\bf{A}}_{i - 1}^T{\bf{a}}_i)^T& f_{ii}
		\end{array}} \right]^T.
	\end{equation}
	Then substitute (\ref{tgfff239sd23New32a3a})
	and
	(\ref{Ai2Aiai320ds3ds4fs})
	into
	${\bf{f}}_i^T{\bf{A}}_i^T{\bf{\underline{A}}}_{k - i}^{}$
	in (\ref{XiXi2390A032ds}) and (\ref{FinalUpdatePi19sd3}),
	to
	obtain
	\begin{multline}\label{fAA2gFFA230ks23ds23ds23}
		{\bf{f}}_i^T{\bf{A}}_i^T{\bf{\underline{A}}}_{k - i}^{} \\
		= {\left[ {\begin{array}{*{20}{c}}
					{ -f_{ii} {\bf{F}}_{i - 1}^{}{\bf{F}}_{i - 1}^T{\bf{A}}_{i - 1}^T{\bf{a}}_i^{}}\\
					f_{ii}
			\end{array}} \right]^T}{\left[ {\begin{array}{*{20}{c}}
					{{\bf{A}}_{i - 1}^{}}&{{\bf{a}}_i^{}}
			\end{array}} \right]^T}{\bf{\underline{A}}}_{k - i}^{} \\
		= f_{ii}\left( {{\bf{a}}_i^T{\bf{\underline{A}}}_{k - i}^{} - {\bf{a}}_i^T{\bf{A}}_{i - 1}^{}{\bf{F}}_{i - 1}^{}{\bf{F}}_{i - 1}^T{\bf{A}}_{i - 1}^T{\bf{\underline{A}}}_{k - i}^{}} \right).
	\end{multline}
	Moreover,
	let us deduce
	\begin{multline}\label{XiaAaAFFAA390sd23d3}
		{{\bf{\Xi }}_{k - i + 1}}(1,2:end) = \\
		{\bf{a}}_i^T{\bf{\underline{A}}}_{k - i}^{} - {\bf{a}}_i^T{\bf{A}}_{i - 1}^{}{\bf{F}}_{i - 1}^{}{\bf{F}}_{i - 1}^T{\bf{A}}_{i - 1}^T{\bf{\underline{A}}}_{k - i}^{}
	\end{multline}
	from (\ref{AAFFAA2390kd4d34}),
	and
	substitute (\ref{XiaAaAFFAA390sd23d3})
	into  (\ref{fAA2gFFA230ks23ds23ds23})
	to obtain
	\begin{equation}\label{fAA2gFFA230ks23ds23ds23fAA38d3}
		{\bf{f}}_i^T{\bf{A}}_i^T{\bf{\underline{A}}}_{k - i}^{}
		= f_{ii} {{\bf{\Xi }}_{k - i + 1}}(1,2:end).
	\end{equation}
	Finally, we substitute (\ref{fAA2gFFA230ks23ds23ds23fAA38d3}) into (\ref{XiXi2390A032ds})  to further simplify it into
	(\ref{XiXi2390A032dsModify}).
	On the other hand,
	we substitute
	(\ref{f1i1i2gPi39d234d3})
	into
	(\ref{Fi2ffiiT3902new934ds})
	to obtain
	\begin{equation}\label{tgfff239sd23New239sa}
		{{\bf{f}}_{i}}=\left[ {\begin{array}{*{20}{c}}
				f_{ii} {{\bf{\Pi }}_{i - 1}}(:,1)^T & f_{ii}
		\end{array}} \right]^T,
	\end{equation}
	and then substitute (\ref{tgfff239sd23New239sa}) and  (\ref{fAA2gFFA230ks23ds23ds23fAA38d3}) into
	(\ref{FinalUpdatePi19sd3})
	to further simplify it into
	${{\bf{\Pi }}_i} = \left( {\begin{array}{*{20}{c}}
			{\bf{\Pi }}_{i - 1}(:,2:end)  \\
			{{\bf{0}}_{k - i}^T}
	\end{array}} \right) - \left[ {\begin{array}{*{20}{c}}
			{  {f_{ii}}{{\bf{\Pi }}_{i - 1}}(:,1)}\\
			{{f_{ii}}}
	\end{array}} \right] f_{ii} {{\bf{\Xi }}_{k - i + 1}}(1,2:end)$,
	i.e.,  (\ref{FinalPi19sd3SimplifiedMore}).

	\section{The Inverse LDL$^T$ Factorization and the Corresponding Parallelization}

	To avoid the square-root operation in
	(\ref{InvCholSaveMem3290ccc}), we can use the inverse LDL$^T$ factorization instead of the inverse Cholesky factorization.
	The inverse LDL$^T$ factors of
	${{\mathbf{R}}_i}$ are the unit upper-triangular
	${\mathbf{L}}_i$ and the diagonal ${\mathbf{D}}_i$ satisfying
	\begin{equation}\label{LDMtFactorsDefine398da32}
		{\bf{L}}_i^{}{\bf{D}}_i^{}{\bf{L}}_i^T = {\bf{R}}_i^{ - 1},
	\end{equation}
	from which
	we can deduce
	\begin{equation}\label{MinvDL2Re4332}
		{\bf{L}}_i^{ - T}{\bf{D}}_i^{ - 1}{\bf{L}}_i^{ - 1} = {\bf{R}}_i^{}.
	\end{equation}
	Notice that in (\ref{MinvDL2Re4332}),
	the unit lower-triangular  ${\bf{L}}_i^{ - T}$
	and the diagonal ${\bf{D}}_i^{ - 1}$ are the conventional
	LDL$^T$ factors~\cite{Matrix_Computations_book} of ${\bf{R}}_i$.

	\subsection{The Block Inverse LDL$^T$ Factorization}
	Obviously,
	the upper-triangular
	${\mathbf{L}}_k$ and the diagonal ${\mathbf{D}}_k$
	can be written
	as
	\begin{subnumcases}{\label{LDMdefineAll313943}}
		{{\mathbf{L}}_{k}}\text{=}\left[ \begin{matrix}
			{{\mathbf{L}}_{i}} & \mathbf{\Gamma}  \\
			\mathbf{0} & \mathbf{\Delta}     \\
		\end{matrix} \right]  &  \label{L2LAF32413}\\
		{{\mathbf{D}}_{k}}=\left[ \begin{matrix}
			{{\mathbf{D}}_{i}} & \mathbf{0}  \\
			\mathbf{0} & \mathbf{\Phi}    \\
		\end{matrix} \right].   &  \label{D2DG3243}
	\end{subnumcases}
	In this subsection, we will deduce an efficient algorithm to
	compute  $\mathbf{\Gamma}$, $\mathbf{\Delta}$ and  $\mathbf{\Phi}$  in (\ref{LDMdefineAll313943})
	by
	\begin{subnumcases}{\label{ABFGEcomputeE84r9834}}
		\mathbf{\Gamma}=-\mathbf{L}_{i}^{{}}\mathbf{D}_{i}^{{}}\mathbf{L}_{i}^{T}\mathbf{P}\mathbf{\Delta}   &  \label{A2LFMVF8430}\\
		\mathbf{\Delta} \mathbf{\Phi}{\mathbf{\Delta}  ^{T}}={{(\mathbf{U}-{{\mathbf{P}}^{T}}\mathbf{L}_{i}^{{}}\mathbf{D}_{i}^{{}}\mathbf{L}_{i}^{T}\mathbf{P})}^{-1}},  &  \label{LDLinvFreeNov17b}
	\end{subnumcases}
	which can be regarded as a special case of the inverse LDM$^T$ factorization introduced in \cite{BLASTLULDLpresented32ds}.
	Notice that in (\ref{LDLinvFreeNov17b}),  the unit  upper-triangular $\mathbf{\Delta}$  and  the diagonal $\mathbf{\Phi}$
	are the inverse
	$\mathbf{LD}\mathbf{L}^T$  factors of $\mathbf{U}-{{\mathbf{P}}^{T}}\mathbf{L}_{i}^{{}}\mathbf{D}_{i}^{{}}\mathbf{L}_{i}^{T}\mathbf{P}$.
	The derivation of
	(\ref{ABFGEcomputeE84r9834}) is in the next paragraph. 

	From (\ref{LDMdefineAll313943}) we can deduce
	$\mathbf{L}_{k}^{-1}\text{=}\left[ \begin{matrix}
		\mathbf{L}_{i}^{-1} & -\mathbf{L}_{i}^{-1}\mathbf{\Gamma}{{\mathbf{\Delta}   }^{-1}}  \\
		\mathbf{0} & {{\mathbf{\Delta}   }^{-1}}  \\
	\end{matrix} \right]$
	and
	$\mathbf{D}_{k}^{-1}=\left[ \begin{matrix}
		\mathbf{D}_{i}^{-1} & \mathbf{0}  \\
		\mathbf{0} & {{\mathbf{\Phi}  }^{-1}}  \\
	\end{matrix} \right]$,
	which are substituted into (\ref{MinvDL2Re4332}) to obtain
	\begin{multline}\label{LDLinvFreeNov17a}
		\left[ \begin{matrix}
			\mathbf{L}_{i}^{-T} & \mathbf{0}  \\
			-{\mathbf{\Delta}  ^{-T}}{{\mathbf{\Gamma}}^{T}}\mathbf{L}_{i}^{-T} & {\mathbf{\Delta}  ^{-T}}  \\
		\end{matrix} \right]\left[ \begin{matrix}
			\mathbf{D}_{i}^{-1} & \mathbf{0}  \\
			\mathbf{0} & {{\mathbf{\Phi}  }^{-1}}  \\
		\end{matrix} \right]  \\
		\times
		\left[ \begin{matrix}
			\mathbf{L}_{i}^{-1} & -\mathbf{L}_{i}^{-1}\mathbf{\Gamma}{{\mathbf{\Delta}   }^{-1}}  \\
			\mathbf{0} & {{\mathbf{\Delta}   }^{-1}}  \\
		\end{matrix} \right]=\mathbf{R}_{k}^{{}}.
	\end{multline}
	Then let us  substitute (\ref{R_def_perhaps_No_R1apdx9ds23}) into (\ref{LDLinvFreeNov17a}) to  deduce
	\begin{multline}\label{}
		\left[ {\begin{array}{*{20}{c}}
				{{\bf{L}}_i^{ - T}{\bf{D}}_i^{ - 1}{\bf{L}}_i^{ - 1}}&{ - {\bf{L}}_i^{ - T}{\bf{D}}_i^{ - 1}{\bf{L}}_i^{ - 1}{\mathbf{\Gamma}}{{\mathbf{\Delta}  }^{ - 1}}}\\
				{\left( \begin{array}{l}
						{\rm{ - }}{{\mathbf{\Delta}}^{ - T}} \times \\
						{{\mathbf{\Gamma}}^T}{\bf{L}}_i^{ - T} \times \\
						{\bf{D}}_i^{ - 1}{\bf{L}}_i^{ - 1}
					\end{array} \right)}&{\left( \begin{array}{l}
						{{\mathbf{\Delta}}^{ - T}}{{\mathbf{\Gamma}}^T}{\bf{L}}_i^{ - T} \times \\
						{\bf{D}}_i^{ - 1}{\bf{L}}_i^{ - 1}{\mathbf{\Gamma}}{{\mathbf{\Delta}  }^{ - 1}} + \\
						{{\mathbf{\Delta}}^{ - T}}{{\mathbf{\Phi}}^{ - 1}}{{\mathbf{\Delta}}^{ - 1}}
					\end{array} \right)}
		\end{array}} \right] \\
		= \left[ {\begin{array}{*{20}{c}}
				{{{\bf{R}}_i}}&{\bf{P}}\\
				{{{\bf{P}}^T}}&{\bf{U}}
		\end{array}} \right],
	\end{multline}
	from which we can obtain
	\begin{subnumcases}{\label{MEYEBMtotal312405}}
		-\mathbf{L}_{i}^{{\text{-}}T}\mathbf{D}_{i}^{{\text{-}}1}\mathbf{L}_{i}^{{\text{-}}1}\mathbf{\Gamma}{{\mathbf{\Delta}   }^{{\text{-}}1}}=\mathbf{P} &  \label{MDLAF2V32143}\\
		{\mathbf{\Delta}  ^{{\text{-}}T}}{{\mathbf{\Gamma}}^{T}}\mathbf{L}_{i}^{{\text{-}}T}\mathbf{D}_{i}^{{\text{-}}1}\mathbf{L}_{i}^{{\text{-}}1}\mathbf{\Gamma}
		{{\mathbf{\Delta}   }^{{\text{-}}1}}+{\mathbf{\Delta}  ^{{\text{-}}T}}{{\mathbf{\Phi}  }^{{\text{-}}1}}{{\mathbf{\Delta}   }^{{\text{-}}1}}=\mathbf{U}.    &  \label{EBMDLAFEGF2T438}
	\end{subnumcases}
	Finally,
	write (\ref{MDLAF2V32143}) as  (\ref{A2LFMVF8430}),
	which is then substituted
	into  (\ref{EBMDLAFEGF2T438})
	to get
	${{\mathbf{P}}^{T}}\mathbf{L}_{i}^{{}}\mathbf{D}_{i}^{{}}\mathbf{L}_{i}^{T}\mathbf{P}+{\mathbf{\Delta}  ^{-T}}{{\mathbf{\Phi}  }^{-1}}{{\mathbf{\Delta}   }^{-1}}=\mathbf{U}$,
	i.e.,  (\ref{LDLinvFreeNov17b}).

	\subsection{The Memory-Saving Parallel Implementation of the Inverse LDL$^T$ Factorization}

	Let us substitute
	(\ref{LDMtFactorsDefine398da32})
	into
	(\ref{L_m_def12431RidgeInvJun30JunY24})
	to obtain
	\begin{equation}\label{FF2LDL032dasasr3ed3}
		{{\bf{F}}_i}{{\bf{F}}_i^{T}}={\bf{L}}_i^{}{\bf{D}}_i^{}{\bf{L}}_i^T.
	\end{equation}
	On the other hand,
	we can write
	the unit upper-triangular
	${\mathbf{L}}_i$ and the diagonal ${\mathbf{D}}_i$ as~\footnote{The following equation (\ref{LLDDsmaller932sd23}) has been presented in \cite{zhfVTC2010DivFree}.}
	\begin{subnumcases}{\label{LLDDsmaller932sd23}}
		{{\bf{L}}_i} = \left[ {\begin{array}{*{20}{c}}
				{{{\bf{L}}_{i - 1}}}&{{{{\bf{\bar l}}}_i}}\\
				{{\bf{0}}_{i - 1}^T}&1
		\end{array}} \right] &  \label{LLDDsmaller932sd23aaa}\\
		{{\bf{D}}_i} = \left[ {\begin{array}{*{20}{c}}
				{{{\bf{D}}_{i - 1}}}&{{\bf{0}}_{i - 1}^{}}\\
				{{\bf{0}}_{i - 1}^T}&{{d_{i}}}
		\end{array}} \right],  &  \label{LLDDsmaller932sd23bbb}
	\end{subnumcases}
	where ${{\bf{\bar l}}}_i$ denotes the $i^{th}$ column of ${{\bf{L}}_i}$
	with the last entry removed, and ${{d_{i}}}$ denotes the $i^{th}$ diagonal entry of
	the diagonal ${{\bf{D}}_i}$.
	Then substitute (\ref{LLDDsmaller932sd23})
	and (\ref{Fi2ffiiT3902}) into (\ref{FF2LDL032dasasr3ed3})
	to obtain
	$\left[ {\begin{array}{*{20}{c}}
			{{{\bf{L}}_{i - 1}}}&{{{{\bf{\bar l}}}_i}}\\
			{{\bf{0}}_{i - 1}^T}&1
	\end{array}} \right]\left[ {\begin{array}{*{20}{c}}
			{{{\bf{D}}_{i - 1}}}&{{\bf{0}}_{i - 1}^{}}\\
			{{\bf{0}}_{i - 1}^T}&{{d_{i}}}
	\end{array}} \right]{\left[ {\begin{array}{*{20}{c}}
				{{{\bf{L}}_{i - 1}}}&{{{{\bf{\bar l}}}_i}}\\
				{{\bf{0}}_{i - 1}^T}&1
		\end{array}} \right]^T} = \left[ {\begin{array}{*{20}{c}}
			{{{\bf{F}}_{i - 1}}}&{{{{\bf{\bar f}}}_i}}\\
			{{\bf{0}}_{i - 1}^T}&{{f_{ii}}}
	\end{array}} \right]{\left[ {\begin{array}{*{20}{c}}
				{{{\bf{F}}_{i - 1}}}&{{{{\bf{\bar f}}}_i}}\\
				{{\bf{0}}_{i - 1}^T}&{{f_{ii}}}
		\end{array}} \right]^T}$, from which we can deduce
	\begin{subnumcases}{\label{dlf3290as23}}
		d_{i}^{} = f_{ii}^2  &  \label{dlf3290as23d} \\
		{{\bf{\bar l}}_i} = {{\bf{\bar f}}_i}/{f_{ii}}.  &  \label{dlf3290as23lLi}
	\end{subnumcases}

	We can substitute (\ref{f1i1i2gPi39d234d3})
	into
	(\ref{dlf3290as23lLi})
	to obtain
	\begin{equation}\label{li2Pi3920sd3d}
		{{\bf{\bar l}}_i} =   {{\bf{\Pi }}_{i - 1}}(:,1).
	\end{equation}
	Moreover,  substitute (\ref{fii2sqrt2390sd23})  into
	(\ref{dlf3290as23d}), (\ref{XiXi2390A032dsModify}) and
	(\ref{FinalPi19sd3SimplifiedMore}) to obtain
	\begin{equation}\label{dii2Xi390ads2d32}
		d_{i}^{}  = 1/{{\bf{\Xi }}_{k - i + 1}}(1,1),
	\end{equation}
	\begin{equation}\label{Xi2k409043fd}
		{{\bf{\Xi }}_{k - i}} = {\bf{\Xi }}_{k - i + 1}^{[ - 1, - 1]} - \frac{{{{\bf{\Xi }}_{k - i + 1}}{{(1,2:end)}^T}{{\bf{\Xi }}_{k - i + 1}}(1,2:end)}}{{{{\bf{\Xi }}_{k - i + 1}}(1,1)}}
	\end{equation}
	and
	\begin{equation}\label{Pi2k9430934ds3}
		{{\bf{\Pi }}_i} = \left[ {\begin{array}{*{20}{c}}
				{{{\bf{\Pi }}_{i - 1}}(:,2:end) - \frac{{{{\bf{\Pi }}_{i - 1}}(:,1){{\bf{\Xi }}_{k - i + 1}}(1,2:end)}}{{{{\bf{\Xi }}_{k - i + 1}}(1,1)}}}\\
				- {\frac{{{{\bf{\Xi }}_{k - i + 1}}(1,2:end)}}{{{{\bf{\Xi }}_{k - i + 1}}(1,1)}}}
		\end{array}} \right],
	\end{equation}
	respectively.

	Since the unit upper-triangular
	${\mathbf{L}}_1 =1$, it can be seen from (\ref{LDMtFactorsDefine398da32}) with $k=1$ that
	the diagonal
	\begin{equation}\label{InitialD1daaded4}
		{\mathbf{D}}_1=d_{11}=1/\mathbf{R}(1,1).
	\end{equation}
	To save memories,  we store ${\mathbf{D}}_1=d_{11}$,
	${{\bf{\Xi }}_{k - 1}}$  and
	${{\mathbf{\Pi }}_{1}}$    in ${\bf{R}}(1,1)$, ${\bf{R}}(2:k,2:k)$
	and $\mathbf{R}(1,2:k)$, respectively, by computing
	\begin{subnumcases}{\label{InvLDLSaveMem3290Initial}}
		{\bf{R}}(1,1) = 1/{\bf{R}}(1,1)  &  \label{InvLDLSaveMem3290cccInitial}\\
		{\begin{split}
				{\bf{R}}(2:k,2:k) = {\bf{R}}(2:k,2:k)-  \quad \quad \quad \quad \quad   \quad \quad \\
				\quad \quad \quad \quad \quad \quad \quad \quad {\bf{R}}{(1,2:k)^T}{\bf{R}}(1,1){\bf{R}}(1,2:k)
		\end{split}}   &  \label{InvLDLSaveMem3290aaaInitial}  \\
		{\bf{R}}(1,2:k) = - {\bf{R}}(1,1){\bf{R}}(1,2:k),   &  \label{InvLDLSaveMem3290bbbInitial}
	\end{subnumcases}
	which are deduced in the rest part of this paragraph.
	We use the $i^{th}$ ($i=1,2,\cdots,k$) diagonal entry of ${\mathbf{R}}$ to
	store the $i^{th}$ diagonal entry of the diagonal ${\mathbf{D}}$. Accordingly, we compute and store
	${\mathbf{D}}_1=d_{11}$ in $\mathbf{R}(1,1)$ by (\ref{InvLDLSaveMem3290cccInitial}),
	which is then substituted into
	(\ref{i1Initial2SaveMem1})
	and
	(\ref{i1Initial2SaveMem2}) to obtain
	(\ref{InvLDLSaveMem3290aaaInitial}) and
	(\ref{InvLDLSaveMem3290bbbInitial}), respectively.

	In the $i^{th}$ ($i \ge 2$) iteration,
	we also save memories by storing
	${{\bf{\bar l}}_i}$ in ${\bf{R}}(1:i - 1,i)$, i.e.,
\begin{equation}\label{li2Pi3920sd3d2R}
	{\bf{R}}(1:i - 1,i)={{\bf{\bar l}}_i},
\end{equation}
and storing $d_{i}$,
${{\mathbf{\Xi }}_{k-i}}$ and  ${{\mathbf{\Pi }}_{i}}$
in $\mathbf{R}(i,i)$,  ${\bf{R}}(i + 1:k,i + 1:k)$ and
${\bf{R}}(1:i,i + 1:k)$, respectively,
by
\begin{small}
	\begin{subnumcases}{\label{InvLDLSaveMem3290}}
		{\bf{R}}(i,i) = 1/{\bf{R}}(i,i)  &  \label{InvLDLSaveMem3290ccc}\\
		{\begin{split}
				{\bf{R}}(i + 1:k,i + 1:k) = {\bf{R}}(i + 1:k,i + 1:k)-     \\
				{\bf{R}}{(i,i + 1:k)^T}{\bf{R}}(i,i){\bf{R}}(i,i + 1:k)
		\end{split}}   &  \label{InvLDLSaveMem3290aaa}  \\
		{\begin{split}
				{\bf{R}}(1:i,i + 1:k) = \quad \quad \quad \quad \quad \quad \quad \quad \quad \quad \quad \quad \quad \quad \\
				\left[ {\begin{array}{*{20}{c}}
						{\left( \begin{array}{l}
								{\bf{R}}(1:i - 1,i + 1:k) - {\bf{R}}(i,i)\\
								\times {\bf{R}}(1:i - 1,i){\bf{R}}(i,i + 1:k)
							\end{array} \right)}\\
						{ - {\bf{R}}(i,i){\bf{R}}(i,i + 1:k)}
				\end{array}} \right].
		\end{split}}   &  \label{InvLDLSaveMem3290bbb}
	\end{subnumcases}
\end{small}
The above  (\ref{li2Pi3920sd3d2R})
and
(\ref{InvLDLSaveMem3290}) are deduced in the next paragraph.

When $i=2$,
we can deduce (\ref{li2Pi3920sd3d2R}) and (\ref{InvLDLSaveMem3290ccc})   from (\ref{li2Pi3920sd3d})
and (\ref{dii2Xi390ads2d32}), respectively,
and then substitute
(\ref{InvLDLSaveMem3290ccc})
into
(\ref{InvCholSaveMem3290aaa}) and
(\ref{InvCholSaveMem3290bbb}) to obtain
(\ref{InvLDLSaveMem3290aaa})
and
(\ref{InvLDLSaveMem3290bbb}), respectively,
since
${{\bf{\Xi }}_{k - 1}}$  and
${{\mathbf{\Pi }}_{1}}$  have been stored  in
${\bf{R}}(2:k,2:k)$
and $\mathbf{R}(1,2:k)$, respectively.  
When $i \ge 3$,
we can also deduce (\ref{li2Pi3920sd3d2R}) and (\ref{InvLDLSaveMem3290ccc})   from (\ref{li2Pi3920sd3d})
and (\ref{dii2Xi390ads2d32}), respectively, and then substitute
(\ref{InvLDLSaveMem3290ccc})
into
(\ref{InvCholSaveMem3290aaa}) and
(\ref{InvCholSaveMem3290bbb}) to obtain
(\ref{InvLDLSaveMem3290aaa})
and
(\ref{InvLDLSaveMem3290bbb}), respectively,
since
${{\mathbf{\Xi }}_{k-(i-1)}}={{\mathbf{\Xi }}_{k-i+1}}$  and  ${{\mathbf{\Pi }}_{i-1}}$
have been  stored
in ${\bf{R}}((i-1) + 1:k,(i-1) + 1:k)={\bf{R}}(i:k,i:k)$ and
${\bf{R}}(1:i-1,(i-1) + 1:k)={\bf{R}}(1:i-1,i:k)$, respectively.
Thus we have verified  (\ref{li2Pi3920sd3d2R})
and
(\ref{InvLDLSaveMem3290}) for any $i=2,3,\cdots,k$.
Obviously,
(\ref{InvLDLSaveMem3290Initial}) is equivalent to (\ref{InvLDLSaveMem3290}) with $i=1$.
Then we can compute (\ref{InvLDLSaveMem3290}) iteratively for $i=1,2,\cdots,k$, to
cover the upper-triangular part of  ${\bf{R}}_k$  with
the unit upper-triangular
${\mathbf{L}}_k$ and the diagonal ${\mathbf{D}}_k$
satisfying ${\bf{L}}_k^{}{\bf{D}}_k^{}{\bf{L}}_k^T = {\bf{R}}_k^{ - 1}$.
\textbf{Algorithm 12} summarizes
the corresponding implementation.

\begin{algorithm}
	\caption{:~\bf The parallel implementation of the inverse $LDL^T$ factorization}
	\begin{algorithmic}[1]
		\Require The upper-triangular part of a  Hermitian matrix ${\bf{R}}$
		\Ensure The inverse $LDL^T$ factors of ${\bf{R}}$ in the upper-triangular part of  ${\bf{R}}$
		\For{$i=1:k$ ($k$ is the size of ${\bf{R}}$)}
		\State  ${\bf{R}}(i,i) = 1/{\bf{R}}(i,i);$
		\State ${\bf{R}}(i + 1:k,i + 1:k) = {\bf{R}}(i + 1:k,i + 1:k)-  \quad \quad \quad$
		${\bf{R}}{(i,i + 1:k)^T}{\bf{R}}(i,i){\bf{R}}(i,i + 1:k);$
		\State ${\bf{R}}(1:i,i + 1:k) =  \quad \quad \quad \quad \quad \quad \quad \quad \quad \quad \quad \quad \quad$
		$\footnotesize  \left[ {\begin{array}{*{20}{c}}
				{{\bf{R}}(1:i - 1,i + 1:k) - {\bf{R}}(1:i - 1,i){\bf{R}}(i,i){\bf{R}}(i,i + 1:k)}\\
				{ - {\bf{R}}(i,i){\bf{R}}(i,i + 1:k)}
		\end{array}} \right];$
		\EndFor
		\State ${\bf{D}} = {\mathop{\rm diag}\nolimits} \left( {{\mathop{\rm diag}\nolimits} ({\bf{R}})} \right)$,  i.e.,  the diagonal entries of  ${\bf{R}}$ form the diagonal matrix  ${\bf{D}}$;
		\State ${\bf{L}} = {\mathop{\rm triu}\nolimits} \left( {{\bf{R}} - {\bf{D}} + {{\bf{I}}_k}} \right)$, i.e.,  the upper-triangular part of ${\bf{R}}$ except the diagonal entries forms
		the unit upper-triangular matrix  ${\bf{L}}$;
	\end{algorithmic}
\end{algorithm}

\subsection{Parallel Performance of the Proposed Inverse LDL$^T$ Factorization}

The inverse LDL$^T$ factorization summarized in \textbf{Algorithm 12}
computes (\ref{InvLDLSaveMem3290}) iteratively for $i=1,2,\cdots,k$,
which can be  written  as
\begin{small}
	\begin{subnumcases}{\label{InvLDLSaveMem3290New2sd2}}
		{\bf{R}}(i,i) = 1/{\bf{R}}(i,i)  &  \label{InvLDLSaveMem3290cccNew2sd2}\\
		{\bf{\xi }}_{k - i}^T =  {\bf{R}}(i,i+1,k)   &  \label{InvLDLSaveMem3290dddNew2sd2}  \\
		{\bf{R}}(i,i+1,k) = - {\bf{R}}(i,i) {\bf{\xi }}_{k - i}^T   &  \label{InvLDLSaveMem3290aaaNew2sd2} \\
		{\begin{split}
				{\bf{R}}([1:i - 1,i + 1:k],i + 1:k) = \quad \quad \quad \quad \quad \quad   \\
				{\bf{R}}([1:i - 1,i + 1:k],i + 1:k) + \quad \quad \quad \quad \quad  \\
				\quad    {\left[ {\begin{array}{*{20}{c}}
							{{\bf{R}}{{(1:i - 1,i)}^T}}& {\bf{\xi }}_{k - i}^T
					\end{array}} \right]^T}(- {\bf{R}}(i,i) {\bf{\xi }}_{k - i}^T).
		\end{split}}   &  \label{InvLDLSaveMem3290bbbNew2sd2}
	\end{subnumcases}
\end{small}
It can be seen that the dominant computational complexity of (\ref{InvLDLSaveMem3290New2sd2}) comes from   (\ref{InvLDLSaveMem3290bbbNew2sd2}),
where all entries in columns $i+1$ to $k$ of the upper-triangular ${\bf{R}}$ except row $i$
are updated  with a multiplication and an addition~\footnote{We do not need  to compute $- {\bf{R}}(i,i) {\bf{\xi }}_{k - i}^T$ in (\ref{InvLDLSaveMem3290bbbNew2sd2}), since it has been computed in  (\ref{InvLDLSaveMem3290aaaNew2sd2})}.
Then the  dominant complexity of (\ref{InvLDLSaveMem3290New2sd2})  is $(i+1-1)+(i+2-1)+\cdots+(k-1)=(i+k-1)(k-i)/2$
multiplication and addition in the $i^{th}$ iteration, and is $\sum\limits_{i = 1}^k {\frac{{(i + k - 1)(k - i)}}{2}}  \approx \frac{{{k^3}}}{3}$ in total.
Accordingly, it can be seen that the proposed parallel implementation does not change the dominant computational complexity.


Let us assume a theoretical situation where the
parallel algorithm runs on $1+2+ \cdots + k=\frac{(k+1)k}{2}$ processors, i.e., each entry in the upper-triangular ${{\bf{R}}_k}$ is stored in an exclusive processor.
Then in (\ref{InvLDLSaveMem3290bbbNew2sd2}),
all entries in columns $i+1$ to $k$ of the upper-triangular ${\bf{R}}$ except row $i$
can be updated simultaneously, and each entry
is updated with
a multiplication and an addition executed serially.

We can also assume a situation where the
parallel algorithm runs on $k$ processors, and each column of the upper-triangular ${{\bf{R}}_k}$ is stored in an exclusive processor.
The corresponding parallel implementation is described in \textbf{Algorithm 13}.
As can be seen from \textbf{Algorithm 13},
${\bf{R}}{{(1:i,i)}}$
and
${\bf{\xi }}_{k - i}(1:j-i)$ are required for  the processor storing the $j^{th}$
($j=i+1,i+2,\cdots,k$)
column of the upper-triangular ${\bf{R}}$ to update that column.
The column vector ${\bf{R}}{{(1:i,i)}}$  is in the processor storing the $i^{th}$ column of ${\bf{R}}$,
while the $(j-i)$
entries in ${\bf{\xi }}_{k - i}^T(1:j-i) =  {\bf{R}}(i,i+1:j)$ are in the processors storing columns $i+1,i+2,\cdots,j$ of ${\bf{R}}$.
Accordingly,  ${\bf{R}}{{(1:i,i)}}, {\bf{R}}(i,i+1), {\bf{R}}(i,i+2),\cdots,{\bf{R}}(i,j-1)$
need to be transmitted to  the processor storing the $j^{th}$ column of  ${\bf{R}}$
from the processors storing the $i^{th}, (i+1)^{th}, (i+2)^{th}, \cdots, (j-1)^{th}$ columns of ${\bf{R}}$,
respectively.

\begin{algorithm}
	\caption{:~\bf The parallel implementation of the inverse $LDL^T$ factorization when each column of the upper-triangular ${{\bf{R}}_k}$ is stored in an exclusive processor}
	\begin{algorithmic}[1]
		\Require The upper-triangular part of a  Hermitian matrix ${\bf{R}}$
		\Ensure The inverse $LDL^T$ factors of ${\bf{R}}$ in the upper-triangular part of  ${\bf{R}}$
		\For{$i=1:k$ ($k$ is the size of ${\bf{R}}$)}
		\State  ${\bf{R}}(i,i) = 1/{\bf{R}}(i,i);$
		\State  ${\bf{\xi }}_{k - i}^T =  {\bf{R}}(i,i+1:k)$;
		\For{$j=i+1:k$}
		\State ${\bf{R}}(i,j) = - {\bf{R}}(i,i) {\bf{\xi }}_{k - i}(j-i)$;
		\State $\footnotesize  \begin{array}{*{20}{c}}
			{\bf{R}}([1:i - 1,i + 1:j],j) ={\bf{R}}([1:i - 1,i + 1:j],j) + \\
			{\left[ {\begin{array}{*{20}{c}}
						{{\bf{R}}{{(1:i - 1,i)}^T}}& {\bf{\xi }}_{k - i}(1:j-i)^T
				\end{array}} \right]^T}{\bf{R}}(i,j);
		\end{array}$
		\EndFor
		\EndFor
		\State ${\bf{D}} = {\mathop{\rm diag}\nolimits} \left( {{\mathop{\rm diag}\nolimits} ({\bf{R}})} \right)$,  i.e.,  the diagonal entries of  ${\bf{R}}$ form the diagonal matrix  ${\bf{D}}$;
		\State ${\bf{L}} = {\mathop{\rm triu}\nolimits} \left( {{\bf{R}} - {\bf{D}} + {{\bf{I}}_k}} \right)$, i.e.,  the upper-triangular part of ${\bf{R}}$ except the diagonal entries forms
		the unit upper-triangular matrix  ${\bf{L}}$;
	\end{algorithmic}
\end{algorithm}


Let us consider a general case where the columns of ${\bf{R}}$ are stored in $\tau \le k$ processors,
and processors $1,2,\cdots,\tau$
store
${\upsilon _1}, {\upsilon _2}, \cdots,  {\upsilon _\tau}$
columns of ${\bf{R}}$, respectively.
Assume that the columns stored in processors $1,2,\cdots,\tau$ are
columns $1:{\tilde \upsilon}_1,({\tilde \upsilon}_1+1):{\tilde \upsilon}_2,\cdots,({\tilde \upsilon}_{\tau-1}+1):{\tilde \upsilon}_\tau$
of   ${\bf{R}}$
with
\begin{equation}\label{ColumnPointAssume3s3}
	1 \le {\tilde \upsilon}_1 < {\tilde \upsilon}_2 < \cdots < {\tilde \upsilon}_{\tau-1} < {\tilde \upsilon}_\tau =k.
\end{equation}
Then it can be seen that
\begin{equation}\label{EachCPUnColumns239sd}
	{\upsilon _1} + {\upsilon _2} + \cdots + {\upsilon _\tau} = k,
\end{equation}
and columns $({\tilde \upsilon}_{\mu-1}+1):{\tilde \upsilon}_\mu$ of
${\bf{R}}$ are stored in
processor $\mu$ ($\mu = 1,2,\cdots,\tau$),
where  ${\tilde \upsilon}_{0}=0$ 
and
\begin{equation}\label{EachCPUnColumns239sdGeNi}
	{\upsilon _1} + {\upsilon _2} + \cdots + {\upsilon _\mu} = {\tilde \upsilon}_\mu.
\end{equation}
Now let us consider communication cost in the $i^{th}$ ($i=1,2,\cdots,k$) iteration of \textbf{Algorithm 13}.
When ${{\tilde \upsilon} _{\mu-1}}+1 \le i \le {{\tilde \upsilon} _{\mu}}$,
it can be seen that  ${\bf{R}}{{(1:i,i)}}$ and ${\bf{R}}(i,i+1:{{\tilde \upsilon} _\mu})$  in  processor $\mu$ need to be transmitted to processors $\mu+1,\mu+2,\cdots,\tau$.
Moreover,
${\bf{R}}(i,{{\tilde \upsilon} _{\varsigma-1}}+1:{{\tilde \upsilon} _{\varsigma}})$ in  processor $\varsigma$ needs to be  transmitted to processors $\varsigma+1,\varsigma+2,\cdots,\tau$,
where $\varsigma=\mu+1,\mu+2,\cdots,\tau-1$.

\subsection{Application of the Proposed Inverse LDL$^T$ Factorization to the Distributed BLS}

In this subsection,
we introduce the
distributed BLS that adopts
the inverse LDL$^T$ factorization
summarized in \textbf{Algorithm 12}.
To implement the distributed BLS with model-parallelism, we can  partition all ${\tilde k}_p$ feature  and
enhancement nodes into $p$ workers. Then  we have
\begin{equation}\label{Anp2AnH954734partition}
	\mathbf{ A}_{k}^{{}}=\left[ {\begin{array}{*{20}{c}}
			{{{\bf{A}}_{{\upsilon _1}}}}&{{{\bf{A}}_{{\upsilon _2}}}}& \cdots &{{{\bf{A}}_{{\upsilon _\tau}}}}
	\end{array}} \right],
\end{equation}
where ${{{\bf{A}}_{{\upsilon _\mu}}}}$ ($\mu=1,2,\cdots,\tau$) denotes the ${\upsilon _\mu}$ nodes partitioned into the $\mu^{th}$ worker.
Then substitute (\ref{Anp2AnH954734partition}) into  (\ref{R_define12321numdaRidge302923}) to write the upper-triangular blocks of ${{\bf{ R}}}_k$ as
\begin{multline}\label{R2BLK34d4f34}
	{\bf{R}}_k^{} = {\bf{A}}_k^T{\bf{A}}_k^{} + \lambda {\bf{I}} \\
	= \left[ {\begin{array}{*{20}{c}}
			{{\bf{A}}_{{\upsilon _1}}^T{{\bf{A}}_{{\upsilon _1}}}}&{{\bf{A}}_{{\upsilon _1}}^T{{\bf{A}}_{{\upsilon _2}}}}& \cdots &{{\bf{A}}_{{\upsilon _1}}^T{{\bf{A}}_{{\upsilon _\tau }}}}\\
			\times &{{\bf{A}}_{{\upsilon _2}}^T{{\bf{A}}_{{\upsilon _2}}}}& \cdots &{{\bf{A}}_{{\upsilon _2}}^T{{\bf{A}}_{{\upsilon _\tau }}}}\\
			\vdots & \vdots & \ddots & \vdots \\
			\times & \times & \cdots &{{\bf{A}}_{{\upsilon _\tau }}^T{{\bf{A}}_{{\upsilon _\tau }}}}
	\end{array}} \right] + \lambda {\bf{I}}.
\end{multline}
It can be seen from (\ref{R2BLK34d4f34}) that
${{{\bf{A}}_{{\upsilon _\mu}}}}$  in  worker $\mu$ need to be transmitted to workers $\mu+1,\mu+2,\cdots,\tau$, where $\mu=1,2,\cdots,\tau$,
and then ${\upsilon}_\mu$  columns of the upper-triangular
${\bf{R}}_k$,  i.e.,  columns $({\tilde \upsilon}_{\mu-1}+1):{\tilde \upsilon}_\mu$,  can be computed from ${{{\bf{A}}_{{\upsilon _1}}}}, {{{\bf{A}}_{{\upsilon _2}}}}, \cdots, {{{\bf{A}}_{{\upsilon _{\mu-1}}}}}, {{{\bf{A}}_{{\upsilon _{\mu}}}}}$ in worker $\mu$.

Now  workers $1,2,\cdots,\tau$
store
${\upsilon _1}, {\upsilon _2}, \cdots,  {\upsilon _\tau}$
columns of ${{\bf{R}}_k}$, respectively, and the columns stored in workers $1,2,\cdots,\tau$ are
columns $1:{\tilde \upsilon}_1,({\tilde \upsilon}_1+1):{\tilde \upsilon}_2,\cdots,({\tilde \upsilon}_{\tau-1}+1):{\tilde \upsilon}_\tau$ that satisfy
(\ref{ColumnPointAssume3s3}).
Accordingly, the parallel implementation  described in \textbf{Algorithm 13}
can be applied to compute  the inverse $LDL^T$ factors of the upper-triangular ${{\bf{R}}_k}$
for the distributed BLS with model-parallelism.

\section{The Division Free Inverse LDL$^T$ Factorization and the Corresponding Parallelization}

There is still a division operation in (\ref{InvLDLSaveMem3290ccc}). To avoid that division operation,
we can use the  alternative
LDL$^T$ factors of ${{\mathbf{R}}_{i}^{-1}}$, 
i.e., ${\mathbf{\tilde{L}}}_{i}$,  ${\mathbf{\tilde{D}}}_{i}$ and ${\sigma}_{i}$ satisfying
\begin{equation}\label{LDLinvFreeNov17c}
	{\mathbf{\tilde{L}}}_{i}   ({\mathbf{\tilde{D}}}_{i}   /{\sigma}_{i}    ){{\mathbf{\tilde{L}}}_{i}^{T}}   ={\mathbf{L}}_i {\mathbf{D}}_i{{\mathbf{L}}_i^T} ={{\mathbf{R}}_i^{-1}}.
\end{equation}
The corresponding division free block inverse LDL$^T$ factorization and its parallelization
will be introduced in this section.

\subsection{Division Free Block Inverse LDL$^T$ Factorization}

The division free block inverse LDL$^T$ factorization
update ${\mathbf{\tilde{L}}}_{i}$,  ${\mathbf{\tilde{D}}}_{i}$ and ${\sigma}_{i}$  into
${\mathbf{\tilde{L}}}_{k}$,  ${\mathbf{\tilde{D}}}_{k}$ and ${\sigma}_{k}$ (with $k>i$) by
\begin{subnumcases}{\label{GeneralEquGroup131}}
	\mathbf{\tilde{\Delta}  }(\mathbf{\tilde{\Phi}}/\eta ){{\mathbf{\tilde{\Delta}  }}^{T}}={{\left( {\sigma}_{i}    \mathbf{U}-{{\mathbf{P}}^{T}}{\mathbf{\tilde{L}}_i}{\mathbf{\tilde{D}}_i}{{\mathbf{\tilde{L}}}_{i}^{T}}   \mathbf{P} \right)}^{-1}}            &  \label{LDLinvFreeNov17e}\\
	{\rm{Scaling}} \quad  {\rm{of}}   \quad \eta \quad {\rm{and}} \quad  {{\bf{\tilde \Delta }}}   &  \label{scalingBlock329sd32}  \\
	{{\mathbf{\tilde{L}}}_{k}}=\left[ \begin{matrix}
		{{{\mathbf{\tilde{L}}}}_{i}} & -{{{\mathbf{\tilde{L}}}}_{i}}{{{\mathbf{\tilde{D}}}}_{i}}\mathbf{\tilde{L}}_{i}^{T}\mathbf{P\tilde{\Delta}  }  \\
		\mathbf{0} & {\sigma}_i \mathbf{\tilde{\Delta}  }  \\
	\end{matrix} \right] &  \label{GeneralEquGroup131LLL}\\
	{{\mathbf{\tilde{D}}}_{k}}=\left[ \begin{matrix}
		\eta {{{\mathbf{\tilde{D}}}}_{i}} & \mathbf{0}  \\
		\mathbf{0} & {\mathbf{\tilde{\Phi}}}  \\
	\end{matrix} \right]   &  \label{etaDcompute39212}\\
	{{\sigma }_{k}}={{\sigma }_{i}}\eta, &  \label{GeneralEquGroup131Sigma}
\end{subnumcases}
where  $\mathbf{\tilde{\Delta}}$,  $\mathbf{\tilde{\Phi}}$
and $\eta$, which satisfy (\ref{LDLinvFreeNov17e}),
are also the alternative LDL$^T$ factors defined by (\ref{LDLinvFreeNov17c}).
The above division free block inverse LDL$^T$ factorization by (\ref{GeneralEquGroup131}) can be regarded as a special case of the division free block inverse LDM$^T$ factorization introduced in \cite{BLASTLULDLpresented32ds}.
We will give the derivation of (\ref{GeneralEquGroup131}) in
what follows.

Firstly,  substitute
(\ref{LDLinvFreeNov17c})   into (\ref{LDLinvFreeNov17b})   to obtain
\begin{gather}
	\mathbf{\Delta  \Phi }{{\mathbf{\Delta}  }^{T}}={{\left( \mathbf{U}-{{\mathbf{P}}^{T}}{\mathbf{\tilde{L}}}_{i}   ({\mathbf{\tilde{D}}}_{i}   /{\sigma}_{i}    ){{\mathbf{\tilde{L}}}_{i}^{T}}   \mathbf{P} \right)}^{-1}}  \notag \\
	\mathbf{\Delta}  (\mathbf{\Phi}/{\sigma}_{i}    ){{\mathbf{\Delta}  }^{T}}={{\left( {\sigma}_{i}    \mathbf{U}-{{\mathbf{P}}^{T}}{\mathbf{\tilde{L}}_i}{\mathbf{\tilde{D}}_i}{{\mathbf{\tilde{L}}}_{i}^{T}}   \mathbf{P} \right)}^{-1}}.  \label{LDLinvFreeNov17d}
\end{gather}
For the right side of (\ref{LDLinvFreeNov17d}),
we can write the alternative LDL$^T$ factors defined by (\ref{LDLinvFreeNov17c}),
i.e.,
$\mathbf{\tilde{\Delta}}$,  $\mathbf{\tilde{\Phi}}$
and $\eta$  satisfying (\ref{LDLinvFreeNov17e}).																																																																																																																																			
Then
(\ref{LDLinvFreeNov17e}) is substituted into
(\ref{LDLinvFreeNov17d})
to obtain
\begin{equation}\label{LDLinvFreeNov17f}
	\mathbf{\tilde{\Delta}  }({\sigma}_{i}    \mathbf{\tilde{\Phi}}/\eta ){{\mathbf{\tilde{\Delta}  }}^{T}}=\mathbf{\Delta  \Phi }{{\mathbf{\Delta}  }^{T}}.
\end{equation}

Now we only need to deduce (\ref{GeneralEquGroup131LLL}),
(\ref{etaDcompute39212}) and (\ref{GeneralEquGroup131Sigma}).
Substitute (\ref{ABFGEcomputeE84r9834})
into
(\ref{LDMdefineAll313943}),
and substitute  (\ref{LDMdefineAll313943})
into (\ref{LDMtFactorsDefine398da32})  to obtain
\begin{footnotesize}
	\begin{align}
		&\mathbf{R}_{k}^{-1}  \notag \\
		&=\left[ \begin{matrix}
			{{\mathbf{L}}_{i}} & -\mathbf{L}_{i}^{{}}\mathbf{D}_{i}^{{}}\mathbf{L}_{i}^{T}\mathbf{P \Delta}  \\
			\mathbf{0} & \mathbf{\Delta}    \\
		\end{matrix} \right]  \left[ \begin{matrix}
			{{\mathbf{D}}_{i}} & \mathbf{0}  \\
			\mathbf{0} & \mathbf{\Phi}  \\
		\end{matrix} \right]\left[ \begin{matrix}
			{{\mathbf{L}}_{i}^T} &  \mathbf{0}  \\
			-{{\mathbf{\Delta}  }^{T}}{{\mathbf{P}}^{T}}\mathbf{L}_{i}^{{}}\mathbf{D}_{i}^{{}}\mathbf{L}_{i}^{T}   & \mathbf{\Delta}  ^T  \\
		\end{matrix} \right]  \notag \\
		&=\left[ {\begin{array}{*{20}{c}}
				{\left( \begin{array}{l}
						{{\bf{L}}_i}{{\bf{D}}_i}{\bf{L}}_i^T + {\bf{L}}_i^{}{\bf{D}}_i^{}{\bf{L}}_i^T{\bf{P}}\\
						\times {\bf{\Delta \Phi }}{{\bf{\Delta }}^T}{{\bf{P}}^T}{\bf{L}}_i^{}{\bf{D}}_i^{}{\bf{L}}_i^T
					\end{array} \right)}&{\left( \begin{array}{l}
						- {\bf{L}}_i^{}{\bf{D}}_i^{}{\bf{L}}_i^T\\
						\times {\bf{P\Delta \Phi }}{{\bf{\Delta }}^T}
					\end{array} \right)}\\
				{ - {\bf{\Delta \Phi }}{{\bf{\Delta }}^T}{{\bf{P}}^T}{\bf{L}}_i^{}{\bf{D}}_i^{}{\bf{L}}_i^T}&{{\bf{\Delta \Phi }}{{\bf{\Delta }}^T}}
		\end{array}} \right],  \notag
	\end{align}
\end{footnotesize}
into which substitute (\ref{LDLinvFreeNov17c})  and (\ref{LDLinvFreeNov17f})  to obtain
\begin{footnotesize}
	\begin{align}
		&\mathbf{R}_{k}^{-1}  \notag \\
		&=\left[ {\begin{array}{*{20}{c}}
				{\left( \begin{array}{l}
						\frac{{{{{\bf{\tilde L}}}_i}{{{\bf{\tilde D}}}_i}{\bf{\tilde L}}_i^T}}{{{\sigma _i}}} + \frac{{{{{\bf{\tilde L}}}_i}{{{\bf{\tilde D}}}_i}{\bf{\tilde L}}_i^T}}{{{\sigma _i}}}{\bf{P}} \times \\
						\frac{{{\bf{\tilde \Delta }}{\sigma _i}{\bf{\tilde \Phi }}{{{\bf{\tilde \Delta }}}^T}}}{\eta }{{\bf{P}}^T}\frac{{{{{\bf{\tilde L}}}_i}{{{\bf{\tilde D}}}_i}{\bf{\tilde L}}_i^T}}{{{\sigma _i}}}
					\end{array} \right)}&{\left( \begin{array}{l}
						- \frac{{{{{\bf{\tilde L}}}_i}{{{\bf{\tilde D}}}_i}{\bf{\tilde L}}_i^T}}{{{\sigma _i}}} \times \\
						{\bf{P}}\frac{{{\bf{\tilde \Delta }}{\sigma _i}{\bf{\tilde \Phi }}{{{\bf{\tilde \Delta }}}^T}}}{\eta }
					\end{array} \right)}\\
				{ - \frac{{{\bf{\tilde \Delta }}{\sigma _i}{\bf{\tilde \Phi }}{{{\bf{\tilde \Delta }}}^T}}}{\eta }{{\bf{P}}^T}\frac{{{{{\bf{\tilde L}}}_i}{{{\bf{\tilde D}}}_i}{\bf{\tilde L}}_i^T}}{{{\sigma _i}}}}&{\frac{{{\bf{\tilde \Delta }}{\sigma _i}{\bf{\tilde \Phi }}{{{\bf{\tilde \Delta }}}^T}}}{\eta }}
		\end{array}} \right]  \notag \\
		&=\left[ {\begin{array}{*{20}{c}}
				{\left( \begin{array}{l}
						\frac{{{{{\bf{\tilde L}}}_i}{{{\bf{\tilde D}}}_i}{\bf{\tilde L}}_i^T}}{{{\sigma _i}}} + \frac{{{{{\bf{\tilde L}}}_i}{{{\bf{\tilde D}}}_i}{\bf{\tilde L}}_i^T}}{{{\sigma _i}}} \times \\
						{\bf{P}}\frac{{{\bf{\tilde \Delta \tilde \Phi }}{{{\bf{\tilde \Delta }}}^T}}}{\eta }{{\bf{P}}^T}{{{\bf{\tilde L}}}_i}{{{\bf{\tilde D}}}_i}{\bf{\tilde L}}_i^T
					\end{array} \right)}&{\left( \begin{array}{l}
						- {{{\bf{\tilde L}}}_i}{{{\bf{\tilde D}}}_i}{\bf{\tilde L}}_i^T\\
						\times {\bf{P}}\frac{{{\bf{\tilde \Delta \tilde \Phi }}{{{\bf{\tilde \Delta }}}^T}}}{\eta }
					\end{array} \right)}\\
				{ - \frac{{{\bf{\tilde \Delta \tilde \Phi }}{{{\bf{\tilde \Delta }}}^T}}}{\eta }{{\bf{P}}^T}{{{\bf{\tilde L}}}_i}{{{\bf{\tilde D}}}_i}{\bf{\tilde L}}_i^T}&{\frac{{{\bf{\tilde \Delta }}{\sigma _i}{\bf{\tilde \Phi }}{{{\bf{\tilde \Delta }}}^T}}}{\eta }}
		\end{array}} \right],  \label{LDLinvFree4scaling} \\
		&=\left[ {\begin{array}{*{20}{c}}
				{{{{\bf{\tilde L}}}_i}}&{ - {{{\bf{\tilde L}}}_i}{{{\bf{\tilde D}}}_i}{\bf{\tilde L}}_i^T{\bf{P\tilde \Delta }}}\\
				{\bf{0}}&{{\sigma _i}{\bf{\tilde \Delta }}}
		\end{array}} \right]\frac{1}{{{\sigma _i}\eta }}\left[ {\begin{array}{*{20}{c}}
				{\eta {{{\bf{\tilde D}}}_i}}&{\bf{0}}\\
				{\bf{0}}&{{\bf{\tilde \Phi }}}
		\end{array}} \right]  \notag \\
		&\quad \quad \quad \quad \quad \quad   \times  \left[ {\begin{array}{*{20}{c}}
				{{\bf{\tilde L}}_i^T}&{\bf{0}}\\
				{ - {{\bf{P}}^T}{{{\bf{\tilde L}}}_i}{{{\bf{\tilde D}}}_i}{\bf{\tilde L}}_i^T}&{{\sigma _i}{{{\bf{\tilde \Delta }}}^T}}
		\end{array}} \right].   \label{LDLinvFree3290sd23sd}
	\end{align}
\end{footnotesize}
From (\ref{LDLinvFree3290sd23sd}),  we can deduce  (\ref{GeneralEquGroup131LLL}),
(\ref{etaDcompute39212}) and (\ref{GeneralEquGroup131Sigma}) finally, which
compute ${\mathbf{\tilde{L}}}_{k}$,   ${\mathbf{\tilde{D}}}_{k}$  and  ${\sigma}_{k}$   satisfying
(\ref{LDLinvFreeNov17c}).

The iterations in (\ref{GeneralEquGroup131}) will lead to numerically unlimited
results, which may cause a problem in fixed-point implementations
\cite{WCNCcholesky}. We alleviate this problem by scaling in (\ref{scalingBlock329sd32}), as in
\cite{WCNCcholesky}. Scaling is achieved by dividing (or
multiplying) only by powers of $2$ \cite{WCNCcholesky}, which is a
shift operation in binary fixed-point implementation.
We can multiply   ${\eta}$
by $c_i^2$ and multiply
${{\bf{\tilde \Delta }}}$ by $c_i$ accordingly, which will not change $\mathbf{R}_{k}^{-1}$ computed by (\ref{LDLinvFree4scaling}).   Then  we
end up with ${\eta} c_i^2$ and ${{\bf{\tilde \Delta }}} c_i$,
where ${\eta} c_i^2$ is always between $0.5$ and $2$,
and $c_i$ is powers of $2$.

In the special case with $k-i=1$,
${{\mathbf{P}}}$ and ${{\mathbf{U}}}$ in
(\ref{GeneralEquGroup131})
becomes the column vector ${{\mathbf{p}}}$ and the scalar $u$,
respectively,
and
(\ref{GeneralEquGroup131}) with $i=i-1$ becomes~\footnote{The following equation (\ref{GeneralEquGroup131toi1}) has been presented in \cite{zhfVTC2010DivFree}, while the scaling scheme presented in
	\cite{zhfVTC2010DivFree} is different from the scaling scheme proposed in this paper.}

\begin{subnumcases}{\label{GeneralEquGroup131toi1}}
	\eta ={{\sigma }_{i-1}}u-\mathbf{p}^{T}{\mathbf{\tilde{L}}_{i-1}}{\mathbf{\tilde{D}}_{i-1}}{{\mathbf{\tilde{L}}}_{i-1}^{T}}{{\mathbf{p}}}      &  \label{LDLinvFreeNov17iAddVecGroup}\\
	{{\bf{\tilde l}}_i^{}}= -{{{\mathbf{\tilde{L}}}}_{i-1}}{{{\mathbf{\tilde{D}}}}_{i-1}}\mathbf{\tilde{L}}_{i-1}^{T}\mathbf{p}      &  \label{GeneralEquGroup131toi1LLL}\\
	{{l_{ii}}}={{\sigma }_{i-1}}       &  \label{lii2sigmaiMinus1ada23}\\
	{\rm{Scaling}} \quad  {\rm{of}}   \quad \eta, \quad {{\bf{\tilde l}}_i^{}} \quad {\rm{and}} \quad  {{l_{ii}}}   &  \label{scaling2i2one302ds23ds}  \\
	{{{\bf{\tilde L}}}_i} = \left[ {\begin{array}{*{20}{c}}
			{{{{\bf{\tilde L}}}_{i - 1}}}&  {{\bf{\tilde l}}_i^{}}\\
			{\bf{0}}&{{l_{ii}}}
	\end{array}} \right]  &  \label{divFreeLaddCol3290sd}\\
	{{\mathbf{\tilde{D}}}_{i}}=\left[ \begin{matrix}
		\eta {{{\mathbf{\tilde{D}}}}_{i-1}} & \mathbf{0}  \\
		\mathbf{0} & 1  \\
	\end{matrix} \right]   &  \label{GeneralEquGroup131toi1DDD}\\
	{{\sigma }_{i}}={{\sigma }_{i-1}}\eta.  &  \label{GeneralEquGroup131toi1sigma}
\end{subnumcases}
Notice that
$\mathbf{\tilde{\Delta}  }=\mathbf{\tilde{\Phi}}=1$
and $\eta$ computed by
(\ref{LDLinvFreeNov17iAddVecGroup}), which
satisfy  (\ref{LDLinvFreeNov17e}),
are substituted into
(\ref{GeneralEquGroup131LLL}) and
(\ref{etaDcompute39212})
to obtain
(\ref{GeneralEquGroup131toi1LLL}) and
(\ref{GeneralEquGroup131toi1DDD}),
respectively.
Moreover,  it can be seen from  (\ref{LDLinvFreeNov17c}) with  $i=1$ that we can let
the initial
\begin{subnumcases}{\label{S16NoDIVkOne}}
	{\bf{{\tilde{L}}}}_{1}=1, & \label{S16NoDIVkOne1}\\
	{\bf{{\tilde{D}}}}_{1}=1, & \label{S16NoDIVkOneD}\\
	{\sigma _{1}}={\bf{R}}_{1}=\mathbf{R}(1,1), & \label{S16NoDIVkOne2}\\
	{\rm{Scaling}} \quad  {\rm{of}}   \quad {\sigma _{1}} \quad {\rm{and}} \quad  {\bf{{\tilde{L}}}}_{1}.   &  \label{scalingOfL1Delta1adwesdwe}
\end{subnumcases}
From the initial  ${{\mathbf{\tilde{L}}}_{1}}$,   ${{\mathbf{\tilde{D}}}_{1}}$ and ${{\sigma }_{1}}$
computed by (\ref{S16NoDIVkOne}),
we can use (\ref{GeneralEquGroup131toi1})
to compute
${{\mathbf{\tilde{L}}}_{i}}$,   ${{\mathbf{\tilde{D}}}_{i}}$ and ${{\sigma }_{i}}$ from ${{\mathbf{\tilde{L}}}_{i-1}}$,   ${{\mathbf{\tilde{D}}}_{i-1}}$ and ${{\sigma }_{i-1}}$ iteratively,
till we obtain
${{\mathbf{\tilde{L}}}_{k}}$,   ${{\mathbf{\tilde{D}}}_{k}}$ and ${{\sigma }_{k}}$ finally.


The iterations in (\ref{GeneralEquGroup131toi1}) will lead to numerically unlimited
results, and then we alleviate this problem by scaling in (\ref{scaling2i2one302ds23ds}) and the initial  (\ref{scalingOfL1Delta1adwesdwe}).
To implement the scaling in (\ref{scalingOfL1Delta1adwesdwe}), we can multiply   ${\sigma _{1}}$
by $c_1^2$ and multiply
${\bf{{\tilde{L}}}}_{1}$ by $c_1$ accordingly,
which will not change ${{\mathbf{R}}_1^{-1}}$  computed by  (\ref{LDLinvFreeNov17c}) with  $i=1$.
Then we
end up with ${\sigma _{1}} c_1^2$   and ${\bf{{\tilde{L}}}}_{1} c_1$,
where ${\sigma _{1}} c_1^2$ is always between $0.5$ and $2$,
and $c_1$ is powers of $2$.

Moreover,   substitute
(\ref{divFreeLaddCol3290sd}),
(\ref{GeneralEquGroup131toi1DDD})
and
(\ref{GeneralEquGroup131toi1sigma})
into
(\ref{LDLinvFreeNov17c})
to get

\begin{gather}
	{{\mathbf{R}}_i^{-1}}= \left[ {\begin{array}{*{20}{c}}
			{{{{\bf{\tilde L}}}_{i - 1}}}&{{\bf{\tilde l}}_i^{}}\\
			{\bf{0}}&{{l_{ii}}}
	\end{array}} \right]
	\frac{\left[ {\begin{array}{*{20}{c}}
				{\eta {{{\bf{\tilde D}}}_{i - 1}}}&{\bf{0}}\\
				{\bf{0}}&1
		\end{array}} \right]}{{{\sigma _{i - 1}}\eta }}
	{\left[ {\begin{array}{*{20}{c}}
				{{{{\bf{\tilde L}}}_{i - 1}}}&{{\bf{\tilde l}}_i^{}}\\
				{\bf{0}}&{{l_{ii}}}
		\end{array}} \right]^T}  \notag \\
	= \frac{1}{{{\sigma _{i - 1}}}}\left[ {\begin{array}{*{20}{c}}
			{{{{\bf{\tilde L}}}_{i - 1}}{{{\bf{\tilde D}}}_{i - 1}}{\bf{\tilde L}}_{i - 1}^T + \frac{1}{\eta }{\bf{\tilde l}}_i^{}{\bf{\tilde l}}_i^T}&{\frac{1}{\eta }{l_{ii}}{\bf{\tilde l}}_i^{}}\\
			{\frac{1}{\eta }{l_{ii}}{\bf{\tilde l}}_i^T}&{\frac{1}{\eta }l_{ii}^2}
	\end{array}} \right]. \label{deduceScaling239ds32}
\end{gather}
It can be seen that ${{\mathbf{R}}_i^{-1}}$ computed by
(\ref{deduceScaling239ds32})  will not change, if
we implement the scaling in (\ref{scaling2i2one302ds23ds}) by multiplying   ${\eta}$
with $c_i^2$ and multiply
${\bf{\tilde l}}_i$ and $l_{ii}$  with $c_i$ accordingly. Then we
end up with ${\eta} c_i^2$,  ${\bf{\tilde l}}_i c_i$  and $l_{ii} c_i$,
where ${\eta} c_i^2$ is always between $0.5$ and $2$,
and $c_i$ is powers of $2$.

\subsection{The Memory-Saving Parallel Implementation of the Division Free Inverse LDL$^T$ Factorization}

Let us define $-{{{\mathbf{\tilde{L}}}}_{k}}{{{\mathbf{\tilde{D}}}}_{k}}\mathbf{\tilde{L}}_{k}^{T}\mathbf{P}$ in (\ref{GeneralEquGroup131LLL}) and  ${\sigma}_{k}    \mathbf{U}-{{\mathbf{P}}^{T}}{\mathbf{\tilde{L}}_k}{\mathbf{\tilde{D}}_k}{{\mathbf{\tilde{L}}}_{k}^{T}}   \mathbf{P}$ in (\ref{LDLinvFreeNov17e})
as
\begin{subnumcases}{\label{PiXi239sdk23sdboth}}
	{{\mathbf{\tilde \Pi }}_{i}}= -{{{\mathbf{\tilde{L}}}}_{i}}{{{\mathbf{\tilde{D}}}}_{i}}\mathbf{\tilde{L}}_{i}^{T}\mathbf{P} &  \label{pi2FFAA2390923DivFree}\\
	{{\mathbf{\tilde \Xi }}_{k-i}}={\sigma}_{i}    \mathbf{U}-{{\mathbf{P}}^{T}}{\mathbf{\tilde{L}}_i}{\mathbf{\tilde{D}}_i}{{\mathbf{\tilde{L}}}_{i}^{T}}.  &  \label{Xi2AAFFAA23094d3DivFree}
\end{subnumcases}
The above ${{\mathbf{\tilde \Pi }}_{i}}$ and ${{\mathbf{\tilde \Xi }}_{k-i}}$ satisfy
\begin{subnumcases}{\label{PiXi430990dfdfde3}}
	{{\bf{\Pi }}_i} = {{\bf{\tilde \Pi }}_i}/{\sigma _i} &  \label{pi2FFAA2390923DivFree222}\\
	{{\bf{\Xi }}_{k - i}} = {{\bf{\tilde \Xi }}_{k - i}}/{\sigma _i}.  &  \label{Xi2AAFFAA23094d3DivFree222}
\end{subnumcases}
To deduce (\ref{pi2FFAA2390923DivFree222}) and (\ref{Xi2AAFFAA23094d3DivFree222}),  substitute
(\ref{LDLinvFreeNov17c})
into
(\ref{L_m_def12431RidgeInvJun30JunY24})
to obtain
\begin{equation}\label{FF2LDL032dasasr3ed3DivFree}
	{\mathbf{\tilde{L}}}_{i}   {\mathbf{\tilde{D}}}_{i}   {{\mathbf{\tilde{L}}}_{i}^{T}}={\sigma}_{i} {{\bf{F}}_i}{{\bf{F}}_i^{T}},
\end{equation}
which is substituted into (\ref{PiXi239sdk23sdboth})
to obtain
\begin{subnumcases}{\label{}}
	{{\mathbf{\tilde \Pi }}_{i}}= -{\sigma}_{i} {{\bf{F}}_i}{{\bf{F}}_i^{T}}\mathbf{P} &  \label{pi2FFAA2390923DivFree111}\\
	{{\mathbf{\tilde \Xi }}_{k-i}}={\sigma}_{i}    \mathbf{U}-{{\mathbf{P}}^{T}}{\sigma}_{i}  {{\bf{F}}_i}{{\bf{F}}_i^{T}}   \mathbf{P}.  &  \label{Xi2AAFFAA23094d3DivFree111}
\end{subnumcases}
Then
we  substitute  (\ref{UP2AA23ds23BothPP})
into  (\ref{pi2FFAA2390923}) to write
${{\mathbf{\Pi }}_{i}}=-{{\mathbf{F}}_{i}}\mathbf{F}_{i}^{T}\mathbf{P}$,
which is substituted
into (\ref{pi2FFAA2390923DivFree111})
to obtain (\ref{pi2FFAA2390923DivFree222}).

We also substitute (\ref{UP2AA23ds23BothPP}) and
(\ref{UP2AA23ds23BothUU}) into (\ref{Xi2AAFFAA23094d3}) to write
${{\mathbf{\Xi }}_{k-i}}= \mathbf{U} - \mathbf{P}^{T}{{\mathbf{F}}_{i}}\mathbf{F}_{i}^{T}\mathbf{P}$,
which is substituted into
(\ref{Xi2AAFFAA23094d3DivFree111}) to obtain (\ref{Xi2AAFFAA23094d3DivFree222}).
Let us substitute
(\ref{S16NoDIVkOne2}) into
(\ref{pi2FFAA2390923DivFree222})
and
(\ref{Xi2AAFFAA23094d3DivFree222})
with $i=1$ to obtain
${{\mathbf{\tilde \Pi }}_{1}}= \mathbf{R}(1,1){{\mathbf{\Pi }}_{1}}$
and
${{\mathbf{\tilde \Xi }}_{k-1}}= \mathbf{R}(1,1){{\mathbf{\Xi }}_{k-1}}$,
respectively,
into which  substitute
(\ref{InitialPi2FFRXiR932ds23}) to obtain  
\begin{footnotesize}
	\begin{subnumcases}{\label{InitialPi2FFRXiR932ds23DivFree}}
		{{\mathbf{\tilde \Pi }}_{1}} = - \mathbf{R}(1,2:k) &    \label{InitialPi2FFRXiR932ds23aaaDivFree}\\
		{{\mathbf{\tilde \Xi }}_{k-1}} = \mathbf{R}(1,1) \mathbf{R}(2:k,2:k)-\mathbf{R}(1,2:k)_{{}}^{T}\mathbf{R}(1,2:k).   &  \label{InitialPi2FFRXiR932ds23bbbDivFree}
	\end{subnumcases}
\end{footnotesize}%
In order to save memories,  we store ${{\bf{\tilde \Xi }}_{k - 1}}$,
${{\mathbf{\tilde \Pi}}_{1}}$    and ${\bf{{\tilde{L}}}}_{1}$ in ${\bf{R}}(2:k,2:k)$,
$\mathbf{R}(1,2:k)$  and ${\bf{R}}(1,1)$, respectively, by computing
\begin{footnotesize}
	\begin{subnumcases}{\label{iTo1InvLDLSaveMemDivFree}}
		\sigma  = {\bf{R}}(1,1)  &  \label{iTo1InvLDLSaveMemDivFree1}\\
		{\bf{R}}(1,1) = 1    &  \label{iTo1InvLDLSaveMemDivFree2}\\
		{\rm{Scaling}} \quad  {\rm{of}}   \quad  \sigma  \quad {\rm{and}} \quad {\bf{R}}(1,:)    & \label{scalingRmatrix329sd32}  \\
		{\bf{R}}(2:k,2:k) =\sigma {\bf{R}}(2:k,2:k)-{\bf{R}}{(1,2:k)^T}{\bf{R}}(1,2:k)   &  \label{iTo1InvLDLSaveMemDivFree5}  \\
		{\bf{R}}(1,2:k) =  - {\bf{R}}(1,1){\bf{R}}(1,2:k)   &  \label{iTo1InvLDLSaveMemDivFree6}  \\
		{\bf{D}}(1,1) = 1,   &  \label{iTo1InvLDLSaveMemDivFree6d11}
	\end{subnumcases}
\end{footnotesize}
where
(\ref{iTo1InvLDLSaveMemDivFree1}), (\ref{iTo1InvLDLSaveMemDivFree2}),    (\ref{iTo1InvLDLSaveMemDivFree5}),
(\ref{iTo1InvLDLSaveMemDivFree6})
and (\ref{iTo1InvLDLSaveMemDivFree6d11})
are deduced from  (\ref{S16NoDIVkOne2}),  (\ref{S16NoDIVkOne1}),   (\ref{InitialPi2FFRXiR932ds23bbbDivFree}),
(\ref{InitialPi2FFRXiR932ds23aaaDivFree})
and (\ref{S16NoDIVkOneD}),
respectively.  In  (\ref{scalingRmatrix329sd32}),  the scaling of $\sigma$ and ${\bf{R}}(1,:)$
is achieved by multiplying $\sigma$ and ${\bf{R}}(1,:)$
by $c_k^2$ and $c_k$, respectively, to
end up with $\sigma c_k^2$ and ${\bf{R}}(1,:) c_k$,
where $\sigma c_k^2$ is always between $0.5$ and $2$,
and $c_k$ is powers of $2$.

In the $i^{th}$ ($i \ge 2$) iteration, ${{\bf{\tilde \Xi }}_{k - i}}$ and ${{\bf{\tilde \Pi }}_i}$
can be computed efficiently by
\begin{multline}\label{Xi2Xi2vec342ds43}
	{{\bf{\tilde \Xi }}_{k - i}} = \\
	\eta {\bf{\tilde \Xi }}_{k - i + 1}^{[ - 1, - 1]} - {{\bf{\tilde \Xi }}_{k - i + 1}}{(1,2:end)^T}{{\bf{\tilde \Xi }}_{k - i + 1}}(1,2:end)
\end{multline}
and
\begin{small}
	\begin{multline}\label{Pi2PiSigma2390kds3ds}
		{{\bf{\tilde \Pi }}_i} = \\
		\left[ {\begin{array}{*{20}{c}}
				{\eta {{{\bf{\tilde \Pi }}}_{i - 1}}(:,2:end) - {{{\bf{\tilde \Pi }}}_{i - 1}}(:,1){{{\bf{\tilde \Xi }}}_{k - i + 1}}(1,2:end)}\\
				{ - {\sigma _{i - 1}}{{{\bf{\tilde \Xi }}}_{k - i + 1}}(1,2:end)}
		\end{array}} \right],
	\end{multline}
\end{small}
respectively.
To deduce
(\ref{Xi2Xi2vec342ds43}) and (\ref{Pi2PiSigma2390kds3ds}),
substitute  (\ref{GeneralEquGroup131toi1sigma}),
(\ref{divFreeLaddCol3290sd}),
(\ref{GeneralEquGroup131toi1DDD}),
(\ref{LLDDsmaller932sd23aaa})  and
(\ref{LLDDsmaller932sd23bbb}),
(i.e., ${{\sigma }_{i}}={{\sigma }_{i-1}}\eta$, the partitioned
${{{\bf{\tilde L}}}_i}$,
${{\mathbf{\tilde{D}}}_{i}}$, ${{\bf{L}}_i}$ and
${{\bf{D}}_i}$)
into
(\ref{LDLinvFreeNov17c}) (i.e., ${\mathbf{\tilde{L}}}_{i}   ({\mathbf{\tilde{D}}}_{i}   /{\sigma}_{i}    ){{\mathbf{\tilde{L}}}_{i}^{T}}   ={\mathbf{L}}_i {\mathbf{D}}_i{{\mathbf{L}}_i^T}$)
to obtain $d_i^{} = {{\bf{\tilde l}}_i^{2}}/(\eta  {\sigma _{i - 1}})$, into which substitute (\ref{lii2sigmaiMinus1ada23}) (i.e., ${{l_{ii}}}={{\sigma }_{i-1}}$) to obtain
\begin{equation}\label{di2sigmaeta32sd23d}
	d_i^{} = {\sigma _{i - 1}}/\eta.
\end{equation}
Then substitute
(\ref{dii2Xi390ads2d32})
and
(\ref{Xi2AAFFAA23094d3DivFree222})
into
(\ref{Xi2k409043fd})
to obtain
$\frac{{{{{\bf{\tilde \Xi }}}_{k - i}}}}{{{\sigma _i}}} = \frac{{{\bf{\tilde \Xi }}_{k - i + 1}^{[ - 1, - 1]}}}{{{\sigma _{i - 1}}}} - d_i^{}\frac{{{{{\bf{\tilde \Xi }}}_{k - i + 1}}{{(1,2:end)}^T}}}{{{\sigma _{i - 1}}}}\frac{{{{{\bf{\tilde \Xi }}}_{k - i + 1}}(1,2:end)}}{{{\sigma _{i - 1}}}}$,

into which substitute (\ref{GeneralEquGroup131toi1sigma}) and (\ref{di2sigmaeta32sd23d}) to obtain

$\frac{{{{{\bf{\tilde \Xi }}}_{k - i}}}}{{{\sigma _{i - 1}}\eta }} = \frac{{{\bf{\tilde \Xi }}_{k - i + 1}^{[ - 1, - 1]}}}{{{\sigma _{i - 1}}}} - \frac{\sigma _{i - 1}}{\eta}\frac{{{{{\bf{\tilde \Xi }}}_{k - i + 1}}{{(1,2:end)}^T}}}{{{\sigma _{i - 1}}}}\frac{{{{{\bf{\tilde \Xi }}}_{k - i + 1}}(1,2:end)}}{{{\sigma _{i - 1}}}}$, i.e.,  (\ref{Xi2Xi2vec342ds43}).

On the other hand, let us
substitute (\ref{dii2Xi390ads2d32}) and (\ref{PiXi430990dfdfde3})  into
(\ref{Pi2k9430934ds3})
to obtain
$\frac{{{{{\bf{\tilde \Pi }}}_i}}}{{{\sigma _i}}} = \left[ {\begin{array}{*{20}{c}}
		{\frac{{{{{\bf{\tilde \Pi }}}_{i - 1}}(:,2:end)}}{{{\sigma _{i - 1}}}} - \frac{{{{{\bf{\tilde \Pi }}}_{i - 1}}(:,1)}}{{{\sigma _{i - 1}}}}d_i^{}\frac{{{{{\bf{\tilde \Xi }}}_{k - i + 1}}(1,2:end)}}{{{\sigma _{i - 1}}}}}\\
		{ - d_i^{}\frac{{{{{\bf{\tilde \Xi }}}_{k - i + 1}}(1,2:end)}}{{{\sigma _{i - 1}}}}}
\end{array}} \right]$,
into which substitute (\ref{GeneralEquGroup131toi1sigma})  and (\ref{di2sigmaeta32sd23d})  to obtain

$\frac{{{{{\bf{\tilde \Pi }}}_i}}}{{{\sigma _{i - 1}}\eta }} = \left[ {\begin{array}{*{20}{c}}
		{\frac{{{{{\bf{\tilde \Pi }}}_{i - 1}}(:,2:end)}}{{{\sigma _{i - 1}}}} - \frac{{{{{\bf{\tilde \Pi }}}_{i - 1}}(:,1)}}{{{\sigma _{i - 1}}}} \frac{\sigma _{i - 1}}{\eta} \frac{{{{{\bf{\tilde \Xi }}}_{k - i + 1}}(1,2:end)}}{{{\sigma _{i - 1}}}}}\\
		{ -  \frac{\sigma _{i - 1}}{\eta}\frac{{{{{\bf{\tilde \Xi }}}_{k - i + 1}}(1,2:end)}}{{{\sigma _{i - 1}}}}}
\end{array}} \right]$,  i.e., (\ref{Pi2PiSigma2390kds3ds}).

To save memories in the $i^{th}$ ($i \ge 2$) iteration,
we also  store
${{\mathbf{\Xi }}_{k-i}}$ and  ${{\mathbf{\Pi }}_{i}}$
in  ${\bf{R}}(i + 1:k,i + 1:k)$ and
${\bf{R}}(1:i,i + 1:k)$, respectively, by computing
\begin{footnotesize}
	\begin{subnumcases}{\label{InvLDLSaveMemDivFree}}
		\eta  = {\bf{R}}(i,i)  &  \label{InvLDLSaveMemDivFree1}\\
		{\bf{R}}(i,i) = \sigma    &  \label{InvLDLSaveMemDivFree2}\\
		{\rm{Scaling}} \quad  {\rm{of}}   \quad \eta, \quad {\bf{R}}(1:i,i) \quad {\rm{and}} \quad  {\bf{R}}(i,i+1:k)   & \label{scalingRmatrixGENERALi}   \\
		{\bf{\tilde D}}(1:i - 1,1:i - 1) = \eta {\bf{\tilde D}}(1:i - 1,1:i - 1)    &  \label{InvLDLSaveMemDivFree3}\\
		\sigma  = \sigma  \times \eta     &  \label{InvLDLSaveMemDivFree4}\\
		{\begin{split}
				{\bf{R}}(i + 1:k,i + 1:k) =\eta {\bf{R}}(i + 1:k,i + 1:k)  \quad  \quad \quad \quad      \\
				\quad \quad \quad \quad \quad \quad  \quad \quad  -{\bf{R}}{(i,i + 1:k)^T}{\bf{R}}(i,i + 1:k)
		\end{split}}   &  \label{InvLDLSaveMemDivFree5}  \\
		{\begin{split}
				{\bf{R}}(1:i,i + 1:k) = \quad  \quad  \quad  \quad  \quad  \quad  \quad \quad \quad  \quad  \quad \quad \quad  \quad \quad \quad \quad \\
				\left[ {\begin{array}{*{20}{c}}
						{\eta {\bf{R}}(1:i - 1,i + 1:k) - {\bf{R}}(1:i - 1,i){\bf{R}}(i,i + 1:k)}\\
						{ - {\bf{R}}(i,i){\bf{R}}(i,i + 1:k)}
				\end{array}} \right]
		\end{split}}   &  \label{InvLDLSaveMemDivFree6} \\
		{\bf{D}}(i,i) = 1,   &  \label{iTo1InvLDLMemDivFreeGeneraLi93dse}
	\end{subnumcases}
\end{footnotesize}
which will be deduced in what follows.
Firstly, we compare (\ref{Xi2AAFFAA23094d3DivFree}) (with $i=i-1$)
and (\ref{LDLinvFreeNov17iAddVecGroup}) to obtain
$\eta={{\mathbf{\tilde \Xi }}_{k-i+1}}(1,1)$,
from which we  deduce
(\ref{InvLDLSaveMemDivFree1}) since  ${{\mathbf{\Xi }}_{k-i+1}}$ has been stored  in ${\bf{R}}(i:k,i:k)$.
Then   (\ref{InvLDLSaveMemDivFree2}) is deduced from (\ref{lii2sigmaiMinus1ada23})  since  ${{l_{ii}}}$ is stored in $\mathbf{R}(i,i)$,
while (\ref{InvLDLSaveMemDivFree3}) and (\ref{iTo1InvLDLMemDivFreeGeneraLi93dse}) are deduced from (\ref{GeneralEquGroup131toi1DDD}).
Lastly,
(\ref{InvLDLSaveMemDivFree4}),
(\ref{InvLDLSaveMemDivFree5})
and
(\ref{InvLDLSaveMemDivFree6})
are deduced
from
(\ref{GeneralEquGroup131toi1sigma}),
(\ref{Xi2Xi2vec342ds43}) and (\ref{Pi2PiSigma2390kds3ds}),
respectively.
Moreover, we can
compare
(\ref{pi2FFAA2390923DivFree}) (with $i=i-1$)
and
(\ref{GeneralEquGroup131toi1LLL}) to
deduce that ${{\bf{\tilde l}}_i^{}}$ is the first column of ${{\mathbf{\tilde \Pi }}_{i-1}}$,
while  ${{\mathbf{\tilde \Pi }}_{i-1}}$
is stored
in
${\bf{R}}(1:i-1,i:k)$. Thus
we can conclude
that ${{\bf{\tilde l}}_i^{}}$ is stored
in
${\bf{R}}(1:i-1,i)$.

In  (\ref{scalingRmatrixGENERALi}),  the scaling
of   ${\bf{R}}(1:i,i)$,     ${\bf{R}}(i,i+1:k)$ and $\eta$
is achieved by multiplying  ${\bf{R}}(1:i,i)$ and     ${\bf{R}}(i,i+1:k)$
by $c_k$, and multiplying $\eta$ by $c_k^2$. Then we
end up with  ${\bf{R}}(1:i,i) c_k$,      ${\bf{R}}(i,i+1:k) c_k$
and
$\eta c_k^2$,
where $\eta c_k^2$ is always between $0.5$ and $2$,
and $c_k$ is powers of $2$.

When we set  the initial  $\sigma =1$,
(\ref{iTo1InvLDLSaveMemDivFree}) is equivalent to (\ref{InvLDLSaveMemDivFree}) with $i=1$.
Then we can compute (\ref{InvLDLSaveMemDivFree}) iteratively for $i=1,2,\cdots,k$, to obtain
${\mathbf{\tilde{D}}}_{k}$ and ${\sigma}_{k}$,
and
cover the upper-triangular part of  ${\bf{R}}_k$  with
the upper-triangular
${\mathbf{\tilde L}}_k$,
where ${\mathbf{\tilde L}}_k$, ${\mathbf{\tilde{D}}}_{k}$ and ${\sigma}_{k}$  satisfy
${\mathbf{\tilde{L}}}_{k}   ({\mathbf{\tilde{D}}}_{k}   /{\sigma}_{k}    ){{\mathbf{\tilde{L}}}_{k}^{T}}  ={{\mathbf{R}}_k^{-1}}$.

The corresponding implementation is summarized in \textbf{Algorithm 14}, where we
set the initial ${\bf{\tilde D}}={{\mathbf{I}}}_{k}$ to avoid the execution
of   (\ref{iTo1InvLDLMemDivFreeGeneraLi93dse}) in each iteration.

\begin{algorithm}
\caption{:~\bf The parallel implementation of the division-free inverse $LDL^T$ factorization}
\begin{algorithmic}[1]
	\Require The upper-triangular part of a  Hermitian matrix ${\bf{R}}$
	\Ensure The division-free inverse $LDL^T$ factors of ${\bf{R}}$, i.e.,
	${\bf{\tilde L}}$, ${\bf{\tilde D}}$ and  $\sigma$ satisfying
	${\bf{\tilde L\tilde D}}{{\bf{\tilde L}}^T}/\sigma  = {{\bf{R}}^{ - 1}}$;
	\State  The initial ${\bf{\tilde D}}={{\mathbf{I}}}_{k}$ and $\sigma =1$;
	\For{$i=1:k$ ($k$ is the size of ${\bf{R}}$)}
	\State  $\eta  = {\bf{R}}(i,i)$;
	\State  ${\bf{R}}(i,i) = \sigma $;
	\State  Scaling of $\eta$,  ${\bf{R}}(1:i,i)$  and  ${\bf{R}}(i,i+1:k)$;
	\State  ${\bf{\tilde D}}(1:i - 1,1:i - 1) = \eta {\bf{\tilde D}}(1:i - 1,1:i - 1)$;
	\State  $\sigma  = \sigma  \times \eta $;
	\State  ${\bf{R}}(i + 1:k,i + 1:k) =\eta {\bf{R}}(i + 1:k,i + 1:k) \quad$
	$-{\bf{R}}{(i,i + 1:k)^T}{\bf{R}}(i,i + 1:k)$;
	\State  ${\bf{R}}(1:i,i + 1:k) =\quad \quad \quad \quad \quad \quad \quad \quad \quad \quad \quad \quad \quad$
	$\footnotesize  \left[ {\begin{array}{*{20}{c}}
			{\eta {\bf{R}}(1:i - 1,i + 1:k) - {\bf{R}}(1:i - 1,i){\bf{R}}(i,i + 1:k)}\\
			{ - {\bf{R}}(i,i){\bf{R}}(i,i + 1:k)}
	\end{array}} \right]$;
	\EndFor
	\State  The diagonal $\bf{\tilde D}$,  the  scalar  $\sigma$ and the upper-triangular ${\bf{\tilde L}} = {\mathop{\rm triu}\nolimits} \left( {{\bf{R}}} \right)$ (i.e., the upper-triangular part of ${\bf{R}}$ forms
	${\bf{L}}$);
\end{algorithmic}
\end{algorithm}

\ifCLASSOPTIONcaptionsoff
  \newpage
\fi

\end{document}